\documentclass{article}

\PassOptionsToPackage{numbers, compress}{natbib}
    \usepackage[final]{neurips_2024}

\usepackage[utf8]{inputenc} % allow utf-8 input
\usepackage[T1]{fontenc}    % use 8-bit T1 fonts
\usepackage{hyperref}       % hyperlinks
\usepackage{url}            % simple URL typesetting
\usepackage{booktabs}       % professional-quality tables
\usepackage{amsfonts}       % blackboard math symbols
\usepackage{nicefrac}       % compact symbols for 1/2, etc.m
\usepackage{microtype}      % microtypography
\usepackage{xcolor}         % colors
\usepackage{subfigure}

\usepackage{amsmath,amssymb,amsfonts}

\usepackage{graphicx}
\usepackage[inline]{enumitem}
\usepackage{amsmath}
\usepackage{multirow}
\usepackage{color}
\usepackage{subcaption}
\usepackage{makecell}

\usepackage{pifont}

\usepackage{xcolor,colortbl}

\usepackage{wrapfig}

\DeclareMathOperator{\argmin}{argmin}

\title{PointAD: Comprehending 3D Anomalies from Points and Pixels for Zero-shot 3D Anomaly Detection}
\author{    % Authors
    Qihang	Zhou\textsuperscript{\rm 1},
    Jiangtao Yan\textsuperscript{\rm 1},
    Shibo He\textsuperscript{\rm 1}\thanks{Corresponding authors.}\thickspace,
    Wenchao	Meng\textsuperscript{\rm 1},
    Jiming	Chen\textsuperscript{\rm 1} \\
\textsuperscript{\rm 1} College of
 Control Science and Engineering, Zhejiang University \\
\textsuperscript{\rm 1} \texttt{\{zqhang, jtaoy, s18he, wmengzju, cjm\}@zju.edu.cn} 
}
\def\eg{\emph{e.g.}} 
\def\ie{\emph{i.e.}}

\begin{document}

\maketitle

\begin{abstract}
Zero-shot (ZS) 3D anomaly detection is a crucial yet unexplored field that addresses scenarios where target 3D training samples are unavailable due to practical concerns like privacy protection. This paper introduces PointAD, a novel approach that transfers the strong generalization capabilities of CLIP for recognizing 3D anomalies on unseen objects. PointAD provides a unified framework to comprehend 3D anomalies from both points and pixels. In this framework, PointAD renders 3D anomalies into multiple 2D renderings and projects them back into 3D space. To capture the generic anomaly semantics into PointAD, we propose hybrid representation learning that optimizes the learnable text prompts from 3D and 2D through auxiliary point clouds. The collaboration optimization between point and pixel representations jointly facilitates our model to grasp underlying 3D anomaly patterns, contributing to detecting and segmenting anomalies of unseen diverse 3D objects. Through the alignment of 3D and 2D space, our model can directly integrate RGB information, further enhancing the understanding of 3D anomalies in a plug-and-play manner. Extensive experiments show the superiority of PointAD in ZS 3D anomaly detection across diverse unseen objects. Code is available at \url{https://github.com/zqhang/PointAD} 

\end{abstract}
\section{Introduction}
\vspace{-0.5em}

\begin{figure}[htbp]
    \centering
   \subfigure[Comparison on anomaly segmentation using different modalities. ]
    {\includegraphics[width=0.42\textwidth]{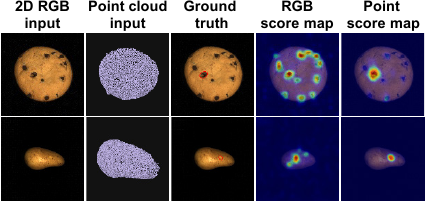}%
    \label{fig: zero-shot}}
    \hfil
    \subfigure[Comparison between ZS and unsupervised settings.]
   {\includegraphics[width=0.26\textwidth]{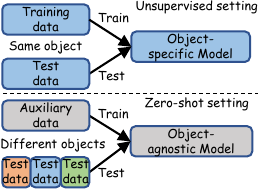}%
   \label{fig: different settings}
    }
    \hfil
    \subfigure[BTF Performance degradation on MVTec3D-AD.]{\includegraphics[width=0.26\textwidth]{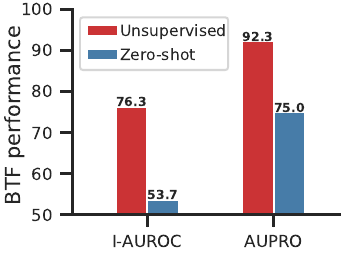}%
    \label{fig: performance degradation}
    }
    \vspace{-0.8em}
 \caption{Motivation of zero-shot 3D anomaly detection. \textbf{(a)}: \textbf{Top:} The hole on the cookies presents a similar appearance to the background. \textbf{Bottom:} Surface damage on the potato is unapparent to the object foreground. In these cases, leveraging RGB information makes it difficult to detect anomalies that imitate the color patterns of the background or foreground. However, effective recognition can be achieved by modeling the point relations within corresponding point clouds. (b) and (c) depicts the setting difference of ZS and unsupervised manner.}
 \vspace{-1.5em}
\end{figure}

Anomaly detection, a significant field within deep learning, has been widely applied to diverse domains, including industrial inspection~\citep{bergmann2019mvtec, bergmann2020uninformed, liznerski2020explainable, pang2021explainable, pang2021deep,roth2022towards, deng2022anomaly,jeong2023winclip, xie2023pushing, cao2023anomaly, gu2023anomalygpt, zhou2023anomalyclip}.
While 2D anomaly detection has been extensively studied by exploring RGB information~\cite{huang2022registration, you2022unified, sun2023when, tian2023self, Cao2024ASO, Li2024MuScZI}, real-world anomalies typically present themselves with abnormal spatial characteristics. Relying solely on RGB information poses challenges in detecting some anomalies in many cases, e.g., when the defect mimics the appearance of the object's background or foreground, as shown in Figure~\ref{fig: zero-shot}. The emerging field of 3D anomaly detection aims to unveil these spatial relations indicative of abnormal patterns~\citep{horwitz2023back,bergmann2023anomaly,rudolph2023asymmetric,chen2023easynet,wang2023multimoda,chu2023shape}. 

However, current 3D anomaly detection methods typically store normal point features during training and identify anomalies by measuring the distance between the test feature and these stored features~\cite{horwitz2023back,wang2023multimoda, chu2023shape}. They all depend on the assumption that target point clouds are available and entirely normal. This assumption does not hold in various situations when the training samples in the target dataset are inaccessible due to privacy protection (\eg, involvement of trade secrets) or the absence of target training data (\eg, a new product never seen before)~\cite{zhou2023anomalyclip}. Figure~\ref{fig: different settings} depicts the setting discrepancy between ZS 3D and unsupervised anomaly detection. These methods mentioned above, which detect anomalies by memorizing or reconstructing normal point features, have limitations in generalizing to unseen objects in Figure~\ref{fig: performance degradation}. While zero-shot (ZS) anomaly detection has been explored in 2D images~\citep{ming2022delving, zhou2023anomalyclip}, ZS 3D anomaly detection remains a research blank. It is a challenging task as ZS 3D anomaly detection necessitates the model to detect 3D anomalies across unseen point clouds with diverse class semantics, requiring a robust generalization capacity in the detection model. Recently, Vision-Language Models (VLMs) with their strong generalization capabilities have been applied to various downstream tasks~\cite{radford2021learning,zhou2022learning,rao2022denseclip,sun2022dualcoop,khattak2023maple,kirillov2023segment}. Particularly, CLIP has demonstrated its strong ZS performance to detect 2D anomalies~\cite{ming2022delving,jeong2023winclip,zhou2023anomalyclip}. Integrating CLIP into the detection model presents a potential solution to the challenging yet unexplored ZS 3D anomaly detection.

In this paper, we propose a unified framework, namely PointAD, to transfer the knowledge of CLIP to detect 3D anomalies in a ZS manner. PointAD comprehends point clouds from both 3D and 2D: \textbf{(1)} deriving 2D representations of point clouds via CLIP by rendering them from multiple views, \textbf{(2)} understanding 3D representations by projecting 2D representations back to 3D, and \textbf{(3)} enhancing 3D comprehension by additional regularization on 2D representations. After grasping point clouds from points and pixels, we propose hybrid representation learning to capture generic normality and abnormality w.r.t. point and pixel information into learnable text prompts~\cite{zhou2023anomalyclip}. Specifically, since 3D representation manifests its 2D renderings from different views, we treat each representation as one instance and achieve 3D representation aggregation via multiple instance learning (MIL). On this basis, PointAD explicitly aligns the 2D anomalies, rendered from 3D anomalies, to further enhance 3D understanding. We formulate these 2D anomaly recognition tasks from the multi-task learning (MTL) perspective. PointAD collaboratively learns point and pixel representations, promoting the in-depth understanding of underlying abnormal patterns and thus achieving superior ZS normality and abnormality point recognition. Furthermore, benefiting from collaboration optimization, PointAD can directly integrate additional RGB information and perform ZS multimodal 3D (M3D) detection without extra modules and retraining. The main contributions of this paper are summarized as follows:
\begin{itemize}[itemsep=4pt,topsep=-2pt,parsep=0pt]
    \item  To the best of our knowledge, we are the first to investigate the challenging yet valuable ZS 3D anomaly detection domain. We propose to transfer the strong recognition generalization of CLIP to detect and segment 3D anomalies over diverse objects.
    \item We introduce a novel ZS 3D anomaly detection approach called PointAD, which provides a unified framework to understand 3D anomalies from points and pixels. Hybrid representation learning is proposed to incorporate the generic normality and abnormality semantics into PointAD, enabling a thorough understanding of 3D anomalies. 
    \item PointAD can incorporate 2D RGB information in a plug-and-play manner for testing. In contrast to other methods that require storing RGB information separately, PointAD offers a unified framework to perform ZS M3D anomaly detection directly.
    \item Extensive experiments are conducted to demonstrate the superiority of our model in detecting and segmenting 3D anomalies, even outperforming some unsupervised SOTA methods that memorize normal features of target objects in certain metrics. We hope that our model will serve as a springboard for future research on ZS 3D and M3D anomaly detection. 
\end{itemize}

\vspace{-1em}
\section{Related Work}
\vspace{-0.8em}
\paragraph{3D Anomaly Detection}
 MVTec3D-AD~\citep{bergmann2023anomaly}, Eyecandies~\cite{Bonfiglioli_2022_ACCV}, and Real3D-AD~\cite{liu2023realdad} provide the point cloud anomalies and the corresponding 2D-RGB information. MVTec3D-AD bridges the connection between 3D and 2D anomaly detection. 3D-ST~\citep{bergmann2023anomaly} uses a teacher net to extract dense local geometric descriptors and design a student net to match such descriptors. AST~\cite{rudolph2023asymmetric} introduces asymmetric teacher and student net to further improve 3D anomaly detection. IMRNet~\cite{Li2023TowardsS3} and 3DSR~\cite{Zavrtanik2023CheatingDE} detect 3D anomalies by reconstruction errors. Instead of only using point clouds, BTF~\citep{horwitz2023back}, M3DM~\citep{wang2023multimoda, wang2024m3dm},  CPFM~\cite{cao2023complementary}, and SDM~\citep{chu2023shape} integrate point features and RGB pixel features to detect 3D anomalies. While these approaches exhibit commendable performance by storing object-specific normal point and pixel features within the unsupervised learning framework, such paradigms simultaneously limit their generalization capacity to point clouds from unseen objects, which is crucial to detecting anomalies when the target object is unavailable. To the best of our knowledge, no solution addresses this valuable yet challenging problem. To fill this gap, we introduce PointAD, designed to identify unseen anomalies across diverse objects. PointAD extends CLIP to the realm of ZS 3D anomaly detection and shows robust generalization in capturing generic normality and abnormality within point clouds.  Furthermore, PointAD serves as a unified framework, allowing seamless integration of point cloud and RGB modality without additional training.

\vspace{-1em}
\paragraph{3D Feature Extraction} Conventional methods of 3D feature extraction typically employ a point-based network like PointNet~\cite{qi2017pointnet} or PointNet++\cite{qi2017pointnet++} to extract 3D features from point clouds. Alternative approaches convert 3D data into a 2D format~\cite{su2015multi, goyal2021revisiting}, enabling 2D image backbones to process 3D information. PointCLIP~\cite{zhang2022pointclip} directly projects raw points onto image planes for efficiency, but this approach causes the produced depth map to lack geometric details. Instead, rendering-based methods~\cite{su2015multi, hamdi2021mvtn} generate 2D renderings by rendering point clouds, allowing for better preservation of local semantics. CPFM~\cite{cao2023complementary} stores normal features of these 2D renderings for unsupervised 3D anomaly detection. In this paper, we apply this rendering strategy to the source samples to capture generic anomaly semantics for recognizing abnormalities in unseen objects.

\vspace{-1em}
\paragraph{Prompt Learning} Instead of fine-tuning the whole network, prompt learning just optimizes the model to adapt the network to downstream tasks. CoOp~\cite{zhou2022learning, zhou2022conditional} introduces global context optimization to update learnable text prompts for few-shot recognition. DenseCLIP~\cite{rao2022denseclip} extends it to the dense classifications. More recently, AnomalyCLIP~\cite{zhou2023anomalyclip} proposes object-agnostic prompt learning to capture the generic normality and abnormality for images. Our model first introduces hybrid representation learning for ZS 3D anomaly detection, enabling the detection of anomalies and abnormal regions.

\vspace{-1.2em}
\section{PointAD}
\vspace{-0.8em}

\subsection{A Review of CLIP}
\vspace{-0.5em}
CLIP, a representative VLM, aligns visual representations to the corresponding textual representations, where an image is classified by comparing the cosine similarity between its visual representation and textual representations of given class-specific text prompts. Specifically, given an image $x_i$ and target class set $\mathcal{C}$, visual encoders output the global visual representation $f_i \in \mathbb{R}^d$ and local visual representations $f_i^m \in \mathbb{R}^{h \times w \times d}$, where $h$, $w$, and $d$ are the height, width, and dimension, respectively. Textual representations $g_c$ are encoded by textual encoder $\mathcal{T}$ with the commonly used text prompt template \texttt{A photo of a [c]}, where $ c \in \mathcal{C}$. The probability of $x_i$ belongs to $c$ can be computed as:
\begin{footnotesize}
\begin{equation}
\setlength{\abovedisplayskip}{3pt}
     % p(y = c | x_i) = P(g_c, f_i) = {\frac{exp(cos(g_c, f_i)/\tau)}{\sum_{c\in \mathcal{C}}exp(cos(g_c, f_i))/\tau)}},
     P(g_c, f_i) = {\frac{exp(cos(g_c, f_i)/\tau)}{\sum_{c\in \mathcal{C}}exp(cos(g_c, f_i))/\tau)}},
\label{equ: softmax}
\setlength{\belowdisplayskip}{0pt}
\end{equation}
\end{footnotesize}

where $cos(\cdot, \cdot)$ and $\tau$ represent the cosine similarity and temperature used in CLIP, respectively. The segmentation $S_{i(c)} \in \mathbb{R}^{h\times w}$ for class $c$ can be computed as $Seg(g_c, f_i^m)$, where each entry ($u$,$v$) is calculated as $P(g_c, f_{i,u,v}^m)$.
\vspace{-1.0em}

\subsection{Overview of PointAD}
\begin{figure}[h]
% \vskip 0.2in
\begin{center}
\centerline{\includegraphics[width=1\textwidth]{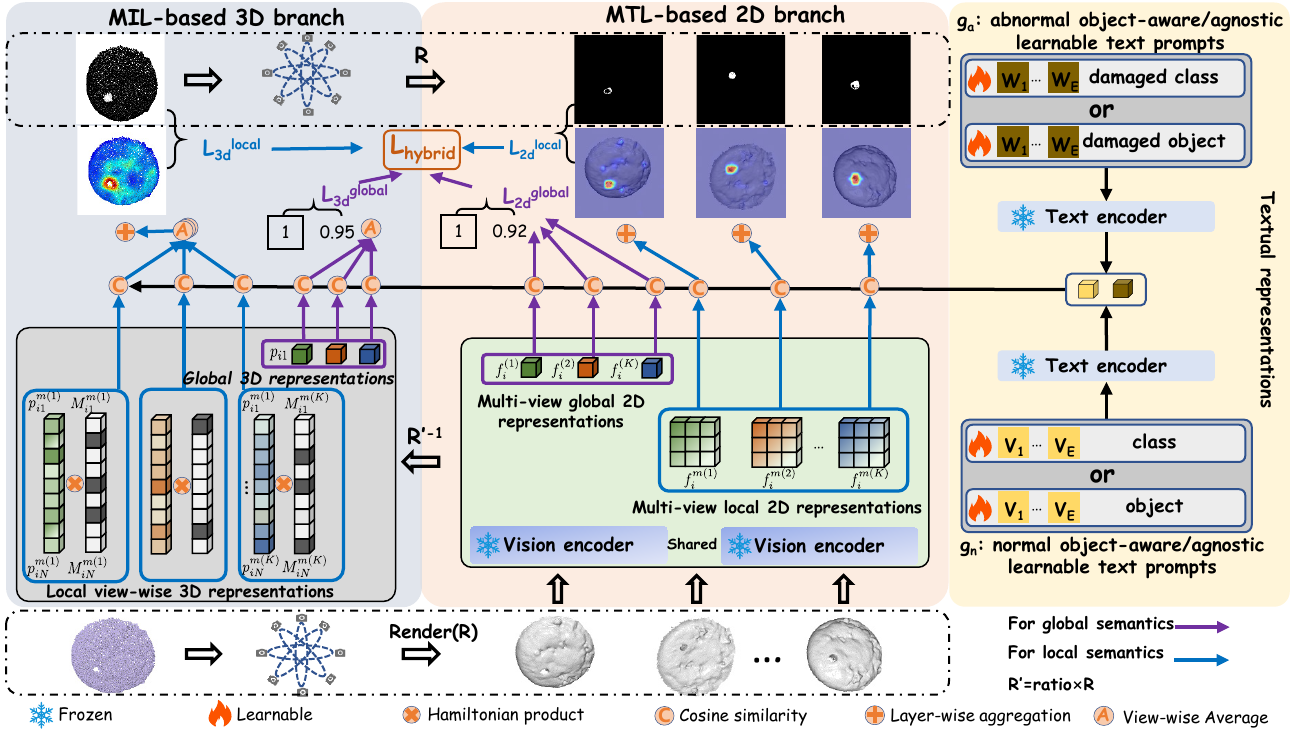}}
\vspace{-0.2em}
\caption{Framework of PointAD. To transfer the strong generalization of CLIP from 2D to 3D, point clouds and corresponding ground truths are respectively rendered into 2D renderings from multi-view. Then, vision encoder of CLIP extracts the renderings to derive 2D global and local representations. These representations are transformed into glocal 3D point representations to learn 3D anomaly semantics within point clouds. Finally, we align the normality and abnormality from both point perspectives (multiple instance learning) and pixel perspectives (multiple task learning) and propose a hybrid loss to jointly optimize the text embeddings from the learnable normality and abnormality text prompts, capturing the underlying generic anomaly patterns.}
\label{f2:overview}
\end{center}
\vspace{-3.0em}
\end{figure}

\vspace{-0.6em}
ZS 3D anomaly detection requires a strong generalization capacity to anomalies on unseen objects with diverse object semantics.
In this paper, we propose a unified framework, namely PointAD, to detect and segment 3D anomalies in a ZS manner. In Figure~\ref{f2:overview}, PointAD understands point clouds from both pixel and point perspectives. To make CLIP understand 3D point clouds, we first render point clouds from multiple views and extract the pixel representations of these generated 2D renderings via the visual encoder of CLIP. And then, we derive point representations by projecting these pixel representations back to 3D. Learning generic normality and abnormality is significant in recognizing across-object anomalies. We propose hybrid representative learning, which focuses on glocal point and pixel abnormality, to optimize normality and abnormality text prompts, enabling PointAD with strong generalization to identify 3D anomalies on diverse objects. Benefiting from the hybrid representation learning, PointAD can directly incorporate 2D RGB information during testing to achieve ZS M3D detection.

\vspace{-1.2em}
\subsection{Multi-View Rendering}
\vspace{-0.5em}
Multi-view projection is a crucial technology for understanding point clouds from 2D perspectives. Some multi-view projection approaches project point clouds into various depth maps, providing adequate shape information for class recognition~\cite{zhang2022pointclip}. However, in this paper, our objective is to learn both generic global and local anomaly semantics. Depth-map projection lacks sufficient resolution to represent fine-grained anomaly semantics accurately. Hence, we adopt high-precision rendering to preserve the original 3D information offline. Specifically, given an auxiliary dataset of point clouds $\mathcal{D}_{3d} = \{(x^{3d}_i, y^{3d}_i)\}_{i=1}^N$, we define the rendering matrix as $R^{(k)}$ for the $k$-$th$ view, with a total of $K$ views. We simultaneously render point clouds and point-level ground truths from different views to obtain their corresponding 2D renderings, which is given by $x_{i}^{(k)} = R^{(k)}(x^{3d}_i)$ and $y_{i}^{(k)} = R^{(k)}(y^{3d}_i)$, where $x_{i}^{(k)} \in \mathbb{R}^{H\times W}$ and $y_{i}^{(k)} \in \mathbb{R}^{H\times W}$ respectively represent the $k$-$th$ 2D renderings and corresponding pixel-level ground truth in the $i$-$th$ point cloud. Note that anomaly pixels are marked as 1, and normal pixels are marked as 0.

\vspace{-1.2em}
\subsection{Representations for 3D and 2D information}
\vspace{-0.5em}
PointAD aims to learn generic anomaly semantics from both 3D and 2D representations, enabling a comprehensive understanding of point and pixel anomaly patterns. For a point cloud $x^{3d}_i$, we first obtain the 2D renderings $\mathcal{X}_i = \{x_{i}^{(k)}\}_{k=1}^K$. Then, these renderings are encoded via the vision encoder of CLIP to obtain global 2D representations \(\mathcal{F}_i \hspace{-0.2em}=\hspace{-0.2em} \{f_{i}^{(k)}\}_{k=1}^K\), and local 2D representations $\mathcal{F}_i^{m} \hspace{-0.2em}=\hspace{-0.2em} \{f_i^{m(k)}\}_{k=1}^{K}$. 
As for point cloud representations, we consider that one point cloud will be projected into multiple 2D renderings. Consequently, global 3D representation $p_{i}$ and local 3D representations $p^m_{i}$ are expected to include their corresponding 2D representations in each view. Formally, $p_{i} = \{p_{i}^{(k)} | p_{i}^{(k)} = f_{i}^{(k)} \}_{k = 1}^{K}$ and $p^m_{i} = \{p^{m(k)}_{i} | p^{m(k)}_{i} = \{p^{m(k)}_{i,j}\}_{j=1}^n\}_{k=1}^K$, where $p^{m(k)}_{i,j} = \{ p^{m(k)}_{i,j} = f_{i,u,v}^{m(k)}, (u, v) = R'^{(k)}(a_{i,j} | b_{i,j}, c_{i,j})\} $ represents the $j$-$th$ point representation of $i$-$th$ point cloud in the $k$-$th$ view, whose 3D coordinate is ($a_{i,j}$, $b_{i,j}$, $c_{i,j}$). $R'^{(k)}$ is the rendering transformation between the point and pixel representation, derived as $R'^{(k)} = \frac{h}{H}R^{(k)}$.

Points at different positions may yield a different number of 2D representations as they are hidden by other points from a specific viewpoint. In this case, we introduce a view-wise visibility mask $M$, where $M_{i,j}^{k}$ indicates whether the $j$-$th$ point of the $i$-$th$ point cloud is visible in the $k$-$th$ view. We compare the depth of points projected into the same pixel in the same view and set the corresponding visibility mask to 1 for the point with the minimum depth, and to 0 for the other points. Let $\mathcal{Q}_{i,u,v}^{(k)}$ denote the depths set of all points that are projected into the same pixel indexed by $(u,v)$ in the $i$-$th$ point cloud in the $k$-$th$ view. $\mathcal{Q}_{i,u,v}^{(k)}$ and  $M_{i,j}^{k}$ are respectively given as 
$ % S_i^{(k)} = \{S_{i,u,v}^{(k)} \mid u \in h, v \in w  \} \nonumber \\ 
\mathcal{Q}_{i,u,v}^{(k)} = \{c_{i,j} \mid R'^{(k)}(a_{i,j}, b_{i,j}, c_{i,j}) = (u, v)\}_{j=1}^{n}$ and $
 M^{(k)}_{i,j} = \mathbb{I}(i,j,k=\argmin \limits_{i,j,k}\{c_{i,j} \mid c_{i,j} \in \mathcal{Q}_{i,u,v}^{(k)}\})$,
where $\mathbb{I}(\cdot)$ is an indicator function. Local 3D representations of the $i$-$th$ point cloud for the $k$-$th$ view are reformulated as:
$ p_{i}^{m(k)} = \{p^{m(k)}_{i,j}*M^{(k)}_{i,j}\}_{j=1}^{n}$.
\vspace{-0.6em}
\subsection{Hybrid representation learning}
\label{sec: hybrid loss}
\vspace{-0.3em}
The key of ZS 3D anomaly detection requires the model to capture generic anomaly semantics, rather than relying on specific object semantics. Since CLIP was originally pre-trained to align object semantics, such alignment harms the generalization capacity of CLIP to recognize anomalies on various objects. To adapt CLIP to 3D anomaly detection, we propose a hybrid representation learning, from both 3D and 2D perspectives, to globally and locally optimize textual representations. This enables PointAD to learn more representative text embedding for glocal anomaly semantics alignment.
Following previous work~\cite{zhou2023anomalyclip, zhou2022learning}, we randomly initialize two learnable text templates $t_n$ and $t_a$, in AnomalyCLIP~\cite{zhou2023anomalyclip} or CoOp manner~\cite{zhou2022learning}, to obtain more overall text embeddings $g_n$ and $g_a$ to recognize normality and abnormality, respectively. 
\vspace{-0.6em}
\begin{footnotesize}
\begin{align}   
    t_n =[V_1]\dots[V_E][object], \quad \quad\quad\quad\enspace\quad&\quad t_n =[V_1]\dots[V_E][class], \nonumber  \\
    \underbrace{t_a =[W_1]\dots[W_E][damaged][object]}_{\textrm{\small PointAD}}, &\quad  \underbrace{t_a =[W_1]\dots[W_E][damaged][class]}_{ \textrm{\small PointAD-CoOp}},\nonumber
\end{align}
\end{footnotesize}
\vspace{-1.5em}

where $V$ and $W$ are learnable word embeddings, respectively.

\vspace{-0.6em}
\paragraph{MIL-based 3D representation learning}To fully incorporate 3D glocal anomaly semantics into PointAD from point information, we respectively devised two losses to capture 3D global anomalies and local anomaly regionals. First, we compute the cosine similarity between the textual representation and its rendering global representations in each view. As point clouds are projected from different views, the resulting renderings in each view reflect certain parts of point clouds. We use view-wise MIL to integrate 2D global representations and then align global labels to capture the global semantics. Formally, the global 3D loss is defined as:
\vspace{-0.5em}
\begin{footnotesize}
\begin{equation}
    \textstyle L_{3d}^{global} = \frac{1}{N}  \sum_{i}\textrm{CrossEntropy}(\frac{1}{K} \textstyle \sum_{f_{i}^{(k)}\in p_i} P(g_c, f_{i}^{(k)}), \max{(y^{3d}_i)}). \nonumber
\end{equation}
\end{footnotesize}
\vspace{-1.5em}

As for local point anomaly semantics, we quantify the cosine similarity between textual representations and local representations of 2D renderings. Since points within point clouds are projected from different views, their projections in each view present part characteristics of themselves. We adopt the pixel-wise MIL to achieve the aggregation of point local representation. The point segmentation can be formulated mathematically as follows:
\vspace{-0.5em}
{\footnotesize
\begin{align}
    \textstyle  S_{i(a)}^{3d} = \frac{1}{K} \sum_{k}Seg(g_a, p_i^{m(k)}), S_{i(n)}^{3d} = \frac{1}{K} \textstyle \sum_{k}Seg(g_n,  p_i^{m(k)}). \nonumber
\end{align}
}%
However, deriving such 3D segmentation requires similarity computation for each point. It brings a significant memory burden, with a huge computational complexity of \( O(Knd) \), which is unaffordable for one NVIDIA RTX 3090 24GB GPU. To address this computational challenge, we resort to the rendering correspondence between points (3D space) and their corresponding pixels within each view (2D space). We first can rewrite 3D segmentation from the view perspective as $S_{i(a)}^{3d} =\frac{1}{K} \textstyle \sum_{k}S_{i(a)}^{3d(k)}$. Then, 
the $k$-$th$ division of 3D segmentation can be transformed into the 2D counterpart through the rendering projection $S_{i(a)}^{3d(k)}=(R^{(k)})^{(-1)}S_{i(a)}^{2d(k)} \otimes M_i^{(k)}$, where $\otimes$ is the Hamiltonian product. The $k$-$th$ 2D counterpart can be computed as $S_{i(a)}^{2d(k)} = \textrm{Up}(Seg(g_a, f_{i}^{m(k)}))$, where the operator $\textrm{Up}(\cdot)$ represents bilinear interpolation from feature space to 2D space. Finally, we can reformulate the 3D segmentation as follows: 
\vspace{-0.6em}
\begin{footnotesize}
\begin{align}
\textstyle S_{i(a)}^{3d} = \frac{1}{K} \sum_{k}\hspace{-0.2em}\Big ((R^{(k)})^{(-1)}\textrm{Up}(Seg(g_a, f_{i}^{m(k)})) \otimes M_i^{(k)}\Big).
\label{equ:3d}
\end{align}
\end{footnotesize}
\vspace{-1.5em}

From the equation, we can observe that the primary computation can be conducted in the feature space, with a computational complexity of \( O(Khwd) \). This is a substantial overhead reduction compared to \( O(Knd) \) since feature space is much smaller than 3D space, \ie, $ h\times w \ll n $. In our experiment, $h\times w = 24 \times 24 = 576$, while $n = 336\times336 = 112896$. With this transformation, the entire experiment can be conducted using only a single NVIDIA RTX 3090 24GB GPU. After that, Dice Loss is employed to precisely model the decision boundary of anomaly regions. Let $I$ represent a full-one matrix of the same size as $y_{i}^{3d}$. Formally, we define 3D local loss $L_{3d}^{local}$ : 
\vspace{-0.6em}
\begin{footnotesize}
\begin{equation}
\textstyle L_{3d}^{local}=\frac{1}{N} \sum_{i}\Big(\textrm{Dice}(S_{i(n)}^{3d}, I - y_{i}^{3d}) + \textrm{Dice}(S_{i(a)}^{3d}, y_{i}^{3d})\Big).  \nonumber
\end{equation}
\end{footnotesize}
\vspace{-1.0em}

\vspace{-1.0em}

\paragraph{MTL-based 2D representation learning} 
We further improve PointAD point understanding by capturing 2D glocal anomaly semantics into the object-agnostic text prompt template. We treat the anomaly recognition for one rendering from the point cloud as a task. Hence, we formulate the anomaly semantics learning for multiple 2D renderings as MTL. MTL-based 2D representation learning is divided into two parts for respective alignment to 2D global and local anomaly semantics.
For 2D global semantics, we use \textrm{CrossEntropy} to quantify the discrepancy between the textual representations and each global 2D representation. Global MTL-based 2D representation learning $L_{2d}^{global}$ is defined as:
\vspace{-0.8em}
\begin{footnotesize}
\begin{gather}
    \textstyle L_{2d}^{global} = \frac{1}{NK} \sum_{i, k}\textrm{CrossEntropy}(P(g_c, f_{i}^{(k)}), \max{(y_{i}^{(k)})}).  \nonumber
\end{gather}
\end{footnotesize}
\vspace{-1.5em}

Also, we focus on 2D abnormal regions to understand pixel-level anomalies. As the anomaly regions are typically smaller than normal regions, we employ \textrm{Focal} Loss to mitigate the class imbalance besides \textrm{Dice} Loss. Let $\oplus$ denote the concatenation operation. Local MTL-based 2D representation learning$L_{2d}^{local}$ is given as follows: 
\vspace{-0.6em}
\begin{footnotesize}
\begin{gather}
\setlength{\abovedisplayskip}{2pt}
\textstyle L_{2d}^{local} = \frac{1}{NK} \sum_{i,k} \textrm{Focal}(S_{i(n)}^{2d(k)} \oplus S_{i(a)}^{2d(k)}, y_{i}^{(k)}) + 
\textrm{Dice}(S_{i(n)}^{2d(k)}, I - y_{i}^{(k)})+ \textrm{Dice}(S_{i(a)}^{2d(k)}, y_{i}^{(k)}). \nonumber
\end{gather}
\end{footnotesize}

\vspace{-1.5em}
\subsection{Training and Inference}
\vspace{-0.5em}
PointAD detects 3D anomalies from both 3D and 2D perspectives and thus combing these above losses to derive hybrid loss $L_{hybrid}$. We minimize $L_{hybrid}$ to incorporate generic anomaly semantics into the text prompt from point and pixel spaces:
\vspace{-0.3em}
\begin{footnotesize}
\begin{align}
    L_{hybrid}  = L_{3d}^{global} + L_{3d}^{local} + L_{2d}^{global} + L_{2d}^{local}. \nonumber
\end{align}
\end{footnotesize}
\vspace{-2em}

During training, we minimize the hybrid loss $L_{hybrid}$, where the original parameters of CLIP are frozen to maintain its strong generalization. Since our model provides a unified framework to understand anomaly semantics from point and pixel, it can not only perform \textbf{ZS 3D anomaly detection} but also \textbf{M3D anomaly detection in a plug-and-pay way}. Next, we will introduce the inference process in detail:
\begin{figure*}[]
  \begin{minipage}{0.48\textwidth}
        \centering
%\vspace{-1em}
 \captionof{table}{Performance comparison on ZS 3D anomaly detection in "one-vs-rest" setting.}
 \vspace{-0.5em}
\label{table: Performance comparison on ZS 3D anomaly detection in "one-vs-rest" setting.}
\tiny
\setlength\tabcolsep{1pt} 
\begin{tabular}{cccccccc}
\toprule
\multirow{2}{*}{\makecell[c]{Detec. \\level}} & Dataset &\multicolumn{2}{c}{MVTec3D-AD(10)}  & \multicolumn{2}{c}{Eyecandies(10)} & \multicolumn{2}{c}{Real3D-AD(12)}   \\  \cline{2-8}
&Metric&I-AUROC&AP&I-AUROC&AP&I-AUROC&AP \\
\hline
\multirow{7}{*}{\makecell[c]{G.}} 
& CLIP + R.   & 61.2& 85.8 & 66.7& 69.2 & 68.8 & 72.3   \\

& Cheraghian  & 53.6& 81.7& 49.5& 48.1 & 50.3 & 54.4\\

& PoinCLIP V2   & 51.2& 80.1 & 46.1& 48.1 &  53.1 & 58.1 \\

& PointCLIP V2$_a$ & 51.1 & 80.6 & 44.4 & 47.0 & 57.5 
  & 58.3 
 \\
& AnomalyCLIP  & 56.4& 83.5 & 57.6& 59.0 & 55.2 & 57.1 \\

\rowcolor{gray!40}
& PointAD-CoOp & \textcolor{blue}{80.9} & \textcolor{blue}{93.9} & \textcolor{blue}{67.7} & \textcolor{blue}{71.8} & \textcolor{blue}{73.9} & \textcolor{blue}{75.9} \\

\rowcolor{gray!40}
& PointAD & \textcolor{red}{82.0} & \textcolor{red}{94.2} & \textcolor{red}{69.1} & \textcolor{red}{73.8} & \textcolor{red}{74.8} & \textcolor{red}{76.9} \\

\midrule

%\vspace{-0.1em}
&Metric&P-AUROC&AUPRO&P-AUROC&AUPRO&P-AUROC&AUPRO \\ \hline
\multirow{7}{*}{\makecell[c]{L.}}
    & CLIP + R.  & -& 54.4 & 81.2 & 37.9 & 45.9 & -  \\
& Cheraghian  & 88.2 & 57.0 & - & - & - & - \\
& PoinCLIP V2 & 87.4& 52.3 & 43.7 & - & 52.9 & -  \\
& PointCLIP V2$_a$  & 87.3& 52.3 & 44.2 & - & 52.2 & - \\
& AnomalyCLIP  & 88.9& 60.9 & 77.7 & \textcolor{blue}{43.4} & 50.3 & - \\

\rowcolor{gray!40}
& PointAD-CoOp & \textcolor{blue}{94.8} & \textcolor{blue}{82.0} & \textcolor{blue}{91.5} & \textcolor{red}{71.3} & \textcolor{blue}{72.6} & - \\

\rowcolor{gray!40}
& PointAD & \textcolor{red}{95.5} & \textcolor{red}{84.4} & \textcolor{red}{92.1} & \textcolor{red}{71.3} & \textcolor{red}{73.5} & - \\

\bottomrule
\end{tabular}%
    % \vspace{-0.8em}
    \end{minipage} 
    \hfill
    \hfill
\begin{minipage}{0.48\textwidth}
    \centering
    %\vspace{-1em}
    \captionof{table}{Performance comparison on ZS M3D anomaly detection in "one-vs-rest" setting.}%
    \vspace{-0.5em}
\label{table: Performance comparison on ZS M3D
anomaly detection in "one-vs-rest" setting.}
\tiny
\setlength\tabcolsep{1pt} 
\begin{tabular}{cccccc}
\toprule
\multirow{2}{*}{\makecell[c]{Detec. \\level}}& Dataset &\multicolumn{2}{c}{MVTec3D-AD(10)}  & \multicolumn{2}{c}{Eyecandies(10)}  \\ \cline{2-6}
&Metric&I-AUROC&AP&I-AUROC&AP \\
\hline
\multirow{7}{*}{\makecell[c]{MG.}} 
& CLIP + R.  & 60.4& 86.4 & 73.0 & 73.9\\
& Cheraghian  &-&-&-& - \\
& PoinCLIP V2   &49.8 & 79.3 & 46.9 & 49.9 \\
& PointCLIP V2$_a$ & 49.4 & 79.8 & 48.5 & 50.5\\
& AnomalyCLIP  & 66.2 & 87.6 & 65.0 & 67.5 \\
\rowcolor{gray!40}
& PointAD-CoOp & \textcolor{blue}{83.4} & \textcolor{blue}{94.9} & \textcolor{blue}{73.7} & \textcolor{blue}{76.0} \\
\rowcolor{gray!40}
& PointAD & \textcolor{red}{86.9} & \textcolor{red}{96.1} & \textcolor{red}{77.7} & \textcolor{red}{80.4} \\
\midrule
%\vspace{-0.1em}
&Metric&P-AUROC&AUPRO&P-AUROC&AUPRO \\ \hline
\multirow{7}{*}{\makecell[c]{ML.}}
& CLIP + R.  &- & 56.0 & 78.0& 31.8  \\

& Cheraghian  &-& - &-& - \\

& PoinCLIP V2 &78.3& 49.4 &46.0& - \\

& PointCLIP V2$_a$  &79.5 &51.6 & 46.2 & -  \\

& AnomalyCLIP  &91.6 & 70.9 &85.0& 56.2 \\

\rowcolor{gray!40}
& PointAD-CoOp & \textcolor{blue}{96.5} & \textcolor{blue}{88.8} & \textcolor{blue}{94.9} & \textcolor{blue}{83.6} \\

\rowcolor{gray!40}
& PointAD & \textcolor{red}{97.2} & \textcolor{red}{90.2} & \textcolor{red}{95.3} & \textcolor{red}{84.3} \\

\bottomrule
\end{tabular}%
    \end{minipage} 
    \vspace{-0.8em}
\end{figure*}

\vspace{-0.9em}
\paragraph{ZS 3D/M3D inference}
Given a point cloud $x_{i}^{3d}$, we regard the 3D segmentation (See Equ.~\ref{equ:3d}) as the anomaly score map: $A^m_i \hspace{-0.2em}=\hspace{-0.2em}$ $G_{\sigma}(S_{i(a)}^{3d}$), where $G_{\sigma}(\cdot)$ represents the Gaussian filter. The global anomaly score incorporates glocal anomaly semantics and is computed as $A_i^s = \frac{1}{2}(\frac{1}{K}\sum_{f_{i}^{(k)}\in \mathcal{F}_i }\hspace{-0.2em}P(g_c, f_{i}^{(k)})\hspace{-0.1em}+\hspace{-0.2em}\max{(A^m_i)})$. When the RGB counterpart is available for testing, PointAD could directly integrate RGB information by feeding RGB images to 2D branch to derive 2D representations. We project these 2D representations back to 3D branch to respectively compute RGB anomaly score map and anomaly score as  $ A^{m(rgb)}_i = P(g_c, f_{i}^{(rgb)})$ and $A_i^{s(rgb)} = G_{\sigma}(S_{i(a)}^{3d(rgb)})$. The final multimodal anomaly score map and anomaly score are defined as $A_i^{m(mod)} = \frac{1}{2}G_{\sigma}\big(A_i^m + A^{m(rgb)}_i\big)$ and $A_i^{s(mod)} = \frac{1}{2}\left[\frac{1}{2}(A_i^{s(rgb)}+A_i^s)+\max{(A_i^{m(mod)})}\right]$, respectively.

\vspace{-1.2em}
\section{Experiment}
\vspace{-0.5em}
\subsection{Experiment Setup}
\vspace{-0.5em}
\paragraph{Dataset} We evaluate the performance of ZS 3D anomaly detection on three public datasets including, MVTec3D-AD, Eyecandies and Real3D-AD. MVTec3D-AD, Eyecandies, and Real3D-AD are multi-class datasets and respectively contain 10 classes, 10 classes, and 12 classes. Since these training datasets only contain all normal samples, \textbf{we use the common zero-shot setting one-vs-rest, where an object test dataset is used to fine-tune PointAD and assess the ZS anomaly detection for the remaining objects.} We also explore a more challenging setting: \textbf{cross-dataset ZS generalization, which requires the detection model to generalize to anomalies on other datasets.} For point cloud anomaly detection, we only use point clouds to detect and localize 3D anomalies. In M3D anomaly detection, the 2D RGB information is utilized only for testing. To comprehensively analyze PointAD, we utilize four metrics to assess its performance in both anomaly detection and segmentation.

\begin{figure*}[h]
% \vskip 0.2in
\begin{center}
\centerline{\includegraphics[width=1\textwidth]{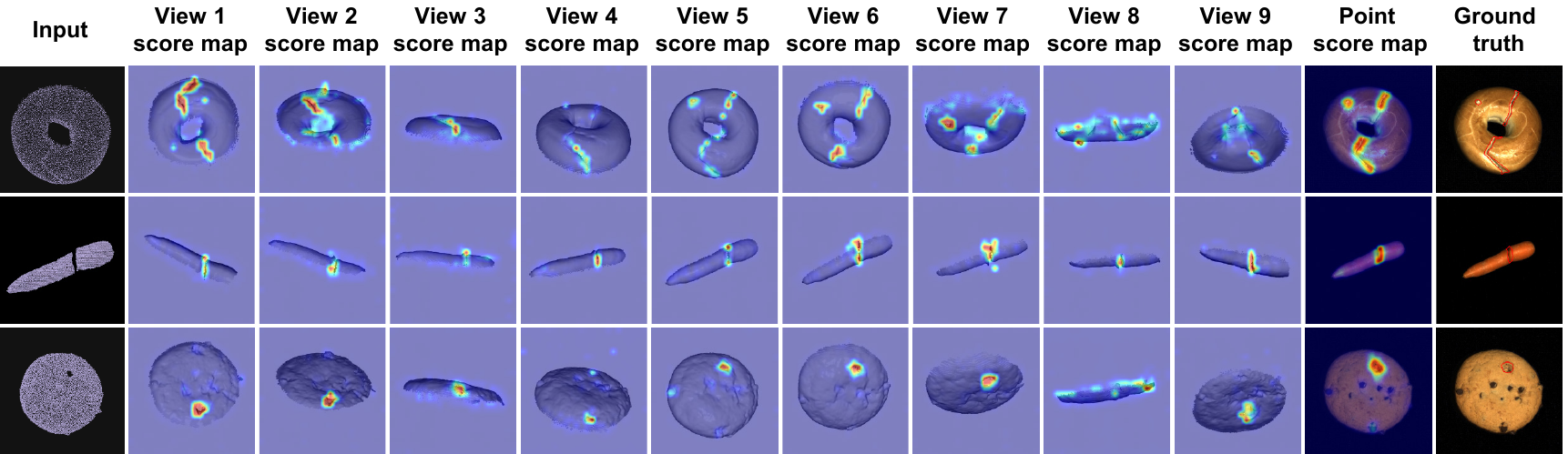}}
\vspace{-0.5em}
\caption{Visualization on anomaly score maps in ZS 3D anomaly detection. Point clouds of diverse objects are input into PointAD to generate 2D and 3D representations. Each row visualizes the anomaly score maps of 2D renderings from different views, and the final point score maps are also presented. More visualizations are provided in Appendix~\ref{appendix: addition_visualization}.}
\label{f5:visualization}
\end{center}
\vspace{-2.8em}
\end{figure*}

\subsection{Implementation Details \& Baselines}
\label{section: implementation details}
\vspace{-0.5em}
Both point clouds and 2D renderings are resized to 336 $\times$ 336. We use Open3d library to generate 9 views by rotating point clouds along the X-axis at angles of $\{-\frac{4}{5}\pi, -\frac{3}{5}\pi, -\frac{2}{5}\pi, -\frac{1}{5}\pi, 0, \frac{1}{5}\pi, \frac{2}{5}\pi, \frac{3}{5}\pi, \frac{4}{5}\pi\}$ for most categories. We circularly set the rendering angles, evenly distributing the angles between $-\pi$ to $\pi$. The backbone of PointAD is the pre-trained CLIP model (\texttt{VIT-L/14@336px} in \texttt{open\_clip}). Following~\cite{zhou2023anomalyclip}, we improve the local visual semantics of vision encoder of CLIP without modifying its parameters. During training, we keep all parameters of CLIP frozen and set the learnable word embeddings in object-agnostic text templates to $12$. All experiments were conducted on a single NVIDIA RTX 3090 24GB GPU using PyTorch-2.0.0. As there is no work to explore the field of ZS 3D anomaly detection, we make a great effort to provide these comparisons. We apply the original CLIP to our framework for 3D detection, called CLIP + Rendering. Also, we reproduce SOTA 3D recognition works including PointCLIP V2~\cite{zhu2023pointclip} and Cheraghian~\cite{cheraghian2022zero}, and adapt them for ZS 3D anomaly detection. We compare the SOTA 2D anomaly detection approach AnomalyCLIP~\cite{zhou2023anomalyclip} by fine-tuning it on depth maps. PointAD by default uses object-agnostic text prompts, whereas PointAD-CoOp employs object-aware prompts. Appendix~\ref{appendix: implementation details} and~\ref{appendix: baselines} provide more details on implementation and baselines.

\vspace{-1.0em}
\subsection{Main Results}
\label{main results}
\vspace{-0.6em}
We fine-tuned PointAD on three objects on MVTec3D-AD, Eyecandies, and Real3D-AD. Over three runs, the averaged results on \textbf{one-vs-rest} and \textbf{cross-dataset} settings are reported. We use the metric pairs (I-AUROC\% $\%\uparrow$ and AP\% $\%\uparrow$) and (P-AUROC\% $\%\uparrow$ and AUPRO\% $\%\uparrow$) to evaluate the glocal detection performance, respectively. Details of experimental settings see Appendix~\ref{Evaluation}. The best and second-best results in ZS are highlighted in~\textcolor{red}{Red} and~\textcolor{blue}{Blue}. G. and L. represent 3D global and local anomaly detection.  M3D global and local anomaly detection are abbreviated as  MG. and ML.

\vspace{-0.8em}
\paragraph{ZS 3D anomaly detection} Table~\ref{table: Performance comparison on ZS 3D
anomaly detection in "one-vs-rest" setting.} presents the comparison of ZS 3D performance. Compared to the point-based method Cheraghian and the projection-based method PointCLIP V2, PointAD achieves superior performance on ZS 3D anomaly detection over all three datasets. Especially, it outperforms CLIP + Rendering from 61.2\% to 82.0\% I-AUROC and from 85.8\% to 94.2\% AP on MVTec3D-AD. In addition, PointAD achieves superior segmentation performance on ZS 3D anomaly detection, improving MVTec3D-AD by a large margin compared to Cheraghian from 88.2\% to 95.5\% P-AUROC and from 57.0\% to 84.4\% AUPRO. This improvement in overall performance is attributed to PointAD adapting CLIP's strong generalization to glocal anomaly semantics through hybrid representation learning. In addition, PointAD advances PointAD-CoOp across all datasets by blocking the class semantics in text prompts~\cite{zhou2023anomalyclip}.

\vspace{-1.0em}
\paragraph{ZS M3D anomaly detection} We also compare the ZS M3D anomaly detection when RGB information is available for testing. As shown in Table~\ref{table: Performance comparison on ZS M3D
anomaly detection in "one-vs-rest" setting.}, the results indicate that PointAD can integrate additional RGB information and further boost its performance from 82.0\% to 86.9\% AUROC and from 94.2\% to 96.1\% AP for global semantics on MVTec3D-AD. Additionally, as for local semantics, the performance improves from 95.5\% to 97.2\% P-AUROC and from 84.4\% to 90.2\% AUPRO. A large performance gain is also obtained on Eyecandies and Real3D-AD. While other methods improve their performance in some metrics, they still suffer from performance degradation in other metrics due to inefficient integration of the two modalities. Instead, PointAD achieves overall improvement across all metrics by incorporating explicit joint constraints on both point and pixel information.

\begin{wrapfigure}{r}{0.7\textwidth}
  \begin{minipage}{0.32\textwidth}
        \centering
    
\vspace{-1.2em}
 \captionof{table}{Performance comparison on ZS 3D
anomaly detection in cross-dataset setting.}
\vspace{-0.5em}
\label{table: Performance comparison on ZS 3D
anomaly detection in across-dataset setting.}
\tiny
\setlength\tabcolsep{1pt} 
\begin{tabular}{cccccc}
\toprule
\multirow{2}{*}{\makecell[c]{Detec. \\level}}& Dataset  & \multicolumn{2}{c}{Eyecandies(10)} & \multicolumn{2}{c}{Real3D-AD(12)}   \\  \cline{2-6}
&Metric&I-AUROC&AP&I-AUROC&AP \\
\hline
\multirow{3}{*}{\makecell[c]{G.}} 
& PointCLIP V2$_a$ & 45.2 & 48.0 & 57.4 & 58.8 \\

& AnomalyCLIP & 56.3 & 57.1 & 52.7 & 55.7 \\

\rowcolor{gray!40}
& PointAD-CoOp & \textcolor{blue}{69.1} & \textcolor{blue}{73.8} & \textcolor{blue}{74.8} & \textcolor{blue}{76.9} \\

\rowcolor{gray!40}
& PointAD & \textcolor{red}{69.5} & \textcolor{red}{74.3} & \textcolor{red}{75.9} & \textcolor{red}{77.9} \\

\midrule

%\vspace{-0.1em}
&Metric&P-AUROC&AUPRO&P-AUROC&AUPRO \\ \hline
\multirow{3}{*}{\makecell[c]{L.}}
& PointCLIP V2$_a$ & 43.9 & - & 51.9 & -\\

& AnomalyCLIP  & \textcolor{blue}{79.6} & 45.4 & 50.3 & - \\

\rowcolor{gray!40}
& PointAD-CoOp & \textcolor{red}{91.8} & \textcolor{blue}{70.5} & \textcolor{blue}{70.1} & - \\

\rowcolor{gray!40}
& PointAD & \textcolor{red}{91.8} & \textcolor{red}{71.4} & \textcolor{red}{71.6} & - \\

\bottomrule
\end{tabular}%
  
    \end{minipage} 
    \hfill
    \hfill
\begin{minipage}{0.32\textwidth}
    \centering
    \vspace{-1.2em}
    \captionof{table}{Performance comparison on ZS M3D
anomaly detection in cross-dataset setting.}%
\vspace{-0.5em}
\label{table: Performance comparison on ZS M3D
anomaly detection in across-dataset setting.}
\tiny
\setlength\tabcolsep{1pt} 
\begin{tabular}{cccc}
\toprule
\multirow{2}{*}{\makecell[c]{Detec. \\level}}& Dataset & \multicolumn{2}{c}{Eyecandies(10)}   \\  \cline{2-4}
&Metric&I-AUROC&AP\\
\hline
\multirow{3}{*}{\makecell[c]{MG.}} 

& PointCLIP V2$_a$ & 48.5 & 50.9 \\

& AnomalyCLIP  & 65.7& 68.1 \\

\rowcolor{gray!40}
& PointAD-CoOp & \textcolor{blue}{76.3} & \textcolor{blue}{78.9} \\

\rowcolor{gray!40}
& PointAD & \textcolor{red}{78.6} & \textcolor{red}{80.8} \\

\midrule

%\vspace{-0.1em}
&Metric&P-AUROC&AUPRO\\ \hline
\multirow{3}{*}{\makecell[c]{ML.}}

& PointCLIP V2$_a$  &46.3 & -  \\

& AnomalyCLIP  & 86.2& 61.3  \\

\rowcolor{gray!40}
& PointAD-CoOp & \textcolor{red}{94.4} & \textcolor{blue}{80.3} \\

\rowcolor{gray!40}
& PointAD & \textcolor{blue}{94.0} & \textcolor{red}{80.7} \\

\bottomrule
\end{tabular}%
    \end{minipage} 
\end{wrapfigure}

\vspace{-0.8em}
\paragraph{Cross-dataset ZS anomaly detection} We perform the cross-dataset anomaly recognition to further evaluate the zero-shot capacity of PointAD, where we use one object as the auxiliary and test objects with totally different semantics and scenes in another dataset. We compare all baselines that need fine-tuning. From Table~\ref{table: Performance comparison on ZS
3D anomaly detection in across-dataset
setting.} and M3D from Table~\ref{table: Performance comparison on ZS
M3D anomaly detection in across-dataset
setting.}, PointAD demonstrates strong cross-dataset generalization performance on Eyecandies and Real3D-AD, with nearly no obvious performance decay compared to the one-vs-rest setting. The strong transfer ability highlights its robust generalization capabilities in detecting anomalies in objects with unseen semantics and backgrounds.

% \paragraph{Ensembling knowledge to other methods}

\begin{figure}[t]
  \centering
    \subfigure[Multimodal visualization with hybrid loss.]
    {\includegraphics[width=0.48\textwidth]{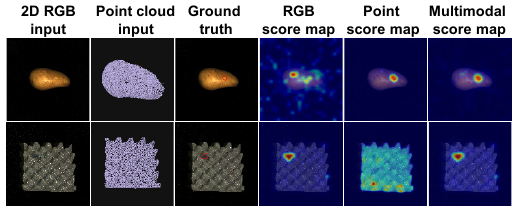}%
    \label{fig6: hybrid loss}}\vspace{-0.8em}
    \subfigure[Multimodal visualization without 2D glocal loss.]
    {\includegraphics[width=0.48\textwidth]{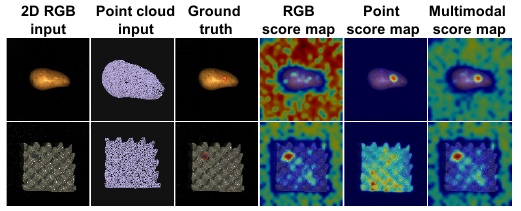}%
    \label{fig6: hybrid loss without 2D glocal loss}}
    \vspace{-0.1em}
 \caption{Visualization comparison between PointAD with hybrid loss and without.}
 \label{fig:auroc_curves}
 \vspace{-1.5em}
\end{figure}

\vspace{-0.8em}
\subsection{Result Analysis}
\vspace{-0.6em}
\paragraph{Visualization analysis.} To intuitively present the strong generalization capacity of our model to unseen anomalies, we visualize the anomaly score maps of the 3D and corresponding 2D counterparts of PointAD on MVTec3D-AD. As shown in Figure~\ref{f5:visualization}, PointAD reveals abnormal spatial relationships of points and further captures the generic point anomaly patterns across diverse objects. And, we also visualize the anomaly score map of corresponding 2D counterparts, where 3D point anomalies are transformed into 2D pixel anomalies. It can be observed that PointAD also has a strong detection ability on such 2D anomalies. The strong representative pixel representations from multiple views facilitate more precise 3D anomaly detection. Quantitative results are provided in Section~\ref{sec: module ablation}. The strong 3D and 2D detection capabilities of PointAD are from hybrid representation learning, which not only enables PointAD to capture the 3D anomalies but also explicitly constrains 2D representations.

\vspace{-1.0em}
\paragraph{How multimodality makes PointAD accurate.}
PointAD is a unified framework that can not only capture point anomalies but also handle 2D information in a plug-and-play manner. As shown in Figure~\ref{fig6: hybrid loss}, we visualize M3D results of PointAD on MVTec3D-AD. The surface damage on the potato presents a similar appearance to the object foreground, which makes it difficult to detect this anomaly with RGB information. On the contrary, the point relations for the color stain on foam are the same as those of normal, but they have a clear distinction in the RGB information. PointAD can integrate these two modalities, thereby complementing their respective advantages. We further investigate the reason why PointAD can directly leverage both modalities. For this purpose, we experiment without 2D glocal loss. As shown in Figure~\ref{fig6: hybrid loss without 2D glocal loss}, without 2D glocal loss, significant noise disrupts and even covers the RGB score maps, resulting in unpromising multimodal fusions. This illustrates the importance of explicit constraints on the 2D space. Hence, we conclude that the robust multimodal detection capability of our model stems from the collaboration optimization in both 3D and 2D spaces during training. We provide more analysis about failure cases and the computation overhead in Appendix~\ref{Failure Cases} and ~\ref{Complexity analysis}.

\vspace{-0.5em}
\section{Ablation Study}
\vspace{-0.3em}
\paragraph{Module Ablation.}
\vspace{-0.6em}
Here, we investigate the effectiveness of the proposed main technologies by progressively adding the proposed modules. Table~\ref{table: Ablation on the proposed modules.} illustrates that \textit{vanilla}, which represents the aforementioned CLIP + rendering, performs poor results on both 3D and M3D anomaly detection because CLIP focuses on alignment for 2D object semantics instead of anomaly semantics. With the 3D global branch, we incorporate the global anomaly semantics into PointAD, improving overall performance in local and global detection. After adding the 3D local branch, the performance is further improved, while the pixel-level performance on M3D detection suffers from performance degradation. This is attributed to the absence of 2D constraints, leading to inefficient multimodality fusion as we integrate 2D RGB information in a plug-and-play way. The inclusion of 2D global branch explicitly incorporates 2D anomaly information, which makes PointAD obtain overall performance gain. Finally, by further focusing on 2D anomaly regions, PointAD has a deeper understanding of point clouds from 2D representations and promotes multimodality fusion. Therefore, our model notably boosts the multimodal segmentation performance from 92.9\% to 97.2\% P-AUROC and from 84.4\% to 90.2\% AUPRO.

\vspace{-0.8em}
\begin{figure}[htbp]
	\begin{minipage}{0.40\textwidth}
    \centering
            \captionof{table}{Ablation on rendering number.}
            \vspace{-0.6em}
             \label{table: Ablation on rendering number.}
             \tiny
                \setlength\tabcolsep{3pt} 
                \begin{tabular}{ccccc}
                \toprule
                \multirow{2}{*}{View number} & \multicolumn{2}{c}{Point detection} & \multicolumn{2}{c}{Multimodal detection} \\
                 & Local & Global & Local & Global \\
                \midrule
                1    & (94.1, 79.1)& (72.6, 90.1)  & (96.0, 87.6)& (80.4, 93.9)\\
                3    & (95.2, 82.5)& (76.8, 92.1)  & (96.9, 89.5)& (83.7, 95.1)\\
                5    & (95.3, 84.3)& (80.8, 93.8)  & (97.1, 89.8)& (85.9, 95.7)\\
                7    & (95.3, \textcolor{red}{84.9})& (81.3, \textcolor{blue}{93.9})  & (\textcolor{red}{97.3}, 90.0)& (\textcolor{blue}{86.5}, \textcolor{blue}{95.9})\\
                9    & (\textcolor{red}{95.5}, \textcolor{blue}{84.4})& (\textcolor{red}{82.0}, \textcolor{red}{94.2})  & (\textcolor{blue}{97.2}, \textcolor{red}{90.2})& (\textcolor{red}{86.9}, \textcolor{red}{96.1})\\
                11   & (\textcolor{blue}{95.4}, 83.8)& (\textcolor{blue}{81.7}, \textcolor{red}{94.2})  & (97.1, \textcolor{blue}{90.1})& (85.4, 95.5)\\
                \bottomrule
                \end{tabular}
	\end{minipage}
	%\qquad
    \hfill
    \hfill
	\begin{minipage}{0.55\textwidth}
    \centering
               \captionof{table}{Ablation on the proposed modules.}
               \vspace{-0.6em}
             \label{table: Ablation on the proposed modules.}
             \tiny
                \setlength\tabcolsep{2.5pt} 
                \begin{tabular}{p{18pt}p{18pt}p{18pt}p{18pt}cccc}
                \toprule
                \multicolumn{4}{c}{Module} & \multicolumn{2}{c}{Point detection} & \multicolumn{2}{c}{Multimodal detection} \\
                  $L_{3d}^{global}$ &$ L_{3d}^{local}$ & $L_{2d}^{global}$ & $L_{2d}^{local}$ & Local & Global & Local & Global \\
                \midrule
                	
                & & & &   (-, 54.4) & (61.2, 85.8)  & (-, 56.0) & (60.4, 86.4)\\

                \ding{51}& &  &  &   (91.9, 71.7)& (75.5, 91.9)  & (92.6, 81.6)& (80.4, 93.9)\\
                \ding{51}& \ding{51} & &  &   (95.2, 82.7)& (81.3, 94.1)  & (92.0, 81.4)& (83.9, 95.0)\\

               &  & \ding{51} &  \ding{51}&  (93.9, 82.8)&	(79.3, 91.6)	&(91.0, 82.2)	&(82.6, 94.1)\\

                 & \ding{51} &\ding{51} & \ding{51} &  (\textcolor{blue}{95.5}, \textcolor{red}{84.7})&	(81.8, 92.3) &(\textcolor{blue}{96.1}, \textcolor{red}{90.6}) &	(83.7, 95.1)\\
                                 
                 \ding{51}&\ding{51} & \ding{51} & &   (\textcolor{red}{95.6}, 82.5)& (\textcolor{red}{82.4}, \textcolor{red}{94.5})  & (92.9, 84.4)& (\textcolor{blue}{85.5}, \textcolor{blue}{95.6})\\
                 \ding{51}& \ding{51} & \ding{51} & \ding{51}&   (\textcolor{blue}{95.5}, \textcolor{blue}{84.4})& (\textcolor{blue}{82.0}, \textcolor{blue}{94.2})  & (\textcolor{red}{97.2}, \textcolor{blue}{90.2})& (\textcolor{red}{86.9}, \textcolor{red}{96.1})\\
                \bottomrule
                \end{tabular}
	\end{minipage}
\end{figure}

\vspace{-0.8em}
\paragraph{View Number Ablation.}
\label{sec: module ablation}
\vspace{-0.5em}
PointAD interprets point clouds from 2D renderings, and the quantity of rendering views directly affects the 3D original information acquired by PointAD. Table~\ref{table: Ablation on rendering number.} depicts that the appropriate number of views benefits point understanding from informative views while alleviating negative effects of subpar views. More ablations about the length of learnable prompts, layers of intermediate vision features, and the number of training sets are provided in Appendix~\ref{appendix: hyperparameter}.

\vspace{-1.2em}
\section{Conclusion}
\vspace{-0.8em}
This paper takes the first attempt to study the challenging yet underexplored tasks of ZS 3D and M3D anomaly detection. We propose a unified framework, namely PointAD, to transfer the strong generalization of CLIP to 3D point clouds. PointAD understands 3D anomalies from 3D and 2D spaces. Benefiting from hybrid representation learning, PointAD can recognize generic 3D normality and abnormality across diverse objects and directly integrate RGB information for ZS M3D. Extensive experiments demonstrate the superior ZS detection capacity of our model, whether single modality or multimodality. \textbf{Code will be made available once the paper is accepted.}
\vspace{-0.8em}
\paragraph{Limitations} PointAD utilizes fixed rendering angles to generate 2D renderings across diverse objects. While experimental results demonstrate its superiority, the development of a fine-grained filtering mechanism to select high-quality 2D renderings, particularly for revealing anomalies, remains an avenue for future research.

\vspace{-0.5em}
\paragraph{Broader Impact}
Our paper aims to enhance automated detection and decision-making in smart manufacturing, which does not involve any potential ethical risks. Since the collection of 3D samples is more labor-intensive and costly, our research on using vision-language models for zero-shot point cloud detection can have significant societal impacts, especially in scenarios where target 3D training samples are unavailable due to privacy concerns or the absence of products. We hope that our first exploration of the ZS 3D anomaly detection field could pave the way for further research in this emerging field.

\subsubsection*{Acknowledgments}
This work was supported by NSFC U23A20326 and NSFC 62088101 Autonomous Intelligent Unmanned Systems.
%%%%%%%%%%%%%%%%%%%%%%%%%%%%%%%%%%%%%%%%%%%%%%%%%%%%%%%%%%%%%%%%%%%%%%%%%%%%%%%
%%%%%%%%%%%%%%%%%%%%%%%%%%%%%%%%%%%%%%%%%%%%%%%%%%%%%%%%%%%%%%%%%%%%%%%%%%%%%%%
% APPENDIX
%%%%%%%%%%%%%%%%%%%%%%%%%%%%%%%%%%%%%%%%%%%%%%%%%%%%%%%%%%%%%%%%%%%%%%%%%%%%%%%
%%%%%%%%%%%%%%%%%%%%%%%%%%%%%%%%%%%%%%%%%%%%%%%%%%%%%%%%%%%%%%%%%%%%%%%%%%%%%%%
\bibliographystyle{plain}
\bibliography{neurips_2024}
\newpage
\appendix
\onecolumn
\section{Dataset}

\paragraph{Dataset} We evaluate the performance of ZS 3D anomaly detection on three publicly available 3D anomaly detection datasets, MVTec3D-AD, Eyecandies, and Real3D-AD. MVTec3D-AD comprises 4147 point clouds across 10 categories. These objects exhibit diverse object semantics, including bagel, cable gland, carrot, cookie, dowel, foam, peach, potato, rope, and tire. The training dataset comprises 2656 normal point clouds, and the validation dataset comprises 294 normal point clouds. The test dataset includes 948 normal and 249 anomaly point clouds, covering several anomaly types. Point-wise annotations are available for the point clouds. MVTec3D-AD also provides corresponding 2D-RGB image counterparts for the point clouds. We remove the background plane of point clouds in the whole dataset like~\cite{horwitz2023back}. Eyecandies also has 10 different classes and provides the corresponding 2D RGB information. Real3D-AD is a recently available dataset, which contains 12 objects. However, it does not provide the RGB information.  

\begin{figure}[h]
  \centering
    \subfigure[Inference for ZS 3D anomaly detection]{\includegraphics[width=0.49\textwidth]{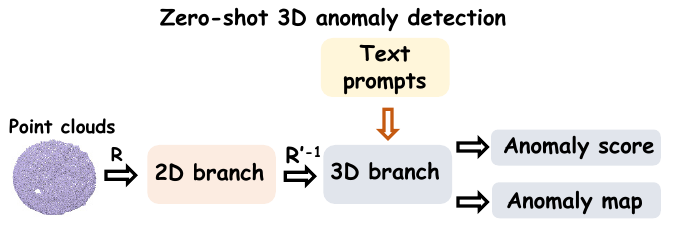}%
    \label{fig: inference_1}}
    \hfil
    \subfigure[Inference for ZS M3D anomaly detection]{\includegraphics[width=0.49\textwidth]{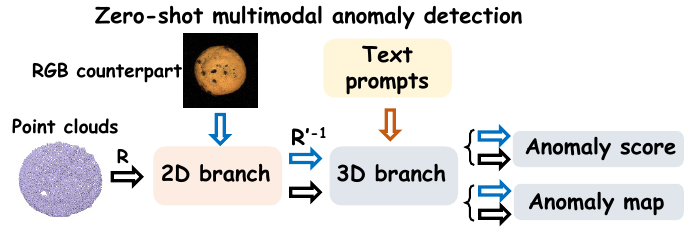}%
    \label{fig: inference_2}}
 \caption{Inference schematic for ZS 3D and M3D anomaly detection.}
\end{figure}

\paragraph{Evaluation Setting and Metric}
\label{Evaluation}
Since the training dataset of MVTec3D-AD only contains all normal samples, \textbf{we use an object test dataset as the auxiliary dataset to fine-tune PointAD and assess the ZS anomaly detection for the remaining objects.} In particular, we report the average results using different objects as the auxiliary, i.e., carrot, cookie, and dowel for MVTec3D-AD;  confetto, LicoriceSandwich, and PeppermintCandy for Eyecandie; seahorse, shell, and starfish for Real3D-AD. Moreover, \textbf{we explore more challenging cross-dataset generalization settings, where we use auxiliary data to test all objects of another dataset.} For point cloud anomaly detection, we only use point clouds to detect and localize 3D anomalies in Figure~\ref{fig: inference_1}. In M3D anomaly detection, both point clouds and their 2D-RGB counterparts are utilized, as shown in Figure~\ref{fig: inference_2}. To comprehensively analyze PointAD, we utilize four metrics to assess its anomaly classification and segmentation performance. For anomaly detection, we use the Area Under the Receiver Operating Characteristic Curve (I-AUROC$\%\uparrow$) and average precision (AP$\%\uparrow$). Regarding anomaly segmentation, we use point-level AUROC (P-AUROC$\%\uparrow$) and a restricted metric called AUPRO$\%(\uparrow)$~\cite{bergmann2020uninformed} to provide a detailed evaluation of subtle anomaly regions.

\section{Implementation Details}
\label{appendix: implementation details}

\begin{table}[h]
\caption{Ablation study on the number of rendering views.}
\label{table5: Ablation study on the number of rendering views.}
\begin{center}
% \begin{small}
% \begin{sc}
\scriptsize
\renewcommand\arraystretch{1.2}
\begin{tabular}{cc}
\toprule
View number & Rendering angles  \\
\midrule
1    & $0$ \\
3    & $-\frac{1}{2}\pi, 0, \frac{1}{2}\pi$\\
5    & $-\frac{2}{3}\pi, -\frac{1}{3}\pi, 0, \frac{1}{3}\pi, \frac{2}{3}\pi$\\
7    & $-\frac{3}{4}\pi, -\frac{1}{2}\pi, -\frac{1}{4}\pi, 0, \frac{1}{4}\pi, \frac{1}{2}\pi, \frac{3}{4}\pi$\\
9    & $-\frac{4}{5}\pi, -\frac{3}{5}\pi, -\frac{2}{5}\pi, -\frac{1}{5}\pi, 0, \frac{1}{5}\pi, \frac{2}{5}\pi, \frac{3}{5}\pi, \frac{4}{5}\pi$\\
11   & $-\frac{5}{6}\pi, -\frac{2}{3}\pi, -\frac{1}{2}\pi, -\frac{1}{2}\pi, \frac{1}{6}\pi, 0, \frac{1}{6}\pi, \frac{1}{3}\pi, \frac{1}{2}\pi, \frac{2}{3}\pi, \frac{5}{6}\pi$\\
\bottomrule
\end{tabular}
% \end{sc}
% \end{small}
\end{center}
\end{table}

Both point clouds and 2D renderings are resized to 336 $\times$ 336. We use Open3d library\footnote{\url{https://github.com/isl-org/Open3D}} to generate 9 views by rotating point clouds along the X-axis at angles of $\{-\frac{4}{5}\pi, -\frac{3}{5}\pi, -\frac{2}{5}\pi, -\frac{1}{5}\pi, 0, \frac{1}{5}\pi, \frac{2}{5}\pi, \frac{3}{5}\pi, \frac{4}{5}\pi \}$ for most categories. Some categories lose their surface completely because they are not stereo point clouds. Table~\ref{table5: Ablation study on the number of rendering views.} also gives the specific rendering angles of different view number settings. We set the rendering angles in a circular manner, evenly distributing the angles between $-\pi$ to $\pi$. The backbone of PointAD is the pre-trained CLIP model\footnote{\url{https://github.com/mlfoundations/open\_clip}} (\texttt{VIT-L/14@336px}). Following~\cite{zhou2023anomalyclip}, we improve the local visual semantics of the vision encoder of CLIP without modifying its parameters and introduce learnable tokens in the text encoder. During training, we keep all CLIP parameters frozen and set the learnable word embeddings in object-agnostic text templates to $12$. We use the Adam optimizer with a learning rate of $0.001$ to optimize the learnable parameters. The experiment runs for $15$ epochs with a batch size of $4$. All experiments were conducted on a single NVIDIA RTX 3090 24GB GPU using PyTorch-2.0.0. 

\section{Baselines}
\label{appendix: baselines}
ZS 3D anomaly detection and M3D anomaly detection have not yet been explored. We first make an adaption for the original CLIP for 3D anomaly detection. Then, we reproduce SOTA ZS 3D classification methods (\ie, PointCLIP V2~\cite{zhu2023pointclip} and Cheraghian~\cite{cheraghian2022zero}) and adapt them to our settings. SOTA unsupervised 3D anomaly detection approaches are reported as the performance upper bound. All hyperparameters in these baselines are kept the same. We will present the detailed reproduction as follows:
\begin{itemize}
    \item CLIP + Rendering is a method, where we apply the original CLIP into our framework for ZS 3D anomaly detection. It uses the same rendering procedure as PointAD. Following~\cite{jeong2023winclip, zhou2023anomalyclip}, we integrate anomaly semantics into CLIP by two class text prompt templates: \verb'A photo of a normal [cls]' and \verb'A photo of an anomalous [cls]', where \verb'cls' denotes the target class name.
    \item PointCLIP V2 (CVPR 2023) is a SOTA ZS 3D classification method based on CLIP, they project point clouds into depth maps from different views. To adapt PointCLIP V2 into ZS anomaly detection, we replace its original text prompts~\texttt{point cloud of a big [c]} with normal text prompts~\texttt{point cloud of a big [c]} and abnormal text prompts~\texttt{point cloud of a big damaged [c]}. 
    \item AnomalyCLIP (ICLR 2024) is a SOTA zero-shot 2D anomaly detection method. AnomalyCLIP introduces object-agnostic learning to capture generic anomaly semantics of images. We adapt AnomalyCLIP in 3D detection by fine-tuning AnomalyCLIP on depth images of corresponding point clouds.
    \item Cheraghian (IJCV 2022) is an approach for ZS 3D anomaly detection without foundation models. They directly extract the point presentations by PointNet and use word2vector~\cite{mikolov2013efficient} to generate the textual embedding of an object. To incorporate the anomaly semantics into Cheraghian, we respectively average the textual embedding of an object name and \texttt{damaged}. We replace the global representation with dense representations to provide the segmentation results.
\end{itemize}

\section{Related Work}
\subsection{2D Anomaly Detection}
2D  anomaly detection has been studied extensively by leveraging RGB information~\citep{Support,Deep,cohen2020sub, tian2021constrained, PANDA,chen2022deep, zhou2022pull,reiss2023mean}. 
Related works can be categorized into two branches: end-to-end and memory-based methods. Representative end-to-end methods exploit knowledge distillation~\citep{bergmann2020uninformed,zhou2022pull,rudolph2023asymmetric} and normalizing flow~\citep{Cflow,yu2021fastflow} to model the normal distribution. Instead, memory-based methods~\citep{roth2022towards, xie2023pushing} store normal features to construct normal prototypes. ZS 2D anomaly detection is proposed to target a challenging problem where training samples are inaccesible~\cite{esmaeilpour2022zero, li2023zero, liznerski2022exposing, aota2023zero, deng2023anovl, chen2023zero}.
WinCLIP~\cite{jeong2023winclip} attempts to explore ZS 2D anomaly detection using CLIP. AnomalyCLIP~\cite{zhou2023anomalyclip} first introduces object-agnostic prompt learning to capture generic normality and abnormality, detecting anomalies across datasets. PromptAD~\cite{Li2024PromptADZA} focuses on effectively fusing these embeddings to enhance zero-shot detection performance.

\begin{figure}[h!]
    \centering
	\begin{minipage}{0.4\textwidth}
    \centering
    \includegraphics[width=1\textwidth]{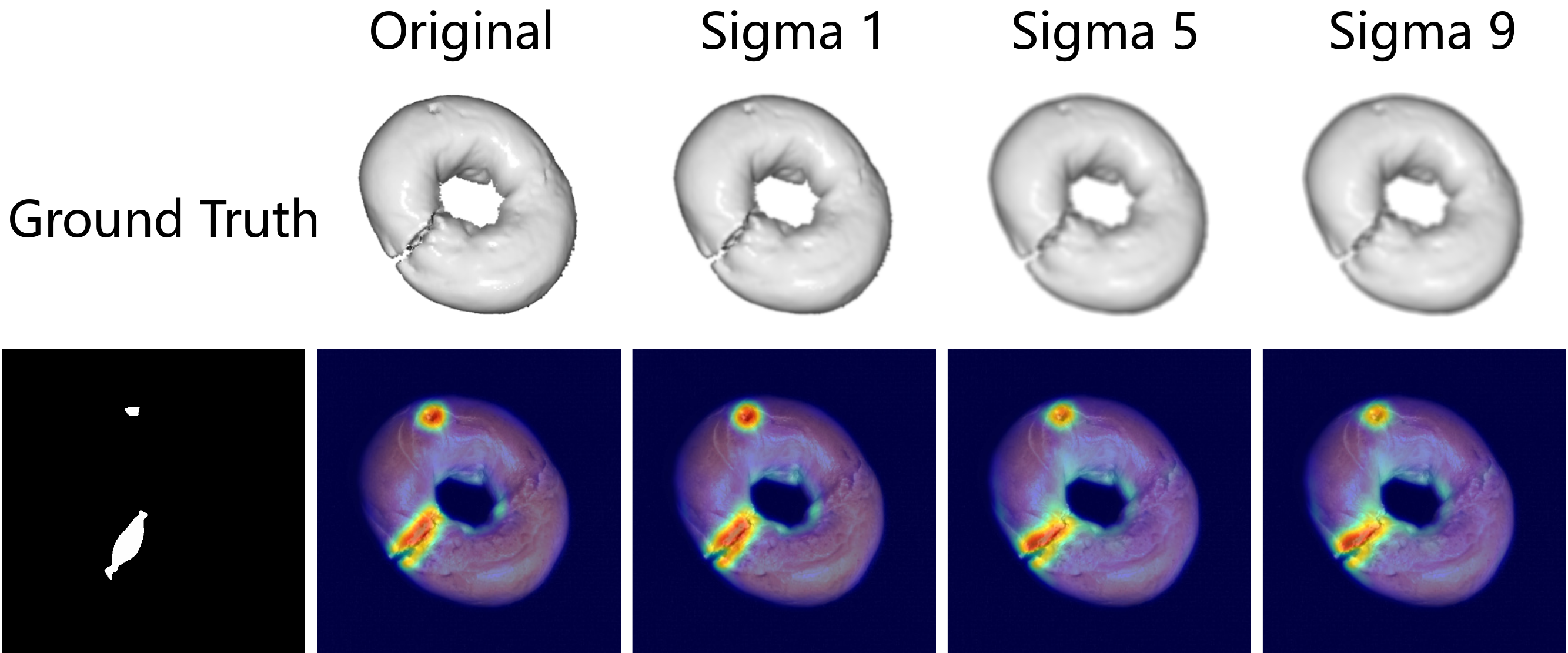}
		\caption{Viusalization with different rendering quality. A larger $\sigma$ represents poorer rendering quality.}
		\label{fig: Viusalization with different rendering quality}
	\end{minipage}
	%\qquad
	\hfil
	\begin{minipage}{0.49\textwidth}
    \centering
    \includegraphics[width=1\textwidth]{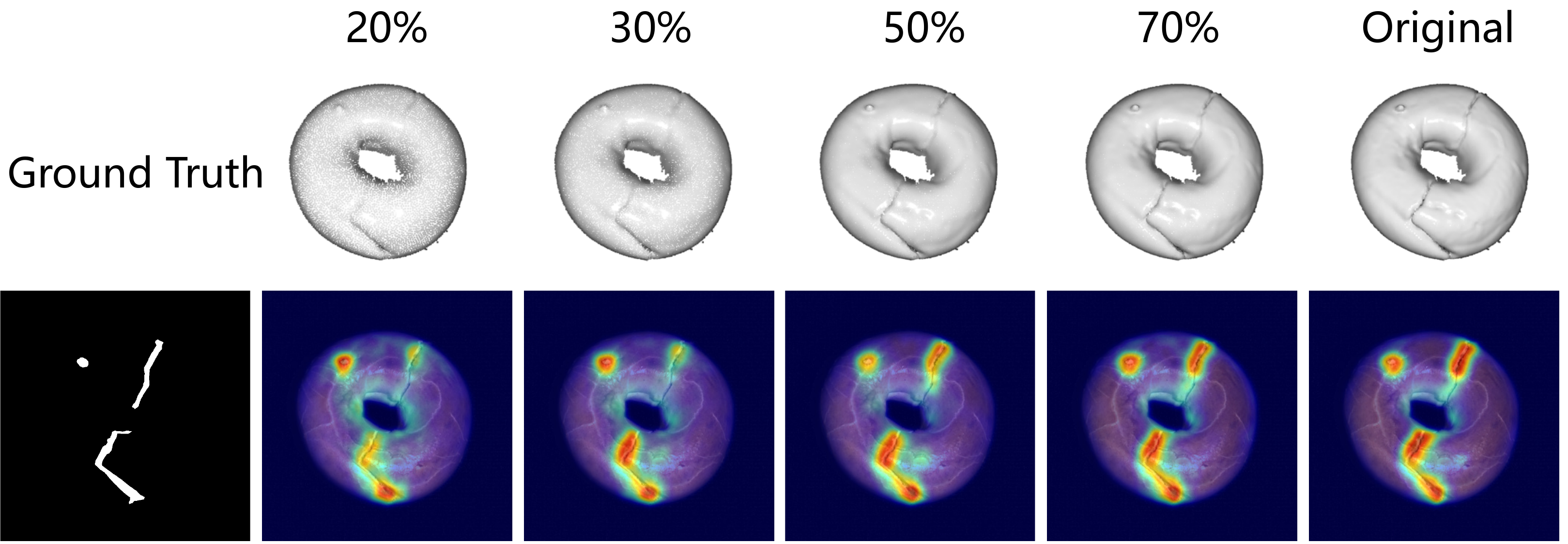}
		\caption{Visualization with different resolutions. We downsample entire point clouds with different ratios to obtain diverse resolutions.}
		\label{fig: Visualization with different resolutions}
	\end{minipage}
\vspace{-0.5em}
\end{figure}

\begin{figure*}[h]
  \begin{minipage}{0.48\textwidth}
        \centering
\vspace{-1em}
 \captionof{table}{Analysis on the rendering quality. The original setting is highlighted in gray.}
 \vspace{-0.5em}
 \label{table: Ablation on the rendering quality.}
 \tiny
    \setlength\tabcolsep{3pt} 
    \begin{tabular}{c|cccccc}
    \toprule
    \multirow{1}{*}{Blur} & \multicolumn{2}{c}{Point detection} & \multicolumn{2}{c}{Multimodal detection} \\ sigma & Global  & Local & Global  & Local  \\ \hline
    \rowcolor{gray!40}
     0 &  (\textcolor{red}{82.0}, \textcolor{red}{94.2}) & (\textcolor{red}{95.5}, \textcolor{red}{84.4}) & (\textcolor{red}{86.9}, \textcolor{red}{96.1}) & (\textcolor{red}{97.2}, \textcolor{red}{90.2})\\
     1 & (80.1, 93.5) & (95.2, 83.2) & (85.5, 95.6) & (97.0, 90.0)\\
    5 & (78.2, 92.5) & (95.1, 82.3) & (83.6, 95.1) & (96.8, 89.7)\\
    9 & (77.6, 92.2) & (95.1, 82.3) & (83.2, 95.0) & (96.5, 89.3)\\
    \bottomrule
    \end{tabular}
    \end{minipage} 
\begin{minipage}{0.48\textwidth}
    \centering
    \captionof{table}{Analysis on the input solution.}%这里必须写table，不然标题就自动设置成figure
        \vspace{-0.5em}
    \tiny
 \label{table: Ablation on the input solution.}
    \setlength\tabcolsep{3pt} 
    \begin{tabular}{c|ccccc}
    \toprule
    \multirow{1}{*}{Downsample}  & \multicolumn{2}{c}{Point detection} & \multicolumn{2}{c}{Multimodal detection} \\ 
    ratio & Global  & Local & Global  & Local  \\ \hline
    \rowcolor{gray!40}
     100\% &  (\textcolor{red}{82.0}, \textcolor{red}{94.2}) & (\textcolor{red}{95.5}, \textcolor{red}{84.4}) & (\textcolor{red}{86.9}, \textcolor{red}{96.1}) & (\textcolor{red}{97.2}, \textcolor{red}{90.2})\\
     70\% & (80.6, 93.3) & (95.2, 82.6) & (86.2, 95.1) & (97.0, 90.0)\\  
    50\% & (77.9, 92.6) & (95.0, 82.0) & (83.6, 94.9) & (96.8, 89.2)\\
    30\% & (74.6, 91.1) & (94.9, 81.4) & (81.9, 94.6) & (96.2, 88.7)\\
    20\% & (73.9, 90.9) & (94.3, 78.7) & (80.3, 93.4) & (95.3, 87.5)\\
    \bottomrule
    \end{tabular}%
	\end{minipage}
\end{figure*}
\vspace{-1em}

\section{Analysis on Rendering Conditions}
\paragraph{Rendering quality} PointAD interprets point clouds through their corresponding 2D renderings, and the quality of these renderings impacts the information that PointAD can extract from the original point clouds. In our manuscript, we used the Open3D Library to render the point clouds, but it does not provide an API for controlling rendering quality. To simulate varying rendering quality, we applied Gaussian blur with different extents $\sigma$ to the 2D renderings. Sample visualizations are included in Figure~\ref{fig: Viusalization with different rendering quality}. Specifically, we conducted experiments on MVTec3D-AD using different blur $\sigma$ values (i.e., 
). Table~\ref{table: Ablation on the rendering quality.} shows that the detection performance of PointAD diminishes as rendering quality decreases (with increasing sigma). However, the degradation is acceptable even when the renderings are heavily blurred ($\sigma$ equals 9). In such cases, PointAD still outperforms baselines that use high-quality renderings.

\paragraph{Input Resolution}
Here, we study the effect of resolutions of input point clouds.  To create low-resolution point clouds, we downsample the entire high-resolution point clouds using Farthest Point Sampling (FPS) with various sampling ratios. This strategy allows us to generate corresponding low-resolution datasets for training and evaluating PointAD. Visualizations of these datasets are provided in Figure~\ref{fig: Visualization with different resolutions}. We train PointAD using the resulting low-resolution samples and test PointAD on the same resolution. Table~\ref{table: Ablation on the input solution.} demonstrates that PointAD maintains a strong detection capacity for low-resolution point clouds when the downsampling ratios are 20\%, 30\%, 50\%, and 70\%. Even at 20\% resolution, PointAD still achieves state-of-the-art performance. This indicates that PointAD is generally applicable to point clouds with various resolutions.

\paragraph{Rendering angles and different numbers of views}

PointAD interprets point clouds from their 2D renderings, where rendering angles and numbers collectively determine the amount of information derived. They have different emphases. For rendering angles, the importance lies in the discrepancy between adjacent angles, as this affects the information granularity that PointAD obtains from adjacent views. When the angle discrepancy is fixed, the number of renderings determines the coverage of 3D information in the resulting 2D renderings. To capture all point cloud information, especially abnormal points, it is crucial to ensure comprehensive coverage. Therefore, our approach in selecting rendering angles and the number of renderings is to guarantee that all points in point clouds are adequately represented.

Based on this principle, we conducted experiments to assess the impact of the number of renderings on PointAD's detection performance, circularly rendering point clouds to ensure even coverage of all points. As shown in Table~\ref{table: Ablation on rendering number.}, increasing the number of views allows PointAD to gather more detailed information from the 2D renderings, benefiting from smaller angle discrepancies, which improves detection and localization results. However, when the number of views increased from 9 to 11, we observed a performance decline in PointAD, with the I-AUROC for global multimodal detection dropping from 87.4\% to 86.4\%. This suggests that incorporating too many views could introduce redundant information, resulting in 2D renderings with extensive overlap and excessive local detail. This overemphasis on local information can impede global recognition. Hence, the appropriate number of views benefits point understanding from informative views while mitigating the adverse effects of redundant local information. To further explore the impact of the absolute angle, we shift the rendering angles while keeping the angle discrepancy unchanged. The original adjacent angle discrepancy in our paper is $\frac{1}{5}\pi$. We divide this discrepancy into four parts and perform angle shifts of $\frac{1}{20}\pi$, $\frac{2}{20}\pi$, and $\frac{3}{20}\pi$ to test the impact of varying rendering angles. Table~\ref{table Analysis on the rendering angle.} shows that PointAD maintains consistent performance across different rendering angles, demonstrating its robustness to variations in angles different from those used during training.

\begin{figure*}[h!]
  \begin{minipage}{0.48\textwidth}
        \centering
 \captionof{table}{Analysis on the rendering angle.}
 \vspace{-0.5em}
 \label{table Analysis on the rendering angle.}
 \tiny
    \setlength\tabcolsep{3pt} 
    \begin{tabular}{c|ccccc}
    \toprule
    \multirow{1}{*}{Angle} & \multicolumn{2}{c}{Point detection} & \multicolumn{2}{c}{Multimodal detection} \\ 
    shift & Global  & Local & Global  & Local  \\ \hline 
    \rowcolor{gray!40}
    0 & (\textcolor{red}{82.0}, \textcolor{red}{94.2}) & (\textcolor{red}{95.5}, \textcolor{red}{84.4}) & (\textcolor{red}{86.9}, \textcolor{red}{96.1}) & (\textcolor{red}{97.2}, \textcolor{red}{90.2})\\  
    $\frac{1}{15} \pi$ & (82.6, 94.6) & (95.4, 83.9) & (86.7, 96.0) & (97.1, 90.7)\\  
    $\frac{2}{15} \pi$ & (82.1, 94.4) & (95.4, 84.1) & (86.4, 95.9) & (97.1, 90.7)\\
    $\frac{3}{15} \pi$ & (82.6, 94.6) & (95.4, 84.3) & (86.4, 95.9) & (97.1, 90.7)\\
    \bottomrule
    \end{tabular}
    \end{minipage} 
    \hfill
    \hfill
\begin{minipage}{0.48\textwidth}
    \centering
    \captionof{table}{Analysis on the rendering lighting.}%这里必须写table，不然标题就自动设置成figure
        \vspace{-0.5em}
    \tiny
 \label{tabel Visualization with different rendering lighting.}
    \setlength\tabcolsep{3pt} 
    \begin{tabular}{c|ccccc}
    \toprule
    % \multicolumn{5}{c|}{Proposed technology}& \multicolumn{4}{c}{MVTEC}\\ \hline
    \multirow{2}{*}{Lighting}  & \multicolumn{2}{c}{Point detection} & \multicolumn{2}{c}{Multimodal detection} \\ 
    & Global  & Local & Global  & Local  \\ \hline
     ++ &  (82.0, 94.3) & (95.4, 83.7) & (85.7, 95.8) & (97.2, 90.7)\\
     + & (82.4, 94.6) & (95.4, 83.8) & (86.1, 95.9) & (97.1, 90.5)\\ 
    \rowcolor{gray!40}
    original  & (\textcolor{red}{82.0}, \textcolor{red}{94.2}) & (\textcolor{red}{95.5}, \textcolor{red}{84.4}) & (\textcolor{red}{86.9}, \textcolor{red}{96.1}) & (\textcolor{red}{97.2}, \textcolor{red}{90.2})\\  
    - &  (82.4, 94.5) & (95.3, 83.9) & (86.4, 95.9) & (97.1, 90.6)\\
    -- &  (81.9, 94.3) & (95.3, 83.4) & (86.1, 95.8) & (97.1, 90.5)\\
    \bottomrule
    \end{tabular}
	\end{minipage}
\end{figure*}

\begin{table}[]
    \centering
    \captionof{table}{Analysis on the point occlusions.}%这里必须写table，不然标题就自动设置成figure
    \small
 \label{table Analysis on the point occlusions.}
    \setlength\tabcolsep{3pt} 
    \begin{tabular}{c|ccccc}
    \toprule
    \multirow{2}{*}{Method} & \multicolumn{2}{c}{Point detection} & \multicolumn{2}{c}{Multimodal detection} \\ 
    & Global  & Local & Global  & Local  \\ \hline
    \rowcolor{gray!40}
     original  & (\textcolor{red}{82.0}, \textcolor{red}{94.2}) & (\textcolor{red}{95.5}, \textcolor{red}{84.4}) & (\textcolor{red}{86.9}, \textcolor{red}{96.1}) & (\textcolor{red}{97.2}, \textcolor{red}{90.2})\\ 
     occlusions & (73.3, 90.6) & (94.3, 80.8) & (83.0, 94.8) & (96.7, 89.5)\\
    \bottomrule
    \end{tabular}
\end{table}

\begin{figure}[h!]
    \centering
	\begin{minipage}{0.65\textwidth}
    \centering
    \includegraphics[width=1\textwidth]{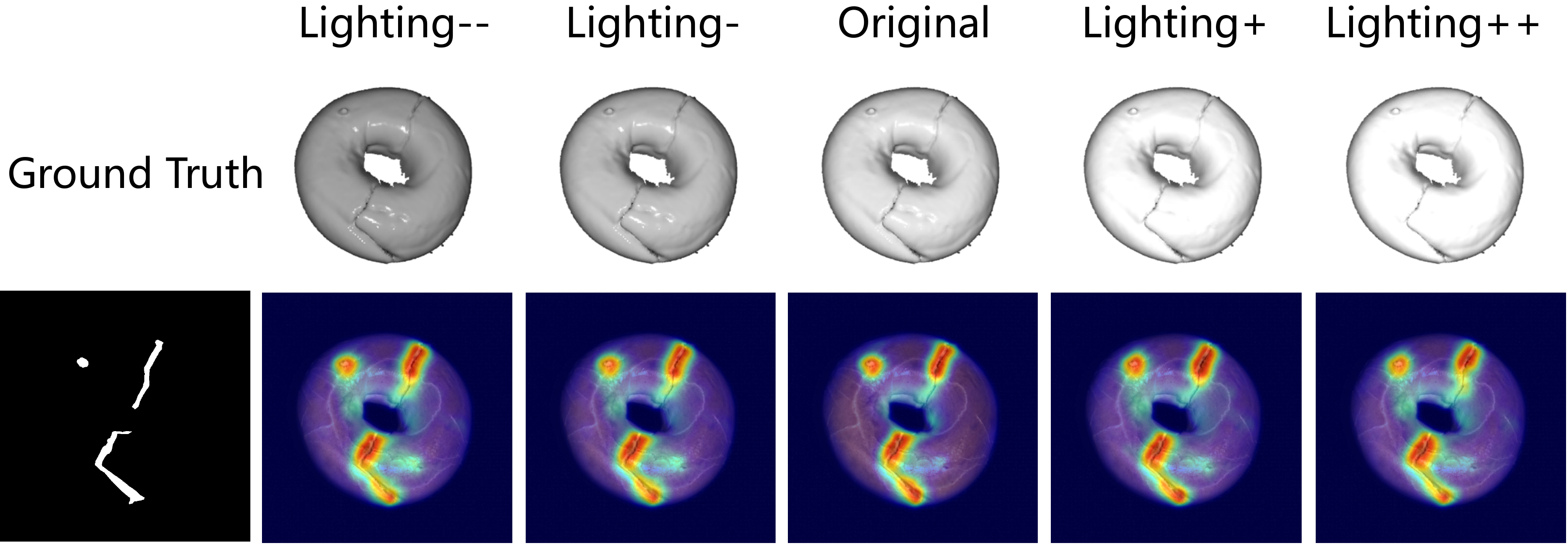}
		\caption{Visualization with different rendering lighting.}
		\label{fig: Visualization with different rendering lighting.}
	\end{minipage}
	%\qquad
	\hfill
	\begin{minipage}{0.30\textwidth}
    \centering
    \includegraphics[width=1\textwidth]{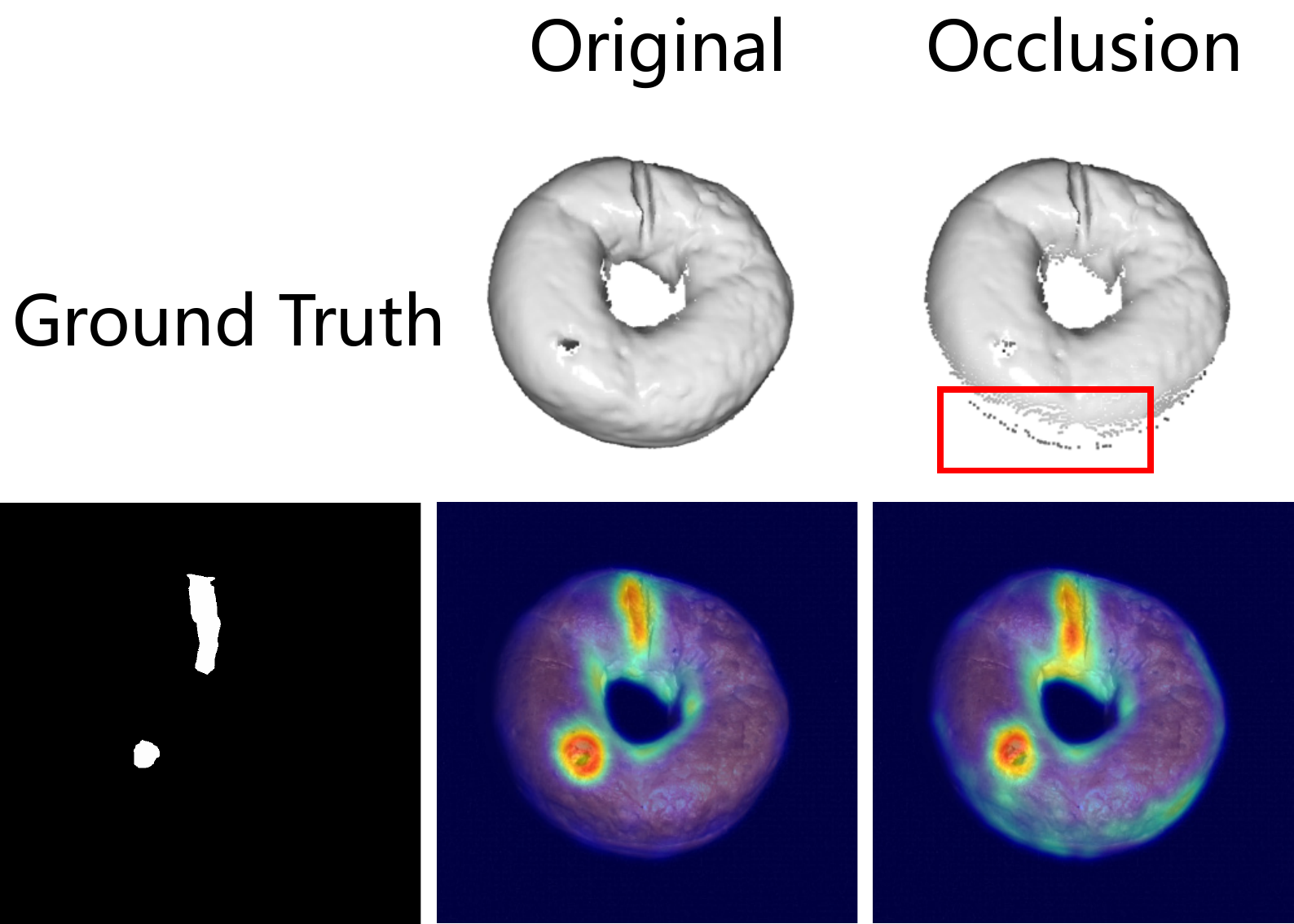}
		\caption{Visualization of occluded point clouds.}
		\label{fig: Visualization of occluded point clouds.}
	\end{minipage}
 \hfill
\end{figure}

\paragraph{Rendering lighting}
Further exploring the robustness of PointAD under different conditions could enhance its generalized detection performance. We conducted ablation studies to test its sensitivity under different lighting conditions and with occluded point clouds below. To evaluate the impact of rendering lighting, we adjusted the lighting conditions to render point clouds, generating variant datasets with different lighting. We used both stronger and weaker lighting to render point clouds compared to the original dataset, covering a broad lighting range. We denote stronger and the strongest lighting as "+" and "++", and weaker and the weakest lighting as "-" and "--". Visualizations of the resulting samples are presented in Figure~\ref{fig: Visualization with different rendering lighting.}. The experiments were conducted on MVTec3D-AD, where we tested PointAD, trained on the original dataset, on these lighting variant datasets. Table~\ref{tabel Visualization with different rendering lighting.} shows that PointAD can still detect anomalies even with significant discrepancies in rendering lighting, suggesting that PointAD is not sensitive to variations in rendering lighting.

\paragraph{Point occlusions}
Next, we evaluated detection performance with occluded point clouds. We occluded point clouds by removing those invisible from a specific rendering angle and then used the same rendering parameters to project the remaining points. During this process, we observed that abnormal regions might be occluded totally. Unlike class classification, where class semantics remain unchanged by point occlusions, removing these anomaly semantics would transform anomaly samples into normal points. Therefore, we selected rendering angles that allow visibility of part or all anomalous regions when the point cloud is an abnormal instance. The occluded point clouds are shown in Figure~\ref{fig: Visualization of occluded point clouds.}. We then used PointAD, trained on the original dataset, to test the resulting occluded dataset. Table~\ref{table Analysis on the point occlusions.} shows that PointAD suffers from performance degradation when the point clouds are occluded. We attribute this to two aspects: 1) Despite this strategy, the occluded point clouds could still lose part of the anomaly semantics; 2) Occluded point clouds could create unexpected sinkholes on the surface, which may cause PointAD to identify these areas as hole anomalies incorrectly.

\section{Hyperparameter Ablation}
\label{appendix: hyperparameter}
\paragraph{Length of learnable prompts}
% \paragraph{Backbone Ablation}

\begin{table}[h]
\caption{Ablation study on the length of the learnable prompt.}
\label{length ablation}
\begin{center}
% \begin{small}
% \begin{sc}
\scriptsize
\begin{tabular}{ccccc}
\toprule
\multirow{2}{*}{\makecell[c]{Length of \\ learnable prompt}} & \multicolumn{2}{c}{Point detection} & \multicolumn{2}{c}{Multimodal detection} \\
 &Pixel level & Image level & Pixel level & Image level \\
\midrule
6    & (94.6, 83.4)& (81.7, \textcolor{blue}{94.2})  & (96.5, 89.8)& (\textcolor{blue}{86.6}, \textcolor{blue}{96.0})\\
8    & (95.2, 83.6)& (\textcolor{red}{82.0}, \textcolor{blue}{94.2})  & (96.8, 90.0)& (\textcolor{blue}{86.6}, 95.8)\\
10    & (95.3, \textcolor{blue}{84.0})& (\textcolor{blue}{81.8}, \textcolor{red}{94.3})  & (\textcolor{blue}{97.0}, \textcolor{blue}{90.1})& (86.5, \textcolor{blue}{96.0})\\
12    & (\textcolor{red}{95.5}, \textcolor{red}{84.4})& (\textcolor{red}{82.0}, \textcolor{blue}{94.2})  & (\textcolor{red}{97.2}, \textcolor{red}{90.2})& (\textcolor{red}{86.9}, \textcolor{red}{96.1})\\
14   & (\textcolor{blue}{95.4}, 83.7)& (81.4, 94.1)  & (96.9, 89.7)& (85.5, 95.6)\\
16   & (95.1, 83.0)& (81.5, 94.1)  & (96.9, 89.8)& (84.7, 95.6)\\
\bottomrule
\end{tabular}
% \end{sc}
% \end{small}
\end{center}
% \vskip -0.1in
\end{table}

We study the sensitivity of important hyperparameters in PointAD. First, we explore the length of learnable text prompt templates, as shown in Table.~\ref{length ablation}. As the length of word embeddings increases, PointAD can better learn 3D and 2D anomaly semantics to improve its performance. Nonetheless, with a further increase in length (i.e., from 12 to 16), a decline in performance becomes noticeable. The excessive or insufficient number of learnable word embeddings can lead to performance degradation. An appropriate length (\ie, 12) is important for PointAD to attain comprehensive performance in both 3D and M3D anomaly detection.

\paragraph{Training set}
We have increased the test data for each category to incorporate more instances on MVTec3D-AD. Originally, we evaluated the rest of the data using only one category as the test data, such as carrot, cookie, and dowel. Now, we attempt to incorporate more instances. Here, we utilized two categories as auxiliary data, including carrot and cookies, carrot and dowel, and cookie and dowel. To further analyze the effect of the size of auxiliary data, we utilized three categories as auxiliary data, selecting all possible combinations from the four sets: carrot, cookie, bagel, and dowel. We present the performance averaged across all groups below. From Table~\ref{table:3D Training set size} and Table~\ref{table: multimodal Training set size}, PointAD can incorporate more knowledge about abnormality and normality, improving point detection and multimodal detection performance. Specifically, from one category to three categories, PointAD exhibits improved performance, with P-AUROC increasing from 95.5\%, 96.1\%, to 96.3\%, and AUPRO increasing from 84.4\%, 86.3\%, to 86.5\%. Moreover, I-AUROC increases from 97.2\%, 97.5\%, to 97.8\%, and AP increases from 90.2\%, 91.8\%, to 92.0\%. This trend is also observed in global anomaly semantics.
\begin{table*}[h]
\centering
\caption{Ablation study on training set size for point detection}
\label{table:3D Training set size}
\tiny
\setlength\tabcolsep{1pt}
\begin{tabular}{cccccccccccc|c}
\toprule
& Training set & Bagel & Cable gland & Carrot & Cookie & Dowel & Foam & Peach & Potato & Rope & Tire & Mean \\
\midrule
\multirow{3}{*}{\makecell[c]{G.}}
& one & (98.3, 99.6) & (\textcolor{red}{53.7}, \textcolor{red}{86.0}) & (97.9, \textcolor{blue}{99.6}) & (\textcolor{blue}{92.1}, \textcolor{blue}{97.9}) & (\textcolor{red}{72.2}, \textcolor{red}{92.0}) & (\textcolor{red}{69.5}, \textcolor{red}{91.2}) & (\textcolor{blue}{91.5}, \textcolor{blue}{97.6}) & (98.8, 99.7) & (\textcolor{red}{91.5}, \textcolor{red}{96.7}) & (54.1, 82.1) & (\textcolor{blue}{82.0}, 94.2) \\
& two & (\textcolor{blue}{99.0}, \textcolor{blue}{99.8}) & (\textcolor{blue}{52.5}, \textcolor{blue}{85.3}) & (\textcolor{blue}{98.3}, \textcolor{red}{99.7}) & (91.8, 97.7) & (70.2, 91.4) & (\textcolor{blue}{68.9}, 90.8) & (\textcolor{blue}{91.5}, \textcolor{blue}{97.6}) & (\textcolor{blue}{99.2}, \textcolor{blue}{99.8}) & (\textcolor{blue}{90.8}, \textcolor{blue}{96.5}) & (\textcolor{red}{56.9}, \textcolor{red}{85.0}) & (81.9, \textcolor{red}{94.4}) \\
& three & (\textcolor{red}{100}, \textcolor{red}{100}) & (52.2, 84.8) & (\textcolor{red}{98.7}, \textcolor{red}{99.7}) & (\textcolor{red}{94.2}, \textcolor{red}{98.4}) & (\textcolor{blue}{71.2}, \textcolor{blue}{91.6}) & (\textcolor{blue}{68.9}, \textcolor{blue}{91.1}) & (\textcolor{red}{92.6}, \textcolor{red}{98.0}) & (\textcolor{red}{99.6}, \textcolor{red}{99.9}) & (88.9, 95.8) & (\textcolor{blue}{54.5}, \textcolor{blue}{83.6}) & (\textcolor{red}{82.1}, \textcolor{blue}{94.3}) \\
\midrule
\multirow{3}{*}{\makecell[c]{L.}}
& one & (98.4, 96.9) & (\textcolor{red}{93.5}, 79.5) & (\textcolor{blue}{99.4}, 96.4) & (87.5, 75.4) & (95.5, \textcolor{blue}{75.2}) & (86.5, 54.1) & (99.5, 98.3) & (\textcolor{red}{99.9}, \textcolor{blue}{99.1}) & (99.3, 89.9) & (95.3, 79.7) & (95.5, 84.4) \\
& two & (\textcolor{blue}{99.1}, \textcolor{blue}{98.0}) & (\textcolor{red}{93.5}, \textcolor{red}{79.9}) & (\textcolor{red}{99.6}, \textcolor{blue}{97.7}) & (\textcolor{blue}{90.4}, \textcolor{blue}{83.8}) & (\textcolor{blue}{95.6}, \textcolor{blue}{75.2}) & (\textcolor{red}{88.3}, \textcolor{red}{57.3}) & (\textcolor{blue}{99.6}, \textcolor{blue}{98.6}) & (\textcolor{red}{99.9}, \textcolor{red}{99.4}) & (\textcolor{red}{99.5}, \textcolor{red}{91.0}) & (\textcolor{red}{96.0}, \textcolor{red}{81.7}) & (\textcolor{blue}{96.1}, \textcolor{blue}{86.3}) \\
& three & (\textcolor{red}{99.3}, \textcolor{red}{98.2}) & (\textcolor{red}{93.5}, \textcolor{blue}{79.8}) & (\textcolor{red}{99.6}, \textcolor{red}{97.8}) & (\textcolor{red}{92.4}, \textcolor{red}{87.7}) & (\textcolor{red}{95.7}, \textcolor{red}{75.7}) & (\textcolor{blue}{87.9}, \textcolor{blue}{57.1}) & (\textcolor{red}{99.7}, \textcolor{red}{98.7}) & (\textcolor{red}{99.9}, \textcolor{red}{99.4}) & (\textcolor{blue}{99.4}, \textcolor{blue}{90.6}) & (\textcolor{blue}{95.7}, \textcolor{blue}{80.3}) & (\textcolor{red}{96.3}, \textcolor{red}{86.5}) \\
\bottomrule
\end{tabular}
\vspace{-1.5em}
\end{table*}

\begin{table*}[h]
\centering
\caption{Ablation study on training set size for multimodal detection.}
\label{table: multimodal Training set size}
\tiny
\setlength\tabcolsep{1pt}
\begin{tabular}{cccccccccccc|c}
\toprule
& Training set & Bagel & Cable gland & Carrot & Cookie & Dowel & Foam & Peach & Potato & Rope & Tire & Mean \\
\midrule
\multirow{3}{*}{\makecell[c]{MG.}}
& one & (\textcolor{red}{98.8}, \textcolor{red}{99.7}) & (79.9, \textcolor{blue}{94.7}) & (95.5, 98.9) & (\textcolor{blue}{86.2}, \textcolor{blue}{95.5}) & (\textcolor{red}{98.5}, \textcolor{red}{90.6}) & (84.4, 96.1) & (\textcolor{blue}{96.6}, \textcolor{blue}{99.1}) & (\textcolor{blue}{90.7}, \textcolor{blue}{97.0}) & (93.6, 97.3) & (\textcolor{blue}{74.6}, 92.0) & (86.9, \textcolor{blue}{96.1}) \\
& two & (\textcolor{blue}{98.6}, \textcolor{red}{99.7}) & (\textcolor{red}{80.8}, \textcolor{red}{94.8}) & (\textcolor{red}{98.3}, \textcolor{red}{99.6}) & (85.6, 95.4) & (64.7, 88.9) & (\textcolor{blue}{84.9}, \textcolor{blue}{96.3}) & (95.8, 99.0) & (90.6, 96.8) & (\textcolor{blue}{94.5}, \textcolor{blue}{97.8}) & (\textcolor{red}{77.2}, \textcolor{red}{93.0}) & (\textcolor{blue}{87.1}, \textcolor{blue}{96.1}) \\
& three & (\textcolor{blue}{98.6}, \textcolor{red}{99.7}) & (\textcolor{blue}{80.1}, 94.5) & (\textcolor{blue}{97.3}, \textcolor{blue}{99.4}) & (\textcolor{red}{91.7}, \textcolor{red}{97.7}) & (\textcolor{blue}{66.8}, \textcolor{blue}{89.5}) & (\textcolor{red}{87.3}, \textcolor{red}{97.0}) & (\textcolor{red}{97.4}, \textcolor{red}{99.4}) & (\textcolor{red}{92.0}, \textcolor{red}{97.4}) & (\textcolor{red}{95.0}, \textcolor{red}{98.0}) & (74.0, \textcolor{blue}{92.3}) & (\textcolor{red}{88.0}, \textcolor{red}{96.5}) \\
\midrule
\multirow{3}{*}{\makecell[c]{ML.}}
& one & (\textcolor{blue}{99.6}, 90.1) & (96.7, \textcolor{red}{97.9}) & (\textcolor{blue}{99.4}, 85.5) & (92.6, 85.4) & (\textcolor{red}{96.1}, 74.0) & (92.4, \textcolor{red}{98.3}) & (99.4, \textcolor{red}{98.9}) & (\textcolor{red}{99.8}, 92.9) & (\textcolor{red}{98.8}, 87.9) & (\textcolor{blue}{97.5}, \textcolor{red}{91.1}) & (97.2, 90.2) \\
& two & (\textcolor{blue}{99.6}, \textcolor{blue}{98.7}) & (\textcolor{blue}{97.0}, \textcolor{blue}{91.6}) & (\textcolor{blue}{99.4}, \textcolor{blue}{97.6}) & (\textcolor{blue}{93.8}, \textcolor{blue}{88.4}) & (95.7, \textcolor{blue}{83.5}) & (\textcolor{red}{93.9}, \textcolor{blue}{78.8}) & (\textcolor{blue}{99.5}, 98.5) & (\textcolor{red}{99.8}, \textcolor{red}{99.3}) & (\textcolor{blue}{98.6}, \textcolor{blue}{90.6}) & (\textcolor{red}{98.1}, \textcolor{blue}{90.7}) & (\textcolor{blue}{97.5}, \textcolor{blue}{91.8}) \\
& three & (\textcolor{red}{99.7}, \textcolor{red}{98.8}) & (\textcolor{red}{97.2}, 91.1) & (\textcolor{red}{99.5}, \textcolor{red}{98.0}) & (\textcolor{red}{95.4}, \textcolor{red}{90.0}) & (\textcolor{blue}{95.9}, \textcolor{red}{86.1}) & (\textcolor{blue}{93.8}, 77.1) & (\textcolor{red}{99.6}, \textcolor{blue}{98.6}) & (\textcolor{red}{99.8}, \textcolor{blue}{99.2}) & (\textcolor{blue}{98.6}, \textcolor{red}{91.9}) & (\textcolor{red}{98.1}, 89.0) & (\textcolor{red}{97.8}, \textcolor{red}{92.0}) \\
\bottomrule
\end{tabular}
\end{table*}

\begin{figure}[h]
\vspace{-0.5em}
\begin{center}
\centerline{\includegraphics[width = 0.4\columnwidth]{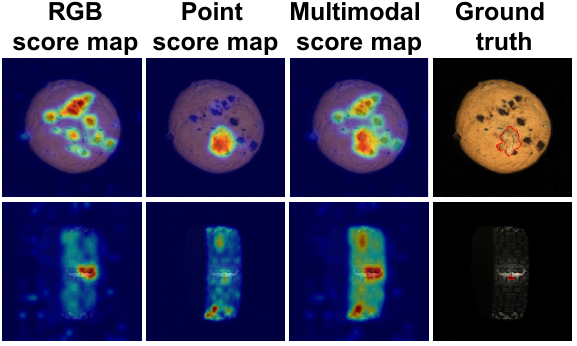}}
\caption{Failure case in PointAD.}
\label{f8:direct multimodality fusion}
\vspace{-0.5em}
\end{center}
% \vskip -0.2in
\end{figure}
\section{Failure Cases}
\label{Failure Cases}
\begin{figure*}[h!]
\centering
 \includegraphics[width=1\textwidth]{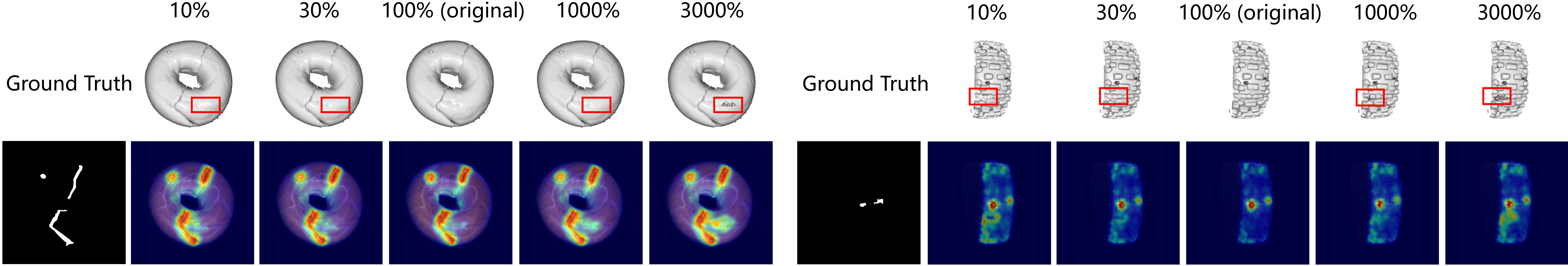}
    \caption{The impact of noise level. We randomly downsample and upsample part of normal regions to create different point densities.}
    \label{figure: The impact of noise level}
\end{figure*}

In this section, we present the failure cases of PointAD, which we attribute to direct multimodality fusion. Since our model uses hybrid loss to incorporate the 3D and 2D anomaly semantics, it performs ZS multimodality 3D anomaly detection in a plug-and-play manner. However, when one modality prediction deviates severely from the ground truth in rare instances, direct fusion may result in an unpromising multimodal score map. As shown in Figure~\ref{f8:direct multimodality fusion}, the hole in the cookie is visually similar to the chocolate on cookies, making it challenging to differentiate the hole anomaly via color information alone. Although PointAD can detect the hole based on its abnormal point relations, the RGB score map heavily influences the final multimodal score map. Conversely, the tire presents an inverse situation where RGB can effectively predict the anomalies, but the point score map fails to recognize it. The false detection could arise from unusual point density and distribution. To demonstrate the effect of point density, we randomly select normal regions of point clouds and subsequently increase or decrease the density of these regions through upsampling and downsampling. We provide a qualitative analysis in~\ref{figure: The impact of noise level}. The visualization shows that PointAD effectively resists noise at reasonable levels. However, when noise levels are extremely low or high, the corresponding regions become excessively sparse or dense. This causes normal regions to appear similar to hole anomalies or squeezed anomalies, leading PointAD to classify these noisy areas as anomalies. Non-parametric score alignment and filter methods could be a potential direction, which we leave for further work.
\begin{table}[]
    \centering
    \setlength\tabcolsep{3pt} 
        \caption{Effect on point density gap.}
    \label{tab: Effect on point density gap.}
    \tiny
\begin{tabular}{c|ccccc}
    \toprule
    % \multicolumn{5}{c|}{Proposed technology}& \multicolumn{4}{c}{MVTEC}\\ \hline
    \multirow{1}{*}{Downsample}  & \multicolumn{2}{c}{Point detection} & \multicolumn{2}{c}{Multimodal detection} \\ 
    ratio & Global  & Local & Global  & Local  \\ \hline
    \rowcolor{gray!40}
     100\% &  (\textcolor{red}{82.0}, \textcolor{red}{94.2}) & (\textcolor{red}{95.5}, \textcolor{red}{84.4}) & (\textcolor{red}{86.9}, \textcolor{red}{96.1}) & (\textcolor{red}{97.2}, \textcolor{red}{90.2})\\
     70\% & (81.3, 94.0 & (94.9, 82.6) & (85.6, 95.7) & (96.8, 90.0)\\  
    50\% & (79.6, 93.4) & (94.7, 81.8) & (84.9, 95.5) & (96.8, 89.9)\\
    30\% & (76.6, 91.8) & (94.5, 79.9) & (83.5, 95.1) & (96.4, 89.5)\\
    20\% & (72.7, 90.5) & (94.2, 78.2) & (82.0, 94.6) & (95.2, 88.6)\\
    \bottomrule
    \end{tabular}
\end{table}
\paragraph{Robustness to point density}
To investigate the impact of point density differences between the training and test datasets, we train PointAD using high-density point clouds (original dataset) and then test it on the low-density versions of the datasets, downsampled as described in \textbf{Input resolution}. Table~\ref{tab: Effect on point density gap.} shows that PointAD can still detect anomalies even when retaining 50\% of the points from the original point clouds. However, when more points are removed (30\% and 20\% sample ratio), PointAD experiences an obvious performance degradation. We attribute the misdetection to the overly sparse point clouds forming holes. Nevertheless, PointAD can still detect anomalies even with a significant gap in point density between the training and test domains (e.g., 100\% vs. 20\%). This demonstrates PointAD's ability to generalize across different point densities.
\section{Complexity analysis}
\label{Complexity analysis}
\begin{table}
    \centering
    \caption{Comparison of computation overhead with SOTA approaches on MVTec3D-AD. The unsupervised method is abbreviated as Un.}
    \tiny
    \begin{tabular}{c|cccccccc}
    \toprule
    \multirow{2}{*}{Methods} & \multirow{2}{*}{\makecell[c]{Inference time (s)}} & \multirow{2}{*}{\makecell[c]{FPS}} & \multirow{2}{*}{\makecell[c]{GPU memory\\ usage (Peak)}} & \multicolumn{2}{c}{Point detection} & \multicolumn{2}{c}{Multimodal detection} \\ 
    &&&& Global  & Local & Global  & Local  \\ \hline
    
    BTF (Un.) & 0.18 & 5.56 & 1934	& (76.3, 91.8) & (97.6, 92.3)&	(89.8, 96.7) &	(99.5, 97.1) \\
    SDM (Un.) & 0.14 & 7.14	& 2716	& (96.7, 90.9) &(97.0, 91.8) & (92.1, 97.6) & (93.3, 98.1)\\
    M3DM (Un)	&2.86	&0.35	&7494	&(85.6, 93.4)&	(92.8, 91.6)	&(93.2, 92.7)&	(98.4, 95.1)\\
    CPFM (Un.) & 0.22&	4.55&	1379&	(94.9, 98.7)&	(97.6, 92.5)&	(-,-)&	(-,-) \\
    3DSR (Un.) & 0.09&	11.11&	3067& (95.1, 94.3) &	(93.8, 91.2)	&(97.3, 98.6) &	(99.3, 97.4) \\
    PointCLIP V2 (ZS) & 1.52& 0.66 & 9747MB& (78.3, 49.4) & (87.4, 52.3) & (49.8, 79.3) & (51.2, 80.1)\\
    CLIP + Rendering (ZS) & 0.27 & 3.61 & 3685MB & (-, 54.4) & (61.2, 85.8) & (-, 56.0) & (60.4, 86.4)\\  
    Cheraghian (ZS) & 0.35&  2.86 & 4847MB& (53.6, 81.7) & (88.2, 57.0) & (-, -) & (-, -)\\   
    WinCLIP (ZS) & 0.29 & 3.45 & 3914MB&  (45.2, 77.9 ) & (85.8, 59.4) & (38.7, 74.1) & (87.5, 64.2)\\ 
     AnomalyCLIP (ZS) & 0.19 & 5.26 & 3348MB& (56.4, 83.5) & ( 88.9, 60.9) & ( 66.2, 87.6) & ( 91.6 70.9)\\  
    \rowcolor{gray!40}
    Ours (ZS) &  0.40 & 2.52& 4275MB & (\textcolor{red}{82.0}, \textcolor{red}{94.2}) & (\textcolor{red}{95.5}, \textcolor{red}{84.4}) & (\textcolor{red}{86.9}, \textcolor{red}{96.1}) & (\textcolor{red}{97.2}, \textcolor{red}{90.2})\\
    \bottomrule
    \end{tabular}
    \label{table7: Comparison of computation overhead with PointCLIP V2.}
\end{table}
In Table~\ref{table7: Comparison of computation overhead with PointCLIP V2.}, we provide a comparison of computation overhead among unsupervised and zero-shot manners~\footnote{Since the official code for WinCLIP is not available, we reproduce its results using the implementation found at \url{https://github.com/zqhang/Accurate-WinCLIP-pytorch}}. The evaluation includes inference time per image, frames per second (FPS), and GPU memory consumption with a batch size of 1. For a fair comparison, we keep NVIDIA RTX 3090 24GB GPU free until we conduct experiments. Compared to PointCLIP V2, our model requires less time to infer an image, achieving higher FPS (2.52 vs. 0.66) with lower graphic memory usage (as discussed in Section~\ref{sec: hybrid loss}). While CLIP + rendering has a slight advantage in computation overhead, our detection performance significantly outperforms it. Therefore, PointAD achieves a favorable trade-off between performance and computation overhead.
\begin{figure}[h]
% \vskip 0.2in
\begin{center}
\centerline{\includegraphics[width = 0.6\columnwidth]{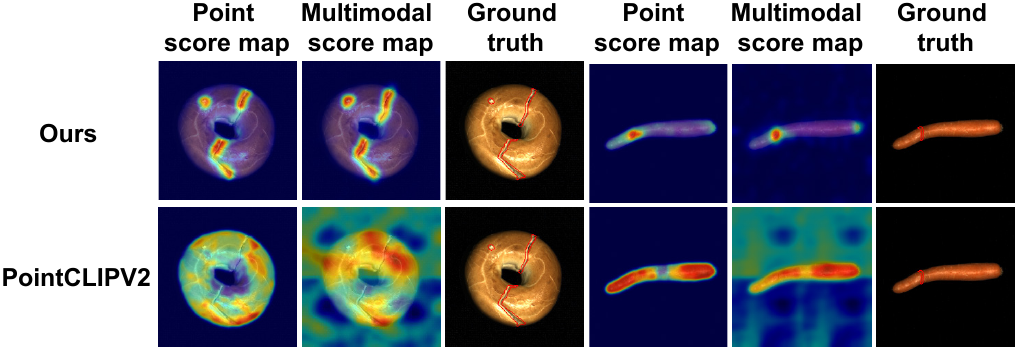}}
\caption{Visualization comparison between PointAD and PointCLIP V2.}
\label{f10:visualization_comparison}
\end{center}
% \vskip -0.2in
\end{figure}

\section{Visualization Comparison}
To provide intuitive results, we compare the visualization of PointAD with PointCLIP V2. In Figure~\ref{f10:visualization_comparison}, our model achieves accurate ZS 3D detection through the point cloud. Moreover, given RGB counterparts, PointAD further improves its detection capacity in M3D detection. However, PointCLIP V2 exhibits noisy activations for normal regions. After incorporating RGB information, PointCLIP V2 appears to struggle to fuse these two modalities in a plug-and-play manner, unlike PointAD.

\section{Additional Visualization}
\label{appendix: addition_visualization}
\begin{figure}
\begin{center}
\centerline{\includegraphics[width=1\textwidth]{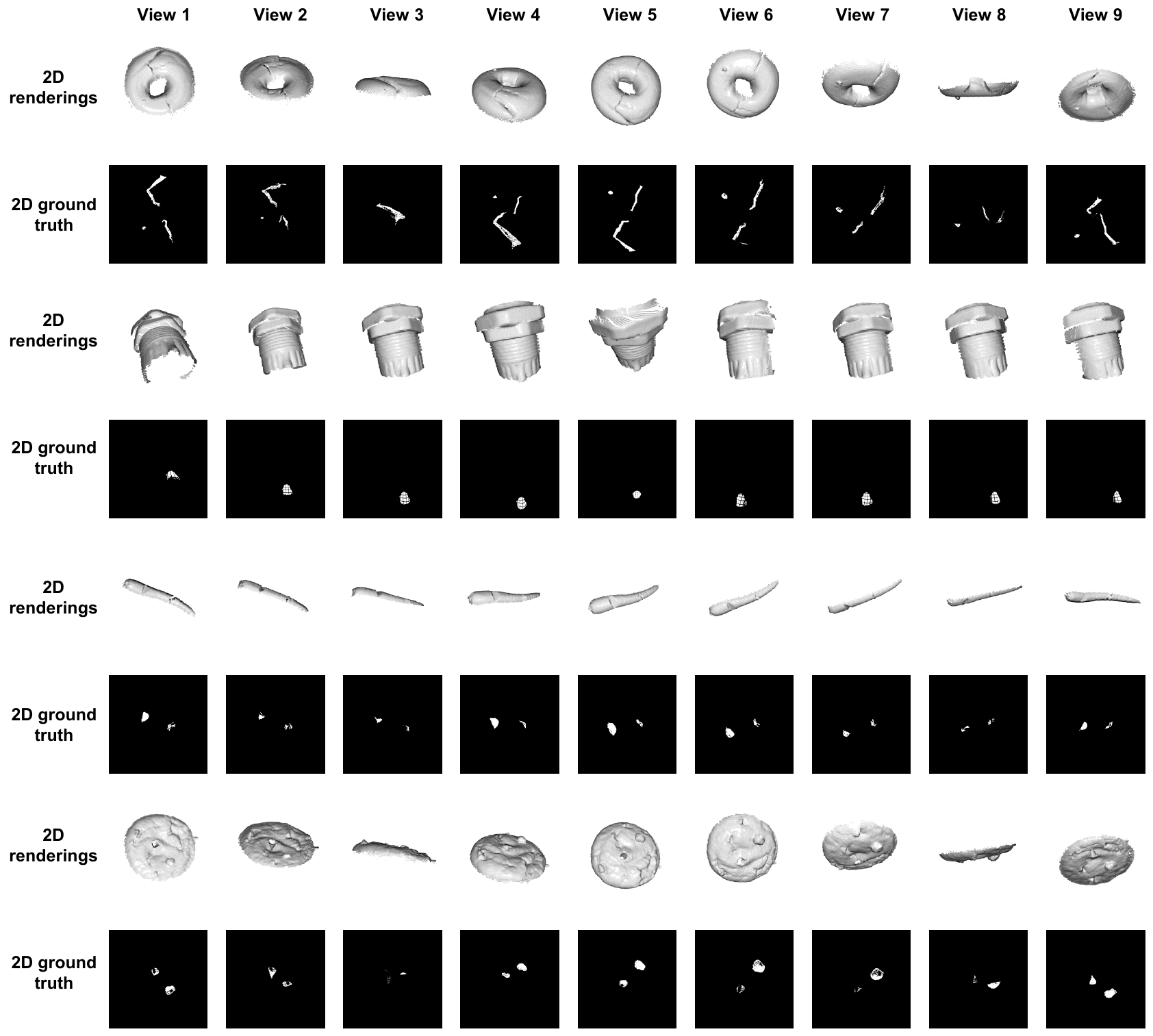}}
\caption{Visualization about 2D renderings and ground truth from different views ($K=9$).}
\label{f9:additional_visualization_2}
\end{center}
% \vskip -0.2in
\end{figure}

\begin{figure}
\begin{center}
\centerline{\includegraphics[width=1\textwidth]{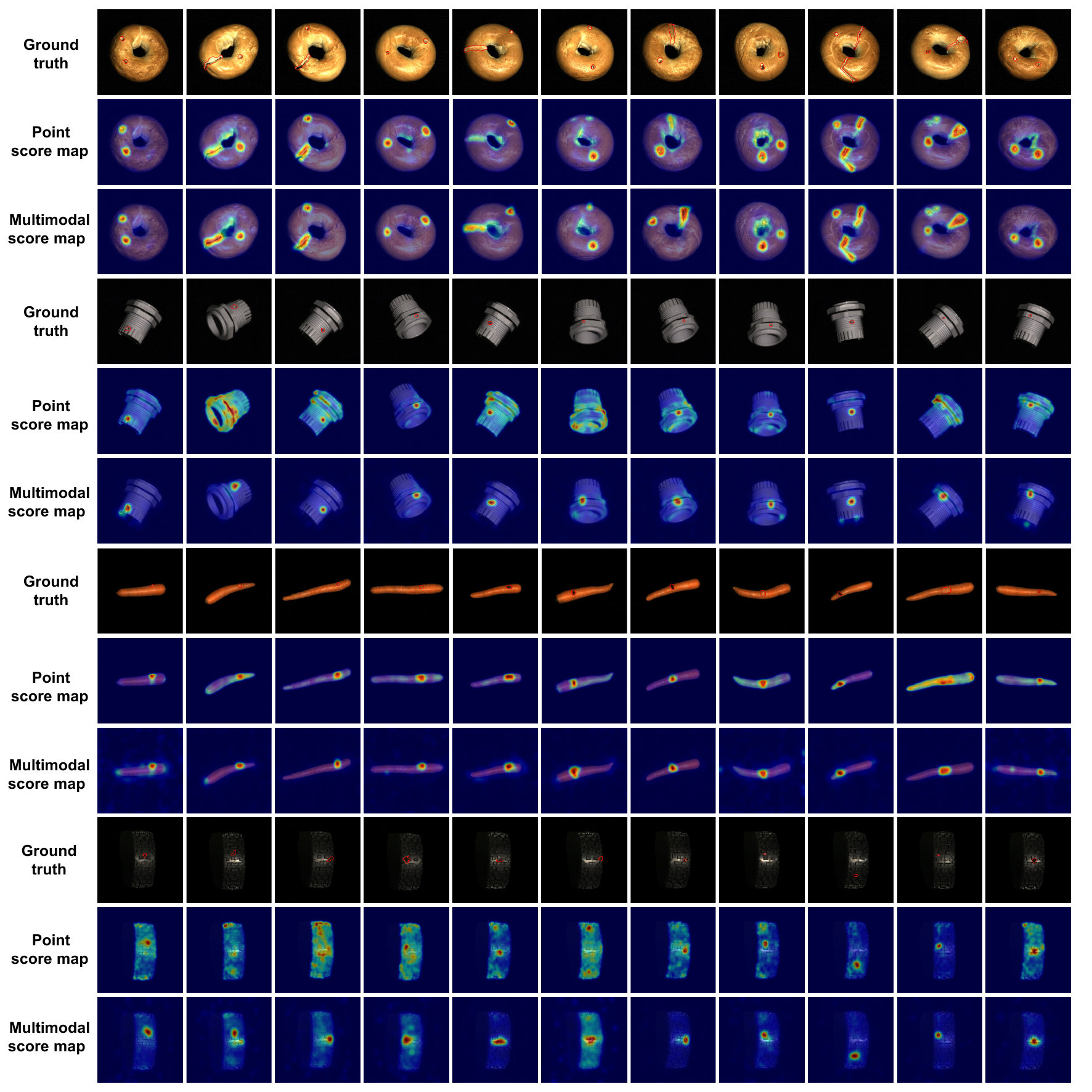}}
\caption{Visualization of point and multimodal score maps in PointAD, which is pre-trained on cookie object.}
\label{f10:visualization}
\end{center}
% \vskip -0.2in
\end{figure}

We respectively visualize the 2D renderings and corresponding 2D ground truths, which are rendered from 3D pint clouds and ground truths, as shown in Figure~\ref{f9:additional_visualization_2}. We also supplement more zero-shot segmentation results of PointAD in Figure~\ref{f10:visualization}.

\newpage
\vspace{-2em}
\section{Detailed results}
\begin{itemize}
    \item We provide the class-level ZS 3D results on MVTec3D-AD in Table~\ref{table1:3D local.}.
    \item We provide the class-level ZS M3D results on MVTec3D-AD in Table~\ref{table2:M3D global.}.
    \item We provide the class-level ZS 3D results on Eyecandies in Table~\ref{table:3D on Eyecandies}.
    \item We provide the class-level ZS M3D results on Eyecandies in Table~\ref{table:multimodal on eyecandies}.
    \item We provide the class-level ZS 3D results on Real3D-AD in Table~\ref{table:3D on Real3D-AD}.
    \item We provide the class-level ZS cross-dataset 3D results on Eyecandies from MVTec3D-AD in Table~\ref{table:3D Eyecandies transfer}.
    \item We provide the class-level ZS cross-dataset 3D results on Eyecandies from MVTec3D-AD in Table~\ref{table:multimodal Eyecandies transfer.}.
    \item We provide the class-level ZS cross-dataset 3D results on Real3D-AD from MVTec3D-AD in Table~\ref{table:3D Real3D-AD transfer.}.
\end{itemize}
\newpage

\begin{table*}[t]
\centering
\caption{Performance comparison on ZS 3D anomaly detection. The best and second-best results in ZS are highlighted in red and blue. G. and L. represent the global and local anomaly detection.} 
\vspace{-0.6em}
\label{table1:3D local.}
\tiny
\setlength\tabcolsep{1.5pt} 
\begin{tabular}{cccccccccccc|c}
\toprule
& Method & Bagel & \makecell[c]{Cable gland} & Carrot & Cookie & Dowel  & Foam & Peach & Potato & Rope & Tire & Mean  \\ \hline
\multirow{4}{*}{\makecell[c]{G.}} 
& CLIP + R.  & (53.4, 85.2) & (49.6, 83.2) & (62.9, 89.9) & (65.0, 88.0) & (65.3, 89.3) & (53.0, 78.9) & (72.0, 89.2) & (58.5, 83.4) & (80.0, 90.7) & (52.4, 80.1) & (61.2, 85.8) \\

& Cheraghian  & (49.3, 80.5) & (47.1, 80.1) & (52.7, 83.8) & (54.4, 83.0) & (43.3, 78.6) & (47.4, 80.6) & (50.0, 80.5) & (59.7, 84.4) & (72.8, 85.2) & (\textcolor{blue}{59.8}, 80.5) & (53.6, 81.7) \\

& PoinCLIP V2  & (\textcolor{blue}{71.7}, 35.9) & (\textcolor{red}{68.6}, 39.2) & (\textcolor{blue}{94.3}, 83.6) & (69.8, 28.5) & (\textcolor{red}{75.5}, 47.7) & (67.1, 51.4) & (69.7, 36.5) & (84.6, 57.6) & (\textcolor{red}{91.8}, 76.1) & (\textcolor{red}{89.8}, 67.5) & (51.2, 80.1) \\

& PointCLIP V2$_a$ & (47.1, 80.7) & (\textcolor{blue}{55.1}, 84.7) & (47.7, 80.8) & (50.1, 79.8) & (50.9, 82.9) & (57.4, 83.7) & (52.4, 83.5) & (48.2, 78.5) & (54.8, 74.4) & (47.8, 76.8) & (51.1, 80.6) \\

& AnomalyCLIP & (62.8, \textcolor{blue}{86.9}) & (51.2, 82.2) & (51.9, 84.4) & (64.9, 86.2) & (50.0, 80.3) & (42.4, 80.1) & (69.4, 90.8) & (61.5, 85.7) & (62.4, 81.5) & (47.8, 77.1) & (56.4, 83.5) \\

& PointAD-CoOp  & (\textcolor{red}{98.3}, \textcolor{red}{99.6}) & (53.8, \textcolor{blue}{85.7}) & (93.2, \textcolor{blue}{98.5}) & (\textcolor{blue}{89.5}, \textcolor{blue}{97.1}) & (66.3, \textcolor{blue}{89.8}) & (\textcolor{red}{70.3}, \textcolor{red}{91.3}) & (\textcolor{blue}{89.1}, \textcolor{blue}{96.9}) & (\textcolor{blue}{97.8}, \textcolor{blue}{99.5}) & (91.1, \textcolor{blue}{96.5}) & (59.2, \textcolor{red}{83.7}) & (\textcolor{blue}{80.9}, \textcolor{blue}{93.9}) \\

\rowcolor{gray!40}
& PointAD  &(\textcolor{red}{98.3}, \textcolor{red}{99.6})& (53.7, \textcolor{red}{86.0}) & (\textcolor{red}{97.9}, \textcolor{red}{99.6}) &(\textcolor{red}{92.1}, \textcolor{red}{97.9}) & (\textcolor{blue}{72.2}, \textcolor{red}{92.0})  &(\textcolor{blue}{69.5}, \textcolor{blue}{91.2}) &(\textcolor{red}{91.5}, \textcolor{red}{97.6}) &(\textcolor{red}{98.8}, \textcolor{red}{99.7})&(\textcolor{blue}{91.5}, \textcolor{red}{96.7})&(54.1, \textcolor{blue}{82.1})&(\textcolor{red}{82.0}, \textcolor{red}{94.2}) \\
\midrule

\multirow{4}{*}{\makecell[c]{L.}}
\vspace{-0.1em}

& CLIP + R. &(-, 22.2)& (-, 67.5) & (-, 77.4) &(-, 6.7) & (-, 65.6)  &(-, 37.4) &(-, 38.8) &(-, 77.2)&(-, 72.2)&(-, 79.4)&(-, 54.4) \\

& Cheraghian  & (73.2, 13.8) & (93.0, 75.6) & (83.9, 45.6) & (82.0, 39.8) & (93.6, 67.0) & (84.4, 45.9) & (84.5, 40.1) & (95.7, 77.1) & (96.2, 74.9) & (\textcolor{red}{95.9}, \textcolor{red}{89.8}) & (88.2, 57.0) \\

& PoinCLIP V2 & (78.2, 36.0) & (90.9, 65.7) & (96.4, 76.0) & (74.7, 24.4) & (\textcolor{blue}{93.8}, 66.0) & (75.1, 18.9) & (86.0, 43.4) & (92.8, 60.1) & (95.6, 71.7) & (90.3, 63.3) & (87.4, 52.3) \\

& PointCLIP V2$_a$ & (79.4, 38.0) & (91.5, 68.0) & (96.3, 75.5) & (74.7, 23.2) & (93.7, 65.3) & (73.0, 16.4) & (86.2, 44.4) & (92.7, 58.8) & (95.3, 69.6) & (90.4, 64.1) & (87.3, 52.3) \\

& AnomalyCLIP & (86.0, 49.0) & (89.1, 59.7) & (94.4, 73.6) & (79.7, 40.2) & (93.4, 73.5) & (78.3, 31.2) & (88.6, 61.2) & (93.6, 75.6) & (96.7, 84.3) & (89.2, 60.7) & (88.9, 60.9) \\

& PointAD-CoOp & (\textcolor{blue}{97.5}, \textcolor{blue}{94.7}) & (\textcolor{blue}{93.3}, \textcolor{blue}{78.5}) & (\textcolor{blue}{99.2}, \textcolor{blue}{95.6}) & (\textcolor{blue}{85.6}, \textcolor{blue}{69.2}) & (\textcolor{red}{95.5}, \textcolor{blue}{74.6}) & (\textcolor{blue}{85.8}, \textcolor{blue}{51.1}) & (\textcolor{blue}{98.9}, \textcolor{blue}{96.6}) & (\textcolor{blue}{99.6}, \textcolor{blue}{97.9}) & (\textcolor{blue}{99.0}, \textcolor{blue}{86.7}) & (94.0, 75.2) & (\textcolor{blue}{94.8}, \textcolor{blue}{82.0}) \\

\rowcolor{gray!40}
& PointAD  &(\textcolor{red}{98.4}, \textcolor{red}{96.9})& (\textcolor{red}{93.5}, \textcolor{red}{79.5}) & (\textcolor{red}{99.4}, \textcolor{red}{96.4}) &(\textcolor{red}{87.5}, \textcolor{red}{75.4}) & (\textcolor{red}{95.5}, \textcolor{red}{75.2})  &(\textcolor{red}{86.5}, \textcolor{red}{54.1}) &(\textcolor{red}{99.5}, \textcolor{red}{98.3}) &(\textcolor{red}{99.9}, \textcolor{red}{99.1})&(\textcolor{red}{99.3}, \textcolor{red}{89.9})&(\textcolor{blue}{95.3}, \textcolor{blue}{79.7})&(\textcolor{red}{95.5}, \textcolor{red}{84.4}) \\
\bottomrule
\end{tabular}%

%\vspace{-1.5em}
\end{table*}

\newpage

\newcommand{\td}[1]{\textcolor{red}{#1}}
\begin{table*}[t]
\centering
\caption{Performance comparison on ZS M3D anomaly detection.
} 
\vspace{-0.6em}
\label{table2:M3D global.}
\tiny
\setlength\tabcolsep{1.5pt} 
\begin{tabular}{cccccccccccc|c}
\toprule
\vspace{-0.1em}
&Method& Bagel & \makecell[c]{Cable gland} & Carrot & Cookie & Dowel  & Foam & Peach & Potato & Rope & Tire & Mean  \\ \hline
\multirow{4}{*}{\makecell[c]{MG.}} 

& CLIP + R. & (55.1, 85.9) & (55.0, 84.1) & (64.5, 90.1) & (50.6, 83.1) & (59.1, 84.6) & (69.0, 90.7) & (72.0, 91.3) & (56.7, 85.5) & (70.8, 86.0) & (51.7, 82.9) & (60.4, 86.4) \\

& Cheraghian  &(-, -)& (-, -) & (-, -) &(-, -) & (-, -)  &(-, -) &(-, -) &(-, -)&(-, -)&(-, -)&(-, -) \\

& PoinCLIP V2  & (51.6, 83.7)& (63.8, 87.6) & (47.7, 83.5) &(47.8, 78.0) & (51.8, 80.5)  &(45.2, 78.5)  & (49.2, 78.7) &(55.4, 82.9)&(39.1, 62.4)&(46.0, 76.9)&(49.8, 79.3)\\

& PointCLIP V2$_a$ & (53.4, 84.4) & (64.7, 89.1) & (48.0, 83.4) & (48.4, 78.4) & (47.1, 81.1) & (45.9, 79.0) & (49.6, 79.2) & (55.5, 85.9) & (34.9, 60.5) & (46.1, 76.9) & (49.4, 79.8) \\

& AnomalyCLIP & (\textcolor{blue}{78.8}, \textcolor{blue}{93.5}) & (58.1, 84.0) & (63.2, 88.7) & (72.3, 89.1) & (53.8, 82.5) & (65.1, 89.8) & (73.7, 91.1) & (64.3, 85.8) & (77.5, 89.1) & (55.2, 82.5) & (66.2, 87.6) \\

& PointAD-CoOp & (\textcolor{red}{98.8}, \textcolor{red}{99.7}) & (\textcolor{blue}{74.6}, \textcolor{blue}{93.0}) & (\textcolor{blue}{90.0}, \textcolor{blue}{97.5}) & (\textcolor{blue}{88.1}, \textcolor{red}{96.3}) & (\textcolor{blue}{66.2}, \textcolor{blue}{88.9}) & (\textcolor{blue}{79.8}, \textcolor{blue}{94.6}) & (\textcolor{blue}{90.8}, \textcolor{blue}{97.7}) & (\textcolor{blue}{83.1}, \textcolor{blue}{93.7}) & (\textcolor{blue}{93.8}, \textcolor{red}{97.6}) & (\textcolor{blue}{68.7}, \textcolor{blue}{89.6}) & (\textcolor{blue}{83.4}, \textcolor{blue}{94.9}) \\

\rowcolor{gray!40}
& PointAD  & (\textcolor{red}{98.8}, \textcolor{red}{99.7})& (\textcolor{red}{79.9}, \textcolor{red}{94.7}) & (\textcolor{red}{95.5}, \textcolor{red}{98.9}) &(\textcolor{red}{86.2}, \textcolor{blue}{95.5}) & (\textcolor{red}{98.5}, \textcolor{red}{90.6})  &(\textcolor{red}{84.4}, \textcolor{red}{96.1})  & (\textcolor{red}{96.6}, \textcolor{red}{99.1}) &(\textcolor{red}{90.7}, \textcolor{red}{97.0})&(\textcolor{red}{93.6}, \textcolor{blue}{97.3})&(\textcolor{red}{74.6}, \textcolor{red}{92.0})&(\textcolor{red}{86.9}, \textcolor{red}{96.1})\\

\midrule
\vspace{-0.1em}

\multirow{3}{*}{\makecell[c]{ML.}} 

& CLIP + R.  &(-, 17.9)& (-, 68.5) & (-, \textcolor{blue}{89.5}) &(-, 4.7) & (-, 74.3)  &(-, 22.1) &(-, 47.5) &(-, 82.7)&(-, 73.6)&(-, 78.9)&(-, 56.0) \\

& Cheraghian  &(-, -)& (-, -) & (-, -) &(-, -) & (-, -)  &(-, -) &(-, -) &(-, -)&(-, -)&(-, -)&(-, -) \\

& PoinCLIP V2  & (40.6, 78.0) & (56.1, 84.4) & (53.8, 84.2) & (52.7, 81.1) & (50.7, \textcolor{blue}{80.4}) & (40.8, \textcolor{blue}{78.1}) & (54.9, 82.8) & (48.9, 77.9) & (54.3, 72.5) & (59.3, 81.9) & (78.3, 49.4) \\

& PointCLIP V2$_a$ & (75.9, 40.8) & (76.2, 47.4) & (92.5, 79.9) & (71.7, 30.7) & (72.8, 44.9) & (62.3, 21.9) & (77.1, 46.4) & (87.4, 63.7) & (87.9, 69.9) & (90.8, 70.8) & (79.5, 51.6) \\

& AnomalyCLIP & (93.7, 71.1) & (90.7, 67.7) & (95.8, 84.7) & (82.0, 45.2) & (93.9, 77.1) & (84.3, 50.0) & (93.5, 79.2) & (95.6, 83.1) & (95.9, 83.4) & (91.2, 67.5) & (91.6, 70.9) \\

& PointAD-CoOp &(\textcolor{blue}{99.4}, \textcolor{red}{97.9}) & (\textcolor{blue}{95.7}, \textcolor{blue}{87.3}) & (\textcolor{blue}{99.3}, \textcolor{red}{97.3}) & (\textcolor{blue}{91.0}, \textcolor{blue}{82.7}) & (\textcolor{blue}{95.9}, \textcolor{red}{85.0}) & (\textcolor{blue}{91.8}, 72.2) & (\textcolor{blue}{98.7}, \textcolor{blue}{96.7}) & (\textcolor{blue}{99.4}, \textcolor{red}{97.9}) & (\textcolor{blue}{98.6}, \textcolor{blue}{91.9}) & (\textcolor{blue}{94.8}, \textcolor{blue}{79.2}) & (\textcolor{blue}{96.5}, \textcolor{blue}{88.8})\\

\rowcolor{gray!40}
& PointAD  & (\textcolor{red}{99.6}, \textcolor{blue}{90.1})& (\textcolor{red}{96.7}, \textcolor{red}{97.9}) & (\textcolor{red}{99.4}, 85.5) &(\textcolor{red}{92.6}, \textcolor{red}{85.4}) & (\textcolor{red}{96.1}, 74.0)  &(\textcolor{red}{92.4}, \textcolor{red}{98.3})  & (\textcolor{red}{99.4}, \textcolor{red}{98.9}) &(\textcolor{red}{99.8}, \textcolor{blue}{92.9})&(\textcolor{red}{98.8}, \textcolor{red}{87.9})&(\textcolor{red}{97.5}, \textcolor{red}{91.1})&(\textcolor{red}{97.2}, \textcolor{red}{90.2})\\

\bottomrule
\end{tabular}%
\end{table*}

\newpage
%%%%%%%%%%%%%%%%%%%%%%%%%%%%%%%%%%%%%%%%%%%%%%%%%%%%%%%%%%%%%%%%%%%%%%%%%%%%%%%
%%%%%%%%%%%%%%%%%%%%%%%%%%%%%%%%%%%%%%%%%%%%%%%%%%%%%%%%%%%%%%%%%%%%%%%%%%%%%%%

%%%%%%%%%%%%%%%%%%%%%%%%%%%%%%%%%%%%%%%%%%%%%%%%%%%%%%%%%%%%

\begin{table*}
\centering
\caption{Performance comparison on ZS 3D anomaly detection on Eyecandies.}
\vspace{-0.6em}
\label{table:3D on Eyecandies}
\tiny
\setlength\tabcolsep{3pt}
\begin{tabular}{cccccccccccc|c}
\toprule
& Method & \makecell{Candy \\ Cane} & \makecell{Chocolate \\ Cookie} & \makecell{Chocolate \\ Praline} & \makecell{Confetto} & \makecell{Gummy \\ Bear} & \makecell{Hazelnut \\ Truffle} & \makecell{Licorice \\ Sandwich} & \makecell{Lollipop} & \makecell{Marsh- \\ mallow} & \makecell{Peppermint \\ Candy} & Mean \\
\midrule
\multirow{5}{*}{\makecell[c]{G.}}
& CLIP + Rendering & (\textcolor{red}{61.8}, \textcolor{red}{60.9}) & (48.3, 53.8) & (61.8, 72.0) & (\textcolor{blue}{82.1}, 86.9) & (\textcolor{red}{81.4}, \textcolor{red}{83.2}) & (57.3, 54.4) & (72.6, 71.9) & (66.2, 55.8) & (60.6, 68.9) & (75.4, 83.9) & (66.7, 69.2) \\

& Cheraghian & (\textcolor{blue}{50.0}, 50.0) & (50.0, 50.0) & (50.0, 50.0) & (50.0, 50.0) & (50.0, 48.0) & (50.0, 50.0) & (50.0, 50.0) & (46.7, 30.8) & (48.0, 52.1) & (50.0, 50.0) & (49.5, 48.1) \\

& PoinCLIPV2 & (45.1, 48.9) & (\textcolor{red}{55.5}, 54.7) & (37.3, 42.1) & (30.9, 40.3) & (33.5, 42.4) & (40.0, 43.0) & (67.0, 62.7) & (41.2, 28.1) & (54.1, 55.6) & (56.0, 63.5) & (46.1, 48.1) \\

& PointCLIP V2$_a$ & (45.8, \textcolor{blue}{51.1}) & (44.3, 54.7) & (30.2, 40.4) & (35.8, 42.4) & (42.5, 46.3) & (33.0, 41.0) & (59.9, 58.2) & (47.1, 30.9) & (59.1, 57.5) & (46.7, 47.9) & (44.4, 47.0) \\

& AnomalyCLIP & (44.8, 47.7) & (34.9, 42.0) & (57.5, 62.0) & (74.3, 76.3) & (49.3, 52.2) & (\textcolor{red}{69.8}, \textcolor{red}{70.7}) & (52.6, 58.2) & (60.1, 46.2) & (64.0, 62.4) & (68.4, 72.7) & (57.6, 59.0) \\

& PointAD-CoOp &(45.7, 48.1) & (56.1, \textcolor{red}{61.4}) & (\textcolor{blue}{72.6}, \textcolor{blue}{82.8}) & (\textcolor{red}{82.5}, \textcolor{red}{87.8}) & (66.2, 71.8) & (\textcolor{blue}{60.8}, \textcolor{blue}{63.9}) & (\textcolor{blue}{80.6}, \textcolor{blue}{84.5}) & (\textcolor{blue}{70.5}, \textcolor{blue}{62.8}) & (\textcolor{blue}{64.1}, \textcolor{blue}{69.7}) & (\textcolor{blue}{77.9}, \textcolor{blue}{85.3}) & (\textcolor{blue}{67.7}, \textcolor{blue}{71.8}) \\

\rowcolor{gray!40}
& Point-AD & (42.8, 51.0) & (\textcolor{blue}{51.2}, \textcolor{blue}{55.4}) & (\textcolor{red}{75.1}, \textcolor{red}{84.0}) & (81.4, \textcolor{blue}{87.2}) & (\textcolor{blue}{70.1}, \textcolor{blue}{78.7}) & (59.9, 61.3) & (\textcolor{red}{81.8}, \textcolor{red}{85.5}) & (\textcolor{red}{80.1}, \textcolor{red}{75.2}) & (\textcolor{red}{68.3}, \textcolor{red}{73.3}) & (\textcolor{red}{80.4}, \textcolor{red}{86.4}) & (\textcolor{red}{69.1}, \textcolor{red}{73.8}) \\

\midrule
\multirow{5}{*}{\makecell[c]{L.}}

& CLIP + Rendering & (97.3, 84.0) & (\textcolor{blue}{77.7}, 24.2) & (71.0, 19.2) & (76.7, 25.6) & (84.5, 37.4) & (76.5, 33.9) & (79.3, 30.9) & (94.6, 66.6) & (70.9, 17.7) & (83.1, 39.4) & (81.2, 37.9) \\

& Cheraghian & (-, -)& (-, -) & (-, -) &(-, -) & (-, -)  &(-, -) &(-, -) &(-, -)&(-, -)&(-, -)&(-, -)\\

& PoinCLIPV2 & (45.0, -) & (38.4, -) & (48.9, -) & (43.3, -) & (45.0, -) & (54.3, 19.6) & (38.6, -) & (43.4, -) & (42.3, -) & (37.9, -) & (43.7, -) \\
& PoinCLIPV2$_a$ & (45.0, -) & (38.4, -) & (51.2, 16.1) & (43.5, -) & (45.2, -) & (55.2, 21.2) & (38.8, -) & (43.4, -) & (43.1, 15.7) & (37.9, -) & (44.2, -) \\

& AnomalyCLIP & (95.9, 84.3) & (73.4, 32.2) & (79.3, 43.0) & (74.6, 37.7) & (78.7, 39.6) & (68.9, 26.1) & (76.9, 38.4) & (92.0, 71.9) & (58.5, 16.2) & (79.1, 39.1) & (77.7, \textcolor{blue}{43.4}) \\

& PointAD-CoOp & (\textcolor{red}{98.1}, \textcolor{red}{88.4}) & (\textcolor{red}{92.9}, \textcolor{red}{75.6}) & (\textcolor{red}{91.4}, \textcolor{red}{68.2}) & (\textcolor{blue}{93.5}, \textcolor{red}{69.4}) & (\textcolor{red}{89.0}, \textcolor{blue}{69.2}) & (\textcolor{blue}{83.3}, \textcolor{blue}{44.8}) & (\textcolor{blue}{91.5}, \textcolor{red}{74.7}) & (\textcolor{blue}{97.3}, \textcolor{blue}{83.0}) & (\textcolor{blue}{86.1}, \textcolor{blue}{64.0}) & (\textcolor{blue}{92.3}, \textcolor{blue}{75.7}) & (\textcolor{blue}{91.5}, \textcolor{red}{71.3}) \\

\rowcolor{gray!40}
& Point-AD & (\textcolor{blue}{98.0}, \textcolor{blue}{87.8}) & (\textcolor{red}{92.9}, \textcolor{blue}{74.8}) & (\textcolor{blue}{90.9}, \textcolor{blue}{65.7}) & (\textcolor{red}{94.3}, \textcolor{blue}{68.2}) & (\textcolor{blue}{88.7}, \textcolor{red}{71.4}) & (\textcolor{red}{84.6}, \textcolor{red}{45.1}) & (\textcolor{red}{93.8}, \textcolor{blue}{72.6}) & (\textcolor{red}{97.8}, \textcolor{red}{86.6}) & (\textcolor{red}{87.8}, \textcolor{red}{64.4}) & (\textcolor{red}{92.6}, \textcolor{red}{76.6}) & (\textcolor{red}{92.1}, \textcolor{red}{71.3}) \\
\bottomrule
\end{tabular}
\vspace{-1.5em}
\end{table*}
\newpage

\begin{table*}
\centering
\caption{Performance comparison on ZS M3D anomaly detection on Eyecandies.}
\label{table:multimodal on eyecandies}
\vspace{-0.6em}
\tiny
\setlength\tabcolsep{3pt}
\begin{tabular}{cccccccccccc|c}
\toprule
& Method & \makecell{Candy \\ Cane} & \makecell{Chocolate \\ Cookie} & \makecell{Chocolate \\ Praline} & \makecell{Confetto} & \makecell{Gummy \\ Bear} & \makecell{Hazelnut \\ Truffle} & \makecell{Licorice \\ Sandwich} & \makecell{Lollipop} & \makecell{Marsh- \\ mallow} & \makecell{Peppermint \\ Candy} & Mean \\
\midrule
\multirow{5}{*}{\makecell[c]{MG.}}
& CLIP + Rendering & (\textcolor{red}{64.3}, \textcolor{red}{67.8}) & (76.6, 77.6) & (64.3, 70.8) & (88.0, 89.6) & (\textcolor{blue}{70.4}, \textcolor{blue}{72.1}) & (55.5, 53.8) & (78.4, 81.9) & (\textcolor{red}{71.5}, \textcolor{red}{64.7}) & (77.1, 77.8) & (83.4, 82.7) & (73.0, 73.9) \\

& Cheraghian &(-, -)& (-, -) & (-, -) &(-, -) & (-, -)  &(-, -) &(-, -) &(-, -)&(-, -)&(-, -)&(-, -) \\
& PoinCLIPV2 & (43.0, 48.3) & (48.0, 55.0) & (46.4, 51.6) & (49.3, 48.4) & (44.7, 49.1) & (48.3, 55.4) & (61.8, 70.0) & (42.1, 30.4) & (54.1, 51.5) & (31.2, 39.6) & (46.9, 49.9) \\
& PoinCLIPV2$_a$ & (44.1, \textcolor{blue}{51.2}) & (44.5, 52.4) & (48.6, 52.3) & (56.7, 54.8) & (42.8, 44.6) & (55.7, 60.3) & (63.5, 68.7) & (43.8, 29.3) & (54.0, 52.1) & (31.3, 39.6) & (48.5, 50.5)\\

& AnomalyCLIP & (\textcolor{blue}{49.7}, 50.8) & (57.1, 62.9) & (66.5, 70.3) & (66.7, 68.0) & (64.0, 69.5) & (61.1, 67.8) & (69.2, 73.9) & (68.8, 56.5) & (77.3, 80.4) & (69.5, 74.6) & (65.0, 67.5) \\

& PointAD-CoOp & (39.4, 45.6) & (\textcolor{blue}{81.6}, \textcolor{blue}{87.2}) & (\textcolor{blue}{84.0}, \textcolor{blue}{87.7}) & (\textcolor{blue}{88.6}, \textcolor{blue}{91.1}) & (66.5, 69.2) & (\textcolor{red}{64.3}, \textcolor{red}{69.2}) & (\textcolor{blue}{82.7}, \textcolor{blue}{84.0}) & (64.9, 53.0) & (\textcolor{blue}{80.0}, \textcolor{blue}{83.5}) & (\textcolor{blue}{84.7}, \textcolor{blue}{89.6}) & (\textcolor{blue}{73.7}, \textcolor{blue}{76.0}) \\

\rowcolor{gray!40}
& Point-AD & (42.8, 49.1) & (\textcolor{red}{85.3}, \textcolor{red}{89.3}) & (\textcolor{red}{86.6}, \textcolor{red}{89.8}) & (\textcolor{red}{89.5}, \textcolor{red}{92.3}) & (\textcolor{red}{75.3}, \textcolor{red}{77.8}) & (\textcolor{blue}{61.4}, \textcolor{blue}{68.5}) & (\textcolor{red}{87.2}, \textcolor{red}{89.0}) & (\textcolor{blue}{70.4}, \textcolor{blue}{63.9}) & (\textcolor{red}{86.8}, \textcolor{red}{89.5}) & (\textcolor{red}{91.8}, \textcolor{red}{94.2}) & (\textcolor{red}{77.7}, \textcolor{red}{80.4}) \\
\midrule
\multirow{5}{*}{\makecell[c]{ML.}}
& CLIP + Rendering & (\textcolor{blue}{97.3}, \textcolor{blue}{89.2}) & (72.8, 10.3) & (65.5, 8.3) & (75.0, 17.4) & (83.9, 31.8) & (72.4, 29.8) & (75.6, 17.4) & (94.4, 75.8) & (66.3, 15.7) & (76.7, 22.2) & (78.0, 31.8) \\

& Cheraghian &(-, -)& (-, -) & (-, -) &(-, -) & (-, -)  &(-, -) &(-, -) &(-, -)&(-, -)&(-, -)&(-, -) \\
& PoinCLIPV2 & (44.8, -) & (44.8, -) & (48.0, -) & (59.6, -) & (48.6, -) & (53.9, -) & (42.2, -) & (33.7, -) & (43.3, -) & (41.4, -) & (46.0, -) \\

& PoinCLIPV2$_a$ & (44.8, -) & (44.7, -) & (49.0, -) & (59.3, -) & (48.2, -) & (54.2, -) & (42.2, -) & (33.6, -) & (45.0, -) & (41.5, -) & (46.2, -) \\

& AnomalyCLIP & (96.7, 88.5) & (89.5, 64.3) & (83.9, 55.4) & (92.1, 74.8) & (77.4, 33.1) & (70.2, 27.3) & (83.0, 52.9) & (94.6, 78.9) & (77.9, 38.2) & (84.8, 48.8) & (85.0, 56.2) \\

& PointAD-CoOp & (\textcolor{red}{97.5}, \textcolor{red}{91.1}) & (\textcolor{blue}{96.8}, \textcolor{blue}{87.9}) & (\textcolor{red}{94.8}, \textcolor{red}{85.2}) & (\textcolor{blue}{98.4}, \textcolor{blue}{93.7}) & (\textcolor{blue}{85.9}, \textcolor{blue}{57.3}) & (\textcolor{red}{88.1}, \textcolor{blue}{61.1}) & (\textcolor{red}{97.3}, \textcolor{red}{89.0}) & (\textcolor{blue}{96.6}, \textcolor{red}{86.3}) & (\textcolor{blue}{97.0}, \textcolor{blue}{87.1}) & (\textcolor{blue}{96.6}, \textcolor{blue}{87.2}) & (\textcolor{blue}{94.9}, \textcolor{blue}{82.6}) \\

\rowcolor{gray!40}
& Point-AD & (96.4, 87.2) & (\textcolor{red}{97.8}, \textcolor{red}{90.3}) & (\textcolor{blue}{93.5}, \textcolor{blue}{83.7}) & (\textcolor{red}{98.7}, \textcolor{red}{94.7}) & (\textcolor{red}{90.8}, \textcolor{red}{74.4}) & (\textcolor{blue}{87.7}, \textcolor{red}{62.5}) & (\textcolor{blue}{96.8}, \textcolor{blue}{88.0}) & (\textcolor{red}{96.7}, \textcolor{blue}{85.4}) & (\textcolor{red}{97.6}, \textcolor{red}{88.5}) & (\textcolor{red}{97.1}, \textcolor{red}{88.6}) & (\textcolor{red}{95.3}, \textcolor{red}{84.3}) \\
\bottomrule
\end{tabular}
\end{table*}
\newpage

\begin{table*}
\centering
\caption{Performance comparison on ZS 3D anomaly detection on Real3D-AD.}
\vspace{-0.6em}
\label{table:3D on Real3D-AD}
\tiny
\setlength\tabcolsep{2pt}
\begin{tabular}{cccccccccccccc|c}
\toprule
& Method & airplane & car & candybar & chicken & diamond & duck & fish & gemstone & seahorse & shell & starfish & toffees & Mean \\
\midrule
\multirow{5}{*}{\makecell[c]{G.}}
& CLIP + Rendering & (51.9, 56.9) & (50, 58.1) & (73.5, 70.8) & (\textcolor{red}{59.5}, \textcolor{red}{70.2}) & (84, 86) & (\textcolor{red}{77}, \textcolor{red}{76.2}) & (61.2, 69.3) & (68.3, 69.9) & (\textcolor{red}{83.6}, \textcolor{red}{87.9}) & (60.7, 54.5) & (69.5, 78.1) & (\textcolor{red}{86.9}, \textcolor{red}{89.6}) & (68.8, 72.3) \\

& Cheraghian & (57.8, 57.8) & (53.5, 57.4) & (50.7, 51.3) & (45.3, 56.2) & (41.9, 47.3) & (47.8, 56.3) & (49.0, 49.9) & (50.2, 53.9) & (45.6, 53.7) & (56.3, 56.7) & (51.5, 57.7) & (53.7, 54.4) & (50.3, 54.4) \\
& PoinCLIPV2 & (49.9, 48.9) & (41.7, 48.3) & (44.8, 49) & (46, 55.9) & (51.5, 48.9) & (\textcolor{blue}{68.4}, \textcolor{blue}{70.4}) & (60.9, 68.6) & (69.6, 69.3) & (53.1, 59.4) & (41.9, 52.8) & (31.1, 43.3) & (\textcolor{blue}{78.6}, \textcolor{blue}{82.4}) & (53.1, 58.1) \\
& PoinCLIPV2$_a$ & (46.5, 46.9) & (47.1, 48.1) & (51.3, 49.9) & (48.7, 57.1) & (47.8, 48.9) & (60.5, 56.2) & (60.6, 67.4) & (59.6, 60.8) & (74.7, 75.9) & (64.5, 58.8) & (61.7, 60.2) & (67.3, 69.8) & (57.5, 58.3)\\

& AnomalyCLIP & (\textcolor{red}{61.7}, 55.4) & (51.2, 52.7) & (49.7, 51.7) & (\textcolor{blue}{57.9}, \textcolor{blue}{67.7}) & (65.0, 65.2) & (56.2, 59.1) & (56.4, 64.0) & (49.1, 50.8) & (56.5, 57.4) & (53.1, 49.7) & (54.8, 58.7) & (51.0, 52.3) & (55.2, 57.1) \\

& PointAD-CoOp & (60.8, \textcolor{blue}{59.5}) & (\textcolor{blue}{68.7}, \textcolor{blue}{71.9}) & (\textcolor{red}{75.1}, \textcolor{red}{78.6}) & (45.3, 59.1) & (\textcolor{blue}{98.4}, \textcolor{blue}{98.5}) & (54.4, 47.5) & (\textcolor{red}{76.3}, \textcolor{red}{78.9}) & (\textcolor{blue}{86.6}, \textcolor{blue}{87.0}) & (\textcolor{blue}{81.4}, 78.9) & (\textcolor{blue}{89.4}, \textcolor{blue}{88.8}) & (\textcolor{red}{83.5}, \textcolor{red}{88.9}) & (67.1, 73.1) & (\textcolor{blue}{73.9}, \textcolor{blue}{75.9}) \\

\rowcolor{gray!40}
& Point-AD & (\textcolor{blue}{60.9}, \textcolor{red}{61.6}) & (\textcolor{red}{73.9}, \textcolor{red}{72.4}) & (\textcolor{blue}{74.1}, \textcolor{blue}{76.8}) & (52.0, 54.2) & (\textcolor{red}{99.2}, \textcolor{red}{99.2}) & (60.1, 62.3) & (\textcolor{blue}{74.3}, \textcolor{blue}{78.7}) & (\textcolor{red}{87.3}, \textcolor{red}{87.6}) & (76.9, \textcolor{blue}{81.1}) & (\textcolor{red}{89.5}, \textcolor{red}{88.9}) & (\textcolor{blue}{80.9}, \textcolor{blue}{87.2}) & (69.0, 72.3) & (\textcolor{red}{74.8}, \textcolor{red}{76.9}) \\
\midrule
\multirow{5}{*}{\makecell[c]{L.}}
& CLIP + Rendering & (48.4, -) & (48, -) & (33.7, -) & (47.1, -) & (31.6, -) & (49.6, -) & (56.5, -) & (50.3, -) & (34.7, -) & (47.5, -) & (54, -) & (49.2, -) & (-) \\

& Cheraghian &(-, -)& (-, -) & (-, -) &(-, -) & (-, -)  &(-, -) &(-, -) &(-, -)&(-, -)&(-, -)&(-, -)&(-, -)&(-, -)\\

& PoinCLIPV2 & (45.1, -) & (56.3, -) & (55.2, -) & (46.5, -) & (52.6, -) & (\textcolor{red}{62.6}, -) & (63.9, -) & (51.5, -) & (48.3, -) & (58.1, -) & (40.6, -) & (53.6, -) & (52.9, -) \\

& PoinCLIPV2$_a$ & (49.6, -) & (54.3, -) & (54.2, -) & (47.1, -) & (53.2, -) & (\textcolor{blue}{61.6}, -) & (59.2, -) & (51.6, -) & (47.2, -) & (59.3, -) & (41.2, -) & (48.1, -) & (52.2, -) \\

& AnomalyCLIP & (51.1, -) & (48.8, -) & (51.7, -) & (50.0, -) & (55.2, -) & (48.9, -) & (46.5, -) & (48.9, -) & (49.2, -) & (50.6, -) & (51.0, -) & (51.1, -) & (50.3, -) \\

& PointAD-CoOp & (\textcolor{blue}{65.5}, -) & (\textcolor{red}{75.5}, -) & (\textcolor{blue}{67.1}, -) & (\textcolor{blue}{64.2}, -) & (\textcolor{blue}{87.2}, -) & (50.8, -) & (\textcolor{blue}{79.1}, -) & (\textcolor{red}{81.7}, -) & (\textcolor{red}{77.0}, -) & (\textcolor{blue}{77.3}, -) & (\textcolor{blue}{77.6}, -) & (\textcolor{blue}{68.4}, -) & (\textcolor{blue}{72.6}, -) \\

\rowcolor{gray!40}
& Point-AD & (\textcolor{red}{67.2}, -) & (\textcolor{blue}{72.3}, -) & (\textcolor{red}{71.3}, -) & (\textcolor{red}{67.7}, -) & (\textcolor{red}{87.7}, -) & (51.0, -) & (\textcolor{red}{80.1}, -) & (\textcolor{blue}{80.2}, -) & (\textcolor{blue}{74.8}, -) & (\textcolor{red}{77.8}, -) & (\textcolor{red}{81.4}, -) & (\textcolor{red}{70.0}, -) & (\textcolor{red}{73.5}, -) \\
\bottomrule
\end{tabular}
\vspace{-1.5em}
\end{table*}
\newpage

\begin{table*}
\centering
\caption{Perfromance comparison on ZS 3D cross-dataset anomaly detection transferring from MVTec3D-AD to Eyecandies.}
\vspace{-0.6em}
\tiny
\setlength\tabcolsep{2pt}
\label{table:3D Eyecandies transfer}
\tiny
\setlength\tabcolsep{3pt}
\begin{tabular}{cccccccccccc|c}
\toprule
& Method & \makecell{Candy \\ Cane} & \makecell{Chocolate \\ Cookie} & \makecell{Chocolate \\ Praline} & \makecell{Confetto} & \makecell{Gummy \\ Bear} & \makecell{Hazelnut \\ Truffle} & \makecell{Licorice \\ Sandwich} & \makecell{Lollipop} & \makecell{Marsh- \\ mallow} & \makecell{Peppermint \\ Candy} & Mean \\
\midrule
\multirow{2}{*}{\makecell[c]{G.}}

& PoinCLIPV2$_a$ & (47.8, 52.7) & (\textcolor{blue}{51.9}, 55.6) & (31.8, 42.4) & (38.6, 44.9) & (46.0, 49.3) & (32.0, 40.3) & (54.5, 54.3) & (42.7, 30.4) & (58.8, 59.8) & (47.7, 49.8) & (45.2, 48.0) \\

& AnomalyCLIP & (49.6, 52.0) & (43.1, 48.1) & (65.2, 67.3) & (69.5, 70.4) & (41.2, 44.6) & (53.8, 56.6) & (51.2, 55.0) & (60.4, 45.4) & (58.1, 58.3) & (70.6, 73.3) & (56.3, 57.1) \\

& PointAD-CoOp & (\textcolor{red}{54.5}, \textcolor{blue}{58.5}) & (51.6, \textcolor{blue}{59.1}) & (\textcolor{blue}{72.8}, \textcolor{blue}{81.1}) & (\textcolor{blue}{79.3}, \textcolor{blue}{86.4}) & (\textcolor{red}{69.1}, \textcolor{red}{76.8}) & (\textcolor{red}{62.6}, \textcolor{red}{64.3}) & (\textcolor{blue}{78.9}, \textcolor{blue}{83.6}) & (\textcolor{blue}{74.4}, \textcolor{blue}{67.5}) & (\textcolor{red}{67.2}, \textcolor{red}{73.7}) & (\textcolor{red}{81.1}, \textcolor{red}{87.0}) & (\textcolor{blue}{69.1}, \textcolor{blue}{73.8}) \\

\rowcolor{gray!40}
& Point-AD & (\textcolor{blue}{51.7}, \textcolor{red}{58.9}) & (\textcolor{red}{57.9}, \textcolor{red}{64.7}) & (\textcolor{red}{76.8}, \textcolor{red}{84.4}) & (\textcolor{red}{79.9}, \textcolor{red}{87.2}) & (\textcolor{blue}{66.5}, \textcolor{blue}{75.0}) & (\textcolor{blue}{61.3}, \textcolor{blue}{61.6}) & (\textcolor{red}{81.0}, \textcolor{red}{85.3}) & (\textcolor{red}{77.3}, \textcolor{red}{71.7}) & (\textcolor{blue}{63.3}, \textcolor{blue}{68.5}) & (\textcolor{blue}{79.3}, \textcolor{blue}{86.0}) & (\textcolor{red}{69.5}, \textcolor{red}{74.3}) \\
\midrule
\multirow{2}{*}{\makecell[c]{L.}}

& PoinCLIPV2$_a$ & (45.0, -) & (38.4, -) & (49.2, -) & (43.2, -) & (45.0, -) & (55.2, 21.6) & (38.7, -) & (43.4, -) & (42.9, -) & (37.9, -) & (43.9, -) \\

& AnomalyCLIP & (\textcolor{blue}{95.9}, 84.0) & (76.7, 34.1) & (80.1, 46.7) & (78.7, 44.2) & (80.4, 42.7) & (73.5, 33.1) & (80.7, 40.8) & (90.8, 66.1) & (60.0, 17.4) & (79.4, 45.0) & (\textcolor{blue}{79.6}, 45.4) \\

& PointAD-CoOp & (\textcolor{red}{97.5}, \textcolor{red}{85.7}) & (\textcolor{blue}{92.4}, \textcolor{blue}{74.4}) & (\textcolor{red}{92.1}, \textcolor{red}{70.6}) & (\textcolor{blue}{94.0}, \textcolor{blue}{67.3}) & (\textcolor{red}{89.0}, \textcolor{red}{69.3}) & (\textcolor{red}{84.4}, \textcolor{red}{46.5}) & (\textcolor{blue}{93.1}, \textcolor{blue}{72.7}) & (\textcolor{blue}{97.8}, \textcolor{blue}{86.8}) & (\textcolor{red}{86.1}, \textcolor{red}{59.4}) & (\textcolor{blue}{92.0}, \textcolor{blue}{71.8}) & (\textcolor{red}{91.8}, \textcolor{blue}{70.5}) \\

\rowcolor{gray!40}
& Point-AD & (\textcolor{red}{97.5}, \textcolor{blue}{85.4}) & (\textcolor{red}{93.2}, \textcolor{red}{76.8}) & (\textcolor{blue}{91.9}, \textcolor{blue}{70.2}) & (\textcolor{red}{94.5}, \textcolor{red}{72.3}) & (\textcolor{blue}{88.6}, \textcolor{blue}{68.3}) & (\textcolor{blue}{82.6}, \textcolor{blue}{45.4}) & (\textcolor{red}{93.3}, \textcolor{red}{74.0}) & (\textcolor{red}{98.0}, \textcolor{red}{89.0}) & (\textcolor{blue}{85.8}, \textcolor{blue}{58.9}) & (\textcolor{red}{92.1}, \textcolor{red}{73.3}) & (\textcolor{red}{91.8}, \textcolor{red}{71.4}) \\
\bottomrule
\end{tabular}
\vspace{-1.5em}
\end{table*}
\newpage

\begin{table*}
\centering
\caption{Perfromance comparison on ZS M3D cross-dataset anomaly detection transferring from MVTec3D-AD to Eyecandies.}
\tiny
\setlength\tabcolsep{2pt}
\label{table:multimodal Eyecandies transfer.}
\tiny
\setlength\tabcolsep{3pt}
\begin{tabular}{cccccccccccc|c}
\toprule
& Method & \makecell{Candy \\ Cane} & \makecell{Chocolate \\ Cookie} & \makecell{Chocolate \\ Praline} & \makecell{Confetto} & \makecell{Gummy \\ Bear} & \makecell{Hazelnut \\ Truffle} & \makecell{Licorice \\ Sandwich} & \makecell{Lollipop} & \makecell{Marsh- \\ mallow} & \makecell{Peppermint \\ Candy} & Mean \\
\midrule
\multirow{2}{*}{\makecell[c]{MG.}}

& PoinCLIPV2$_a$ & (42.1, 50.4) & (45.5, 54.9) & (49.0, 52.5) & (57.1, 54.3) & (44.9, 45.8) & (53.6, 56.6) & (59.0, 64.5) & (47.4, 36.1) & (54.1, 53.3) & (32.6, 40.3) & (48.5, 50.9) \\

& AnomalyCLIP & (44.9, 51.4) & (66.1, 68.9) & (76.4, 77.7) & (79.3, 82.7) & (51.0, 56.7) & (55.1, 58.3) & (75.2, 79.6) & (65.4, 57.9) & (70.3, 72.5) & (72.8, 75.5) & (65.7, 68.1) \\

& PointAD-CoOp & (\textcolor{red}{53.4}, \textcolor{red}{58.6}) & (\textcolor{blue}{72.8}, \textcolor{blue}{80.4}) & (\textcolor{blue}{83.6}, \textcolor{blue}{86.6}) & (\textcolor{blue}{89.1}, \textcolor{blue}{92.8}) & (\textcolor{red}{73.9}, \textcolor{red}{76.2}) & (\textcolor{red}{72.0}, \textcolor{red}{76.2}) & (\textcolor{blue}{84.6}, \textcolor{blue}{88.4}) & (\textcolor{blue}{65.7}, \textcolor{blue}{54.3}) & (\textcolor{blue}{78.4}, \textcolor{blue}{83.3}) & (\textcolor{red}{89.6}, \textcolor{red}{92.7}) & (\textcolor{blue}{76.3}, \textcolor{blue}{78.9}) \\

\rowcolor{gray!40}
& Point-AD & (\textcolor{blue}{49.4}, \textcolor{blue}{55.1}) & (\textcolor{red}{87.4}, \textcolor{red}{91.2}) & (\textcolor{red}{87.5}, \textcolor{red}{88.8}) & (\textcolor{red}{91.0}, \textcolor{red}{94.1}) & (\textcolor{blue}{71.4}, \textcolor{blue}{74.2}) & (\textcolor{blue}{70.2}, \textcolor{blue}{75.0}) & (\textcolor{red}{88.0}, \textcolor{red}{89.5}) & (\textcolor{red}{73.2}, \textcolor{red}{63.5}) & (\textcolor{red}{79.4}, \textcolor{red}{84.4}) & (\textcolor{blue}{88.2}, \textcolor{blue}{91.9}) & (\textcolor{red}{78.6}, \textcolor{red}{80.8}) \\
\midrule
\multirow{2}{*}{\makecell[c]{ML.}}

& PoinCLIPV2$_a$ & (45.1, -) & (47.2, -) & (48.1, -) & (63.1, -) & (49.7, -) & (54.0, 15.6) & (45.2, -) & (33.8, -) & (44.7, -) & (41.9, -) & (47.3, -) \\

& AnomalyCLIP & (\textcolor{red}{95.1}, \textcolor{red}{82.3}) & (91.6, 71.2) & (84.3, 62.0) & (94.1, 80.9) & (80.6, 44.5) & (73.6, 42.7) & (89.6, 66.9) & (92.6, 68.0) & (75.9, 42.8) & (84.9, 52.1) & (86.2, 61.3) \\

& PointAD-CoOp & (\textcolor{blue}{92.0}, \textcolor{blue}{71.0}) & (\textcolor{blue}{96.8}, \textcolor{blue}{86.4}) & (\textcolor{red}{94.2}, \textcolor{red}{86.2}) & (\textcolor{blue}{97.9}, \textcolor{blue}{92.5}) & (\textcolor{blue}{87.9}, \textcolor{blue}{64.8}) & (\textcolor{red}{91.3}, \textcolor{blue}{68.6}) & (\textcolor{red}{96.2}, \textcolor{blue}{83.6}) & (\textcolor{red}{96.5}, \textcolor{red}{82.5}) & (\textcolor{blue}{95.3}, \textcolor{blue}{81.3}) & (\textcolor{red}{96.1}, \textcolor{red}{86.7}) & (\textcolor{red}{94.4}, \textcolor{blue}{80.3})\\

\rowcolor{gray!40}
& Point-AD & (88.8, 63.9) & (\textcolor{red}{97.4}, \textcolor{red}{88.9}) & (\textcolor{blue}{92.5}, \textcolor{blue}{85.1}) & (\textcolor{red}{99.1}, \textcolor{red}{96.6}) & (\textcolor{red}{89.9}, \textcolor{red}{70.4}) & (\textcolor{blue}{89.9}, \textcolor{red}{70.4}) & (\textcolor{blue}{95.5}, \textcolor{red}{84.4}) & (\textcolor{blue}{95.7}, \textcolor{blue}{79.8}) & (\textcolor{red}{95.8}, \textcolor{red}{82.5}) & (\textcolor{blue}{95.2}, \textcolor{blue}{84.8}) & (\textcolor{blue}{94.0}, \textcolor{red}{80.7}) \\
\bottomrule
\end{tabular}
\vspace{-1.5em}
\end{table*}
\newpage

\begin{table*}
\centering
\caption{Perfromance comparison on ZS cross-dataset anomaly detection transferring from MVTec3D-AD to Real3D-AD.}
\tiny
\setlength\tabcolsep{2pt}
\label{table:3D Real3D-AD transfer.}
\tiny
\setlength\tabcolsep{3pt}
\begin{tabular}{cccccccccccccc|c}
\toprule
& Method & airplane & car & candybar & chicken & diamond & duck & fish & gemstone & seahorse & shell & starfish & toffees & Mean \\
\midrule
\multirow{2}{*}{\makecell[c]{G.}}
& PoinCLIPV2$_a$ & (45.9, 46.4) & (47.0, 47.2) & (49.4, 48.7) & (48.3, 57.1) & (47.6, 50.1) & (64.6, 62.1) & (60.3, 67.4) & (60.2, 60.8) & (72.4, 73.8) & (60.8, 58.9) & (60.4, 58.5) & (72.3, 74.7) & (57.4, 58.8) \\

& AnomalyCLIP & (48.3, 51.8) & (\textcolor{blue}{63.3}, \textcolor{red}{67.7}) & (48.8, 50.1) & (51.9, 62.0) & (60.2, 59.2) & (42.6, 46.8) & (57.1, 58.7) & (54.6, 58.2) & (60.0, 60.1) & (51.1, 50.1) & (40.3, 45.6) & (53.8, 58.0) & (52.7, 55.7) \\

& PointAD-CoOp & (\textcolor{blue}{55.2}, \textcolor{blue}{55.7}) & (\textcolor{red}{64.0}, \textcolor{blue}{66.2}) & (\textcolor{blue}{68.3}, \textcolor{blue}{68.7}) & (\textcolor{blue}{57.2}, \textcolor{blue}{68.6}) & (\textcolor{blue}{99.3}, \textcolor{blue}{99.3}) & (\textcolor{red}{75.0}, \textcolor{red}{73.4}) & (\textcolor{red}{78.1}, \textcolor{red}{82.4}) & (\textcolor{blue}{85.8}, \textcolor{blue}{85.5}) & (\textcolor{blue}{73.5}, \textcolor{blue}{78.2}) & (\textcolor{red}{89.8}, \textcolor{blue}{86.8}) & (\textcolor{blue}{76.1}, \textcolor{blue}{81.4}) & (\textcolor{red}{75.6}, \textcolor{blue}{76.2}) & (\textcolor{blue}{74.8}, \textcolor{blue}{76.9}) \\

\rowcolor{gray!40}
& Point-AD & (\textcolor{red}{64.6}, \textcolor{red}{62.8}) & (\textcolor{red}{64.0}, 65.1) & (\textcolor{red}{72.0}, \textcolor{red}{71.5}) & (\textcolor{red}{61.6}, \textcolor{red}{68.9}) & (\textcolor{red}{99.5}, \textcolor{red}{99.6}) & (\textcolor{blue}{69.0}, \textcolor{blue}{67.2}) & (\textcolor{blue}{74.5}, \textcolor{blue}{79.4}) & (\textcolor{red}{90.4}, \textcolor{red}{90.1}) & (\textcolor{red}{74.8}, \textcolor{red}{79.3}) & (\textcolor{blue}{89.7}, \textcolor{red}{90.5}) & (\textcolor{red}{77.9}, \textcolor{red}{82.6}) & (\textcolor{blue}{73.3}, \textcolor{red}{77.9}) & (\textcolor{red}{75.9}, \textcolor{red}{77.9}) \\
\midrule
\multirow{2}{*}{\makecell[c]{L.}}
& PoinCLIPV2$_a$ & (50.5, -) & (53.6, -) & (54.2, -) & (47.0, -) & (54.1, -) & (\textcolor{red}{60.7}, -) & (59.4, -) & (51.2, -) & (47.1, -) & (55.4, 18.2) & (41.6, -) & (47.5, -) & (51.9, -)\\

& AnomalyCLIP & (50.9, -) & (49.6, -) & (49.8, -) & (50.1, -) & (57.5, -) & (47.9, -) & (48.6, -) & (48.3, -) & (50.2, -) & (50.5, -) & (49.3, -) & (51.0, -) & (50.3, -) \\

& PointAD-CoOp & (\textcolor{blue}{61.4}, -) & (\textcolor{red}{71.2}, -) & (\textcolor{blue}{64.6}, -) & (\textcolor{red}{67.7}, -) & (\textcolor{blue}{85.0}, -) & (\textcolor{blue}{54.9}, -) & (\textcolor{red}{76.9}, -) & (\textcolor{blue}{78.0}, -) & (\textcolor{blue}{69.3}, -) & (\textcolor{blue}{77.3}, -) & (\textcolor{blue}{59.7}, -) & (\textcolor{blue}{75.2}, -) & (\textcolor{blue}{70.1}, -) \\

\rowcolor{gray!40}
& Point-AD & (\textcolor{red}{64.8}, -) & (\textcolor{blue}{69.7}, -) & (\textcolor{red}{70.4}, -) & (\textcolor{blue}{67.3}, -) & (\textcolor{red}{86.6}, -) & (50.3, -) & (\textcolor{blue}{75.6}, -) & (\textcolor{red}{78.9}, -) & (\textcolor{red}{74.1}, -) & (\textcolor{red}{78.1}, -) & (\textcolor{red}{66.8}, -) & (\textcolor{red}{76.4}, -) & (\textcolor{red}{71.6}, -) \\
\bottomrule
\end{tabular}
\end{table*}

\clearpage
\section*{NeurIPS Paper Checklist}

\begin{enumerate}

\item {\bf Claims}
    \item[] Question: Do the main claims made in the abstract and introduction accurately reflect the paper's contributions and scope?
    \item[] Answer: \answerYes{} % Replace by \answerYes{}, \answerNo{}, or \answerNA{}.
    \item[] Justification: The abstract and introduction precisely reflect the contribution and scope of this paper.
    \item[] Guidelines:
    \begin{itemize}
        \item The answer NA means that the abstract and introduction do not include the claims made in the paper.
        \item The abstract and/or introduction should clearly state the claims made, including the contributions made in the paper and important assumptions and limitations. A No or NA answer to this question will not be perceived well by the reviewers. 
        \item The claims made should match theoretical and experimental results, and reflect how much the results can be expected to generalize to other settings. 
        \item It is fine to include aspirational goals as motivation as long as it is clear that these goals are not attained by the paper. 
    \end{itemize}

\item {\bf Limitations}
    \item[] Question: Does the paper discuss the limitations of the work performed by the authors?
    \item[] Answer: \answerYes{} % Replace by \answerYes{}, \answerNo{}, or \answerNA{}.
    \item[] Justification: we have created a separate "Limitations" section in our paper.
    \item[] Guidelines:
    \begin{itemize}
        \item The answer NA means that the paper has no limitation while the answer No means that the paper has limitations, but those are not discussed in the paper. 
        \item The authors are encouraged to create a separate "Limitations" section in their paper.
        \item The paper should point out any strong assumptions and how robust the results are to violations of these assumptions (e.g., independence assumptions, noiseless settings, model well-specification, asymptotic approximations only holding locally). The authors should reflect on how these assumptions might be violated in practice and what the implications would be.
        \item The authors should reflect on the scope of the claims made, e.g., if the approach was only tested on a few datasets or with a few runs. In general, empirical results often depend on implicit assumptions, which should be articulated.
        \item The authors should reflect on the factors that influence the performance of the approach. For example, a facial recognition algorithm may perform poorly when image resolution is low or images are taken in low lighting. Or a speech-to-text system might not be used reliably to provide closed captions for online lectures because it fails to handle technical jargon.
        \item The authors should discuss the computational efficiency of the proposed algorithms and how they scale with dataset size.
        \item If applicable, the authors should discuss possible limitations of their approach to address problems of privacy and fairness.
        \item While the authors might fear that complete honesty about limitations might be used by reviewers as grounds for rejection, a worse outcome might be that reviewers discover limitations that aren't acknowledged in the paper. The authors should use their best judgment and recognize that individual actions in favor of transparency play an important role in developing norms that preserve the integrity of the community. Reviewers will be specifically instructed to not penalize honesty concerning limitations.
    \end{itemize}

\item {\bf Theory Assumptions and Proofs}
    \item[] Question: For each theoretical result, does the paper provide the full set of assumptions and a complete (and correct) proof?
    \item[] Answer: \answerNA{} % Replace by \answerYes{}, \answerNo{}, or \answerNA{}.
    \item[] Justification: This paper does not include theoretical results
    \item[] Guidelines:
    \begin{itemize}
        \item The answer NA means that the paper does not include theoretical results. 
        \item All the theorems, formulas, and proofs in the paper should be numbered and cross-referenced.
        \item All assumptions should be clearly stated or referenced in the statement of any theorems.
        \item The proofs can either appear in the main paper or the supplemental material, but if they appear in the supplemental material, the authors are encouraged to provide a short proof sketch to provide intuition. 
        \item Inversely, any informal proof provided in the core of the paper should be complemented by formal proofs provided in appendix or supplemental material.
        \item Theorems and Lemmas that the proof relies upon should be properly referenced. 
    \end{itemize}

    \item {\bf Experimental Result Reproducibility}
    \item[] Question: Does the paper fully disclose all the information needed to reproduce the main experimental results of the paper to the extent that it affects the main claims and/or conclusions of the paper (regardless of whether the code and data are provided or not)?
    \item[] Answer: \answerYes{} % Replace by \answerYes{}, \answerNo{}, or \answerNA{}.
    \item[] Justification: We provide a detailed illustration of our proposed algorithm and baselines in the Appendix~\ref{appendix: baselines}, Appendix~\ref{appendix: implementation details}, and Appendix~\ref{appendix: hyperparameter}.
    \item[] Guidelines:
    \begin{itemize}
        \item The answer NA means that the paper does not include experiments.
        \item If the paper includes experiments, a No answer to this question will not be perceived well by the reviewers: Making the paper reproducible is important, regardless of whether the code and data are provided or not.
        \item If the contribution is a dataset and/or model, the authors should describe the steps taken to make their results reproducible or verifiable. 
        \item Depending on the contribution, reproducibility can be accomplished in various ways. For example, if the contribution is a novel architecture, describing the architecture fully might suffice, or if the contribution is a specific model and empirical evaluation, it may be necessary to either make it possible for others to replicate the model with the same dataset, or provide access to the model. In general. releasing code and data is often one good way to accomplish this, but reproducibility can also be provided via detailed instructions for how to replicate the results, access to a hosted model (e.g., in the case of a large language model), releasing of a model checkpoint, or other means that are appropriate to the research performed.
        \item While NeurIPS does not require releasing code, the conference does require all submissions to provide some reasonable avenue for reproducibility, which may depend on the nature of the contribution. For example
        \begin{enumerate}
            \item If the contribution is primarily a new algorithm, the paper should make it clear how to reproduce that algorithm.
            \item If the contribution is primarily a new model architecture, the paper should describe the architecture clearly and fully.
            \item If the contribution is a new model (e.g., a large language model), then there should either be a way to access this model for reproducing the results or a way to reproduce the model (e.g., with an open-source dataset or instructions for how to construct the dataset).
            \item We recognize that reproducibility may be tricky in some cases, in which case authors are welcome to describe the particular way they provide for reproducibility. In the case of closed-source models, it may be that access to the model is limited in some way (e.g., to registered users), but it should be possible for other researchers to have some path to reproducing or verifying the results.
        \end{enumerate}
    \end{itemize}

\item {\bf Open access to data and code}
    \item[] Question: Does the paper provide open access to the data and code, with sufficient instructions to faithfully reproduce the main experimental results, as described in supplemental material?
    \item[] Answer: \answerNo{} % Replace by \answerYes{}, \answerNo{}, or \answerNA{}.
    \item[] Justification: we will make our code and dataset available once the paper is accepted.
    \item[] Guidelines:
    \begin{itemize}
        \item The answer NA means that paper does not include experiments requiring code.
        \item Please see the NeurIPS code and data submission guidelines (\url{https://nips.cc/public/guides/CodeSubmissionPolicy}) for more details.
        \item While we encourage the release of code and data, we understand that this might not be possible, so “No” is an acceptable answer. Papers cannot be rejected simply for not including code, unless this is central to the contribution (e.g., for a new open-source benchmark).
        \item The instructions should contain the exact command and environment needed to run to reproduce the results. See the NeurIPS code and data submission guidelines (\url{https://nips.cc/public/guides/CodeSubmissionPolicy}) for more details.
        \item The authors should provide instructions on data access and preparation, including how to access the raw data, preprocessed data, intermediate data, and generated data, etc.
        \item The authors should provide scripts to reproduce all experimental results for the new proposed method and baselines. If only a subset of experiments are reproducible, they should state which ones are omitted from the script and why.
        \item At submission time, to preserve anonymity, the authors should release anonymized versions (if applicable).
        \item Providing as much information as possible in supplemental material (appended to the paper) is recommended, but including URLs to data and code is permitted.
    \end{itemize}

\item {\bf Experimental Setting/Details}
    \item[] Question: Does the paper specify all the training and test details (e.g., data splits, hyperparameters, how they were chosen, type of optimizer, etc.) necessary to understand the results?
    \item[] Answer: \answerYes{} % Replace by \answerYes{}, \answerNo{}, or \answerNA{}.
    \item[] Justification: We provide full details in Appendix~\ref{appendix: implementation details} and Appendix~\ref{appendix: hyperparameter}  \item[] Guidelines:
    \begin{itemize}
        \item The answer NA means that the paper does not include experiments.
        \item The experimental setting should be presented in the core of the paper to a level of detail that is necessary to appreciate the results and make sense of them.
        \item The full details can be provided either with the code, in appendix, or as supplemental material.
    \end{itemize}

\item {\bf Experiment Statistical Significance}
    \item[] Question: Does the paper report error bars suitably and correctly defined or other appropriate information about the statistical significance of the experiments?
    \item[] Answer: \answerYes{} % Replace by \answerYes{}, \answerNo{}, or \answerNA{}.
    \item[] Justification: We report the average results across three runs in Section~\ref{main results}.
    \item[] Guidelines:
    \begin{itemize}
        \item The answer NA means that the paper does not include experiments.
        \item The authors should answer "Yes" if the results are accompanied by error bars, confidence intervals, or statistical significance tests, at least for the experiments that support the main claims of the paper.
        \item The factors of variability that the error bars are capturing should be clearly stated (for example, train/test split, initialization, random drawing of some parameter, or overall run with given experimental conditions).
        \item The method for calculating the error bars should be explained (closed form formula, call to a library function, bootstrap, etc.)
        \item The assumptions made should be given (e.g., Normally distributed errors).
        \item It should be clear whether the error bar is the standard deviation or the standard error of the mean.
        \item It is OK to report 1-sigma error bars, but one should state it. The authors should preferably report a 2-sigma error bar than state that they have a 96\% CI, if the hypothesis of Normality of errors is not verified.
        \item For asymmetric distributions, the authors should be careful not to show in tables or figures symmetric error bars that would yield results that are out of range (e.g. negative error rates).
        \item If error bars are reported in tables or plots, The authors should explain in the text how they were calculated and reference the corresponding figures or tables in the text.
    \end{itemize}

\item {\bf Experiments Compute Resources}
    \item[] Question: For each experiment, does the paper provide sufficient information on the computer resources (type of compute workers, memory, time of execution) needed to reproduce the experiments?
    \item[] Answer: \answerYes{} % Replace by \answerYes{}, \answerNo{}, or \answerNA{}.
    \item[] Justification: We point out the specific compute resources in Section~\ref{section: implementation details}
    \item[] Guidelines:
    \begin{itemize}
        \item The answer NA means that the paper does not include experiments.
        \item The paper should indicate the type of compute workers CPU or GPU, internal cluster, or cloud provider, including relevant memory and storage.
        \item The paper should provide the amount of compute required for each of the individual experimental runs as well as estimate the total compute. 
        \item The paper should disclose whether the full research project required more compute than the experiments reported in the paper (e.g., preliminary or failed experiments that didn't make it into the paper). 
    \end{itemize}
    
\item {\bf Code Of Ethics}
    \item[] Question: Does the research conducted in the paper conform, in every respect, with the NeurIPS Code of Ethics \url{https://neurips.cc/public/EthicsGuidelines}?
    \item[] Answer: \answerYes{} % Replace by \answerYes{}, \answerNo{}, or \answerNA{}.
    \item[] Justification: Our paper obeys the NeurIPS Code of Ethics.
    \item[] Guidelines:
    \begin{itemize}
        \item The answer NA means that the authors have not reviewed the NeurIPS Code of Ethics.
        \item If the authors answer No, they should explain the special circumstances that require a deviation from the Code of Ethics.
        \item The authors should make sure to preserve anonymity (e.g., if there is a special consideration due to laws or regulations in their jurisdiction).
    \end{itemize}

\item {\bf Broader Impacts}
    \item[] Question: Does the paper discuss both potential positive societal impacts and negative societal impacts of the work performed?
    \item[] Answer: \answerYes{} % Replace by \answerYes{}, \answerNo{}, or \answerNA{}.
    \item[] Justification: We discuss the broader impacts of our paper.
    \item[] Guidelines:
    \begin{itemize}
        \item The answer NA means that there is no societal impact of the work performed.
        \item If the authors answer NA or No, they should explain why their work has no societal impact or why the paper does not address societal impact.
        \item Examples of negative societal impacts include potential malicious or unintended uses (e.g., disinformation, generating fake profiles, surveillance), fairness considerations (e.g., deployment of technologies that could make decisions that unfairly impact specific groups), privacy considerations, and security considerations.
        \item The conference expects that many papers will be foundational research and not tied to particular applications, let alone deployments. However, if there is a direct path to any negative applications, the authors should point it out. For example, it is legitimate to point out that an improvement in the quality of generative models could be used to generate deepfakes for disinformation. On the other hand, it is not needed to point out that a generic algorithm for optimizing neural networks could enable people to train models that generate Deepfakes faster.
        \item The authors should consider possible harms that could arise when the technology is being used as intended and functioning correctly, harms that could arise when the technology is being used as intended but gives incorrect results, and harms following from (intentional or unintentional) misuse of the technology.
        \item If there are negative societal impacts, the authors could also discuss possible mitigation strategies (e.g., gated release of models, providing defenses in addition to attacks, mechanisms for monitoring misuse, mechanisms to monitor how a system learns from feedback over time, improving the efficiency and accessibility of ML).
    \end{itemize}
    
\item {\bf Safeguards}
    \item[] Question: Does the paper describe safeguards that have been put in place for responsible release of data or models that have a high risk for misuse (e.g., pretrained language models, image generators, or scraped datasets)?
    \item[] Answer: \answerNA{} % Replace by \answerYes{}, \answerNo{}, or \answerNA{}.
    \item[] Justification: our paper poses no such risks
    \item[] Guidelines:
    \begin{itemize}
        \item The answer NA means that the paper poses no such risks.
        \item Released models that have a high risk for misuse or dual-use should be released with necessary safeguards to allow for controlled use of the model, for example by requiring that users adhere to usage guidelines or restrictions to access the model or implementing safety filters. 
        \item Datasets that have been scraped from the Internet could pose safety risks. The authors should describe how they avoided releasing unsafe images.
        \item We recognize that providing effective safeguards is challenging, and many papers do not require this, but we encourage authors to take this into account and make a best faith effort.
    \end{itemize}

\item {\bf Licenses for existing assets}
    \item[] Question: Are the creators or original owners of assets (e.g., code, data, models), used in the paper, properly credited and are the license and terms of use explicitly mentioned and properly respected?
    \item[] Answer: \answerYes{} % Replace by \answerYes{}, \answerNo{}, or \answerNA{}.
    \item[] Justification: we respect the Licenses for existing assets that we use.
    \item[] Guidelines:
    \begin{itemize}
        \item The answer NA means that the paper does not use existing assets.
        \item The authors should cite the original paper that produced the code package or dataset.
        \item The authors should state which version of the asset is used and, if possible, include a URL.
        \item The name of the license (e.g., CC-BY 4.0) should be included for each asset.
        \item For scraped data from a particular source (e.g., website), the copyright and terms of service of that source should be provided.
        \item If assets are released, the license, copyright information, and terms of use in the package should be provided. For popular datasets, \url{paperswithcode.com/datasets} has curated licenses for some datasets. Their licensing guide can help determine the license of a dataset.
        \item For existing datasets that are re-packaged, both the original license and the license of the derived asset (if it has changed) should be provided.
        \item If this information is not available online, the authors are encouraged to reach out to the asset's creators.
    \end{itemize}

\item {\bf New Assets}
    \item[] Question: Are new assets introduced in the paper well documented and is the documentation provided alongside the assets?
    \item[] Answer: \answerNA{} % Replace by \answerYes{}, \answerNo{}, or \answerNA{}.
    \item[] Justification: we will release new assets proposed in our paper once the paper is accepted. 
    \item[] Guidelines:
    \begin{itemize}
        \item The answer NA means that the paper does not release new assets.
        \item Researchers should communicate the details of the dataset/code/model as part of their submissions via structured templates. This includes details about training, license, limitations, etc. 
        \item The paper should discuss whether and how consent was obtained from people whose asset is used.
        \item At submission time, remember to anonymize your assets (if applicable). You can either create an anonymized URL or include an anonymized zip file.
    \end{itemize}

\item {\bf Crowdsourcing and Research with Human Subjects}
    \item[] Question: For crowdsourcing experiments and research with human subjects, does the paper include the full text of instructions given to participants and screenshots, if applicable, as well as details about compensation (if any)? 
    \item[] Answer: \answerNA{} % Replace by \answerYes{}, \answerNo{}, or \answerNA{}.
    \item[] Justification: our paper does not involve crowdsourcing nor research with human subjects.
    \item[] Guidelines:
    \begin{itemize}
        \item The answer NA means that the paper does not involve crowdsourcing nor research with human subjects.
        \item Including this information in the supplemental material is fine, but if the main contribution of the paper involves human subjects, then as much detail as possible should be included in the main paper. 
        \item According to the NeurIPS Code of Ethics, workers involved in data collection, curation, or other labor should be paid at least the minimum wage in the country of the data collector. 
    \end{itemize}

\item {\bf Institutional Review Board (IRB) Approvals or Equivalent for Research with Human Subjects}
    \item[] Question: Does the paper describe potential risks incurred by study participants, whether such risks were disclosed to the subjects, and whether Institutional Review Board (IRB) approvals (or an equivalent approval/review based on the requirements of your country or institution) were obtained?
    \item[] Answer: \answerNA{} % Replace by \answerYes{}, \answerNo{}, or \answerNA{}.
    \item[] Justification: our paper does not involve crowdsourcing nor research with human subjects.
    \item[] Guidelines:
    \begin{itemize}
        \item The answer NA means that the paper does not involve crowdsourcing nor research with human subjects.
        \item Depending on the country in which research is conducted, IRB approval (or equivalent) may be required for any human subjects research. If you obtained IRB approval, you should clearly state this in the paper. 
        \item We recognize that the procedures for this may vary significantly between institutions and locations, and we expect authors to adhere to the NeurIPS Code of Ethics and the guidelines for their institution. 
        \item For initial submissions, do not include any information that would break anonymity (if applicable), such as the institution conducting the review.
    \end{itemize}

\end{enumerate}
\end{document}